\pdfoutput=1
\documentclass{article}

\usepackage[table, dvipsnames]{xcolor}
\usepackage[preprint]{neurips_2026}

\usepackage[utf8]{inputenc} 
\usepackage[T1]{fontenc}    
\usepackage{hyperref}       
\usepackage{url}            
\usepackage{booktabs}       
\usepackage{amsfonts}       
\usepackage{nicefrac}       
\usepackage{microtype}      
\usepackage{amsmath}
\usepackage{amssymb}
\usepackage{array}
\usepackage{tablefootnote}
\usepackage[most]{tcolorbox}
\usepackage{nicematrix}
\usepackage{tabularx}
\usepackage{makecell}
\usepackage{pifont}         
\usepackage{arydshln}       
\usepackage{tikz}
\usepackage{multirow}
\usepackage{caption}
\usepackage{xcolor}
\usepackage{wrapfig}
\usepackage{stackengine}
\usepackage{colortbl}
\usepackage{makecell}
\usepackage{multirow}
\usepackage{xcolor}

\definecolor{citecolor}{RGB}{255, 0, 0}
\definecolor{mvcolor}{RGB}{0, 0, 255}
\definecolor{mvncolor}{RGB}{0, 0, 0}
\definecolor{rhcolor}{RGB}{0, 100, 0}
\definecolor{lavender}{RGB}{245, 240, 255}
\definecolor{softgold}{RGB}{255, 248, 220}
\definecolor{pastelgreen}{HTML}{E2F0D9}

\definecolor{bordeaux}{HTML}{800020}
\definecolor{explicitblue}{HTML}{007FFF}

\newcommand{\dataset}[1]{\textcolor{black}{\textsc{AbstractEdit}#1}}
\newcommand{\eval}[1]{\textcolor{black}{\textsc{Entity-Rubrics}#1}}

\newcommand{\cmark}{\ding{51}}
\newcommand{\xmark}{\ding{55}}

\newcommand{\mv}[1]{\textcolor{black}{#1}}

\newcommand{\greencheck}{{\color{green!70!black}\cmark}}
\newcommand{\redx}{{\color{red!80!black}\xmark}}

\newcommand{\lowred}{{\color{red!80!black}L}}
\newcommand{\medorange}{{\color{orange!90!black}M}}
\newcommand{\highgreen}{{\color{green!70!black}H}}
\newcommand{\sd}[1]{{\tiny\color{gray}{$\pm$#1}}}

\newcolumntype{L}{>{\raggedright\arraybackslash\hsize=1.1\hsize}X} 
\newcolumntype{E}{>{\raggedright\arraybackslash\hsize=1.3\hsize}X} 
\newcolumntype{C}{>{\centering\arraybackslash}p{22pt}}             
\newcolumntype{B}{>{\raggedright\arraybackslash\hsize=0.7\hsize}X}  
\newcolumntype{Z}{>{\raggedright\arraybackslash\itshape\hsize=1.3\hsize\hspace{6pt}}X}  
\newcolumntype{S}{>{\centering\arraybackslash}p{20pt}}

\DeclareRobustCommand{\failureprofile}[2]{%
  \begin{tikzpicture}[x=25pt, y=6pt, baseline=-0.2ex]
    \draw[black!60, line width=0.5pt] (0,-0.1) -- (0,1.1); 
    \fill[violet!90!black, rounded corners=0.1pt] (0,0) rectangle (-#1,1); 
    \fill[red!80!white, rounded corners=0.1pt] (0,0) rectangle (#2,1); 
    \draw[black!15, line width=0.1pt, dashed] (-0.5,0) -- (-0.5,1); 
    \draw[black!15, line width=0.1pt, dashed] (0.5,0) -- (0.5,1);
  \end{tikzpicture}%
}

\title{\mv{Editor's Choice}: Evaluating Abstract Intent in Image Editing through Atomic Entity Analysis}

\author{%
  Mor Ventura$^{1,2}$\thanks{Work done during an internship at Google Research.} \quad 
  Roy Hirsch$^{2}$ \quad 
  Yonatan Bitton$^{2}$ \\
  \textbf{Regev Cohen}$^{2}$ \quad 
  \textbf{Roi Reichart}$^{1}$ \\
  $^1$Technion -- Israel Institute of Technology \quad $^2$Google Research \\
  \texttt{mor.ventura@campus.technion.ac.il, \{royhirsch, yonatanbitton\}@google.com}
}

\begin{document}

\maketitle



\begin{figure*}[ht]
\centering
\includegraphics[width=0.9\linewidth]{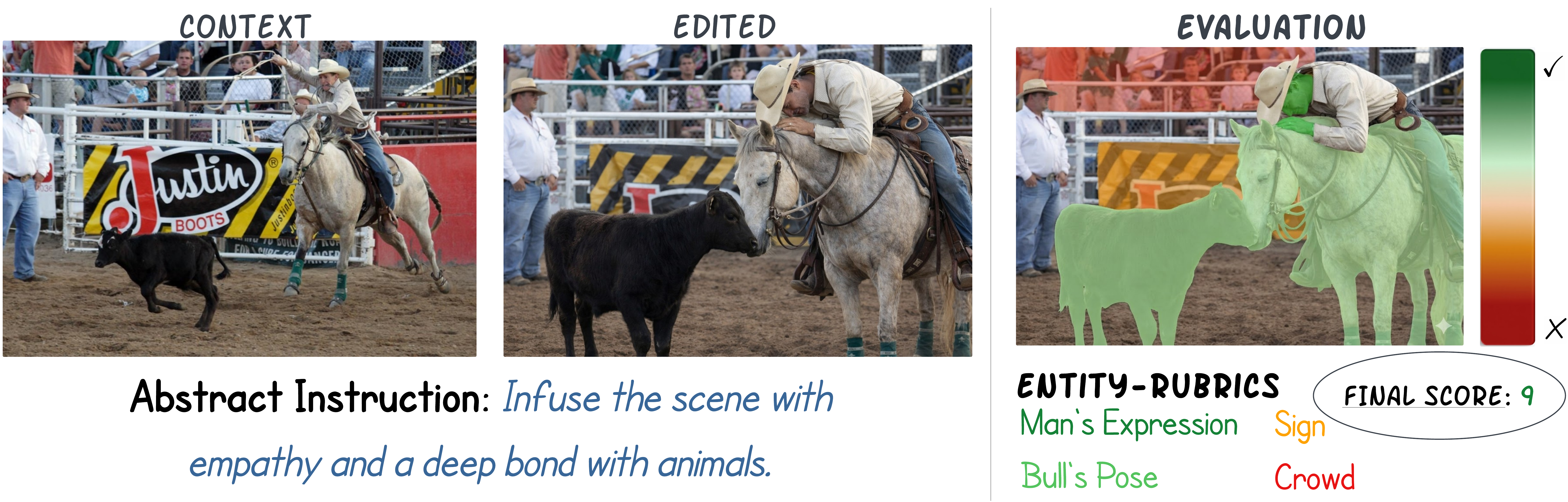}
\caption{\textbf{Evaluating Abstract Image Editing}. Given a context image and an abstract instruction from our \dataset\ test set, our \eval\ framework assesses the edited image by decomposing the scene at a granular level, yielding a per-edit rank that determines the final instruction-following score. As shown, some entity edits lead to better prompt alignment (e.g., the man's expression), while others introduce unnecessary changes that hurt preservation (the crowd).}
\label{fig:main_fig_example}
\end{figure*}


\begin{abstract}
Humans naturally communicate through abstract concepts like ``mood''. However, current image editing benchmarks focus primarily on explicit, literal commands, leaving abstract instructions largely underexplored. In this work, we first formalize the definition and taxonomy of abstract image editing. To measure instruction-following in this challenging domain, we introduce \eval, a framework that breaks down abstract edits into individual, entity-level assessments and achieves strong correlation with human judgment. Alongside this framework, we contribute \dataset, the first benchmark dedicated to abstract image editing across diverse real-world scenes. Evaluating 11 leading models on this dataset reveals a fundamental challenge: standard architectures struggle to balance intent and preservation, commonly defaulting to under-editing or over-editing. Our analysis demonstrates that driving meaningful improvements relies heavily on integrating advanced LLM text encoders and iterative thinking. Looking forward, our entity-based paradigm can generalize beyond assessment to serve as a reward model, enable models to correctly interpret abstract communication, or highlight specific failures in test-time critique loops. Ultimately, we hope this work serves as a stepping stone toward seamless multimodal interaction, closing the gap between rigid machine execution and the natural, open-ended way humans communicate.\footnote{Code and data available via the project page: \url{https://venturamor.github.io/EditorsChoice/}.}


\end{abstract}

\section{Introduction}

Image generation models have made significant progress in following explicit commands \citep{brooks2023instructpix2pix, betker2023improving}, yet their ability to interpret abstract instructions remains a critical bottleneck \citep{liu2022design, fu2023guiding}. In creative applications, humans rarely think in terms of exhaustive, technical specifications; instead, they communicate through high-level concepts, such as altering an image's mood or season \citep{ norman2014some, zamfirescu2023johnny, oppenlaender2024taxonomy}. Natural communication relies on a shared understanding of the world, allowing humans to convey complex ideas without specifying every granular detail \citep{grice1975logic, clark1991grounding}. However, current visual generative models still struggle to achieve this level of \textit{abstract} semantic alignment.

Prior work has recognized this issue but has typically approached it from narrow angles (\S\ref{sec:related}). Studies have predominantly focused on limited implicit instructions \citep{pu2025picabench, zeng2025editworld, yu2025anyedit}, or on physical reasoning \citep{he2025reasoning, zeng2025editworld, yu2025anyedit}, where target outcomes are singular and easily quantifiable, leaving little room for true abstraction. This hesitancy to fully tackle abstract instructions in real-world scenes stems from their inherent difficulty: abstract prompts are highly subjective, allowing many valid interpretations. Therefore, in this work, we establish a framework to tackle the broader challenge of abstract instruction following through a formal problem definition, \mv{and a novel evaluation suite}.

First, we argue that instruction-based \textbf{image editing} provides a uniquely highly practical testbed for bridging this gap compared to a pure text-to-image setup. Generally, image editing is considered more complex than text-to-image generation due to a dual objective: the model must adhere to the text instruction while rigorously preserving the original image context. However, when evaluating abstract instructions, this dual constraint is actually a profound advantage. The requirement to preserve the context image significantly narrows the infinite space of possible interpretations, anchoring the abstract request to a grounded visual baseline and making systematic evaluation tractable.

\begin{figure*}[!ht] 
    \centering
    \includegraphics[width=0.95
    \textwidth]{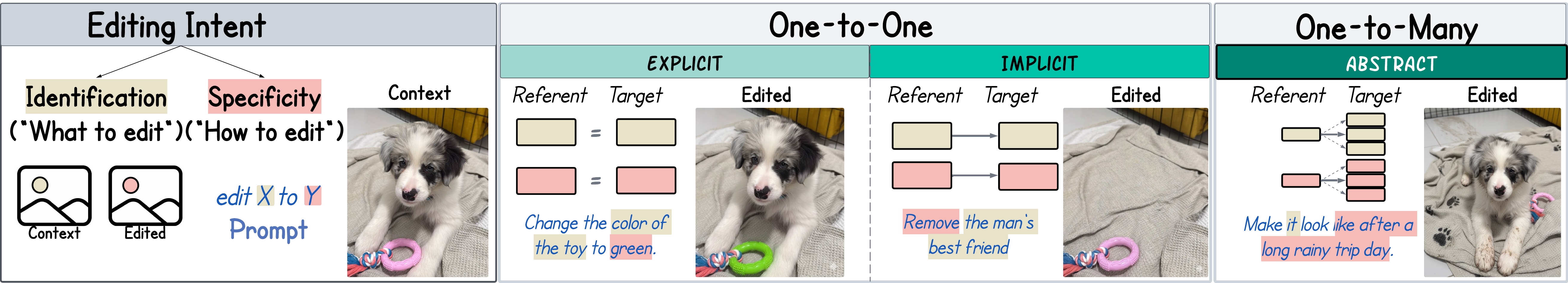}
    \caption{\textbf{Taxonomy of Image Editing Intent}. We formalize intent across two orthogonal axes: Identification (``what to edit'') and Specificity (``how to edit''). Intent is categorized based on the mapping between these axes: Explicit and Implicit editing provide a one-to-one mapping through high consensus on both axes, whereas abstract editing introduces a one-to-many relationship.}
    \label{fig:intent}
\end{figure*}

Next, we define \textit{abstract image editing} (\S\ref{sec:defintion}). It is determined by both the text instruction and the context image, and is measured along two axes spanning human intent: Identification and Specificity. This concept can be conveyed by asking a simple question: \textit{would a majority of human observers identify the exact same target entities to be edited, and would they expect the exact same visual transformations?} If either of these conditions fails, the prompt is considered an abstract edit. 
While explicit instructions, such as ``add mud to the paws'' provide a clear roadmap, the intent in Fig. \ref{fig:intent} to ``make the dog look like it just returned from a long rainy trip'' is entirely abstract; specifying neither target nor transformation. Such abstract intent allows diverse valid edits, such as making the fur wet, moving to a bathtub, or adding an umbrella, requiring the model to autonomously deduce multiple required edits. This definition of abstractness as a subjective, multi-path problem naturally dictates a shift toward \textbf{granularity} and \textbf{precision} in the evaluation.

To achieve this granularity, we draw inspiration from factuality evaluation in NLP. In that domain, complex generated text is assessed by decomposing it into ``atomic facts'' and verifying each one against a source document \citep{gekhman2023trueteacher, min2023factscore, honovich2022true, wei2024long}. We posit that abstract visual transformations can be similarly evaluated by analyzing the discrete modifications applied to specific image entities. Despite the subjectivity of what and how to edit, results must be objectively grounded in the input narrative. Since holistic transformations are more than a ``sum of parts'' we treat entity-level consistency as a necessary, but not sufficient, requirement for overall alignment.

Operationalizing this perspective, we propose \textbf{\eval}, a VLM-based, novel granular automatic evaluation framework. 
\eval\ identifies and treats individual image entities, such as objects and stuff, as the atomic units of assessment (\S\ref{sec:eval}), and provides a highly interpretable rationale, as opposed to existing evaluation methods which at most provide only a single global score and a short explanation \citep{ku2024viescore, yang2025texttt}. Since abstract prompts allow for multiple valid interpretations, target edits are not known a priori. We therefore prioritize precision over recall: rather than checking if a model captured every possible way to interpret a prompt, we verify that the specific edits it chose are logically grounded in the prompt. By that, our approach captures the multiple local edits triggered by abstract instruction, and distinguishes intent-driven transformations from arbitrary alterations that hurt the context image preservation (Fig. \ref{fig:main_fig_example}), yielding a strong correlation with human judgment.

To address the lack of structured resources, we introduce \dataset (\S\ref{sec:dataset}), a diverse dataset of editing scenarios designed specifically to evaluate abstract intent. The test set consists of 470 human-verified abstract samples spanning four core domains, ranging from physical and logical changes to emotional and social shifts. Each sample is a triplet consisting of a context image, an abstract edit instruction, and a corresponding explicit instruction serving as a representative concrete example. To support model development, we further provide a training expansion of 4k samples. \dataset\ is produced via a novel, fully automated VLM-powered pipeline designed to ensure the systematic generation of contextually relevant, diverse prompts grounded in real-world scenes.

Leveraging the \dataset\ and the \eval\ framework, we benchmark 11 leading models (\S\ref{sec:results}) and reveal a stark performance gap. The models exhibit divergent behaviors: closed-source models thrive on abstract intent but often over-edit and sacrifice original context, whereas open-source models exhibit variable performance, relying heavily on explicit instructions and frequently under-editing by missing latent requirements. We find that advanced LLM text encoders and iterative ``thinking'' mechanisms are key to boosting abstract instruction following. \mv{Ultimately, our analysis confirms that abstract prompts unlock model creativity and a quantifiable manifold of visual diversity.}

In summary, our work provides a comprehensive framework for navigating the complex landscape of abstract image editing. Our core contributions are as follows: (1) a formal definition of \textit{abstract} image editing; (2) an \eval\ evaluation framework; (3) The \dataset\ benchmark and its automated generation pipeline; and (4) an in-depth analysis of open-source versus closed-source editing capabilities. Ultimately, this framework moves the field beyond rigid instruction-following toward true alignment with complex human intent, providing the roadmap for models to autonomously master the one-to-many nature of real-world editing.

\section{Related Work}
\label{sec:related}

\paragraph{Instruction-Guided Image Editing Datasets.}
Early image editing models relied on local masks \citep{avrahami2022blended, joseph2024iterative, yuan2024flexedit}, but frameworks like InstructPix2Pix \citep{brooks2023instructpix2pix} and MagicBrush \citep{zhang2023magicbrush} shifted the paradigm toward instruction-guided editing to facilitate natural human communication \citep{clark1991grounding}. However, these datasets primarily focus on simple, explicit commands \citep{fu2023guiding, ma2024i2ebench, qian2025gie}. To push the boundaries of edit complexity, subsequent datasets incorporated multi-step instructions \citep{yang2025texttt}, implicit prompts requiring specific domain knowledge \citep{huang2024smartedit, jia2025compbench}, or hypothetical scenarios within the physical domain \citep{he2025reasoning, zeng2025editworld, yu2025anyedit} (see \S\ref{sec:dataset} and Table~\ref{tab:benchmarks_x_small} for comparison). Yet, these benchmarks evaluate rigid, one-to-one visual mappings rather than true abstract instructions, which require models to translate vague intents into specific edits that align with a user's mental model \citep{norman2014some}. To bridge this gap, we formalize abstract image editing along the axes of Identification and Specificity, introducing the \dataset, a benchmark specifically designed to evaluate abstract image editing.

\begin{table}[h]
\centering
\scriptsize
\setlength{\tabcolsep}{1.8pt} 
\caption{\textbf{Comparison of benchmark subsets for complex image editing}. Size refers to the test set size; Dom indicates the number of covered domains categorized as {\color{red}Low} ($1-2$), {\color{orange}Medium} ($3-4$), and {\color{green}High} ($\geq 5$); Ctx Img counts the unique context images; Nat. denotes if prompts are natural ($\checkmark$) or templated ($\xmark$); and Global/Local specifies the edit \mv{spatial} scope. eDoF (editing Degree of Freedom) describes the intent mapping: $\text{one-to-one}$ ($1:1$) or $\text{one-to-many}$ ($1:N$) abstract.}
\label{tab:benchmarks_x_small}
\begin{tabularx}{\columnwidth}{@{} l >{\raggedright\arraybackslash}p{1.3cm} c c c c c c >{\raggedright\arraybackslash\itshape}X @{}}
\toprule
\textbf{Bench.} & \textbf{Subset} & \textbf{Size} & \textbf{Dom} & \textbf{\makecell{Ctx.\\Img}} & \textbf{\makecell{Nat.\\}} & \textbf{\makecell{Glob/\\Loc}} & \textbf{eDoF} & \textbf{\makecell[l]{Example\\Instruction}} \\
\midrule
\textbf{CompBench} \citep{jia2025compbench} & Implicit & 100 & \lowred & 100 & \redx & L & 1:1 & Remove the farthest tiger from the water \\
\addlinespace[3pt]
\textbf{SmartEdit} \citep{huang2024smartedit} & Implicit & 60 & \medorange & 30 & \redx & L & 1:1 & Remove the object that can be used to have meals \\
\addlinespace[3pt]
\textbf{EditWorld} \citep{zeng2025editworld} & Logic & 60 & \lowred & 60 & \redx & - & 1:1 & What happens if a hole appears in the balloon? \\
\addlinespace[3pt]
\textbf{AnyEdit} \citep{yu2025anyedit} & Implicit & 10k & \lowred & 10k & \redx & - & 1:1 & What would happen if the man can't catch the wave? \\
\midrule
\textbf{EMU-Edit} \citep{sheynin2024emu} & Global & 440 & \medorange & 219 & \greencheck & G & 1:N & Change the scene to night time \\
\addlinespace[3pt]
\textbf{Kontext} \citep{labs2025flux} & Global & 262 & \medorange & 88 & \greencheck & G & 1:N & Make this image real \\
\midrule
\textbf{\dataset} & \textbf{Full Abs.} & \textbf{470} & \highgreen & 257 & \greencheck & G,L & 1:N & Make the dog look like after a long rainy trip day \\
\bottomrule
\end{tabularx}
\end{table}

\paragraph{Evaluating Instruction Alignment.}
Evaluating image editing requires balancing instruction following with context preservation \citep{brooks2023instructpix2pix}. However, traditional encoding-based metrics like CLIP \citep{radford2021learning} or DINO \citep{simeoni2025dinov3} struggle with fine-grained semantics \citep{thrush2022winoground,liang2022mind, hu2023tifa, zhang2025re}, often rewarding aggressive over-editing over logical though subtle changes. \mv{To address this, VQA metrics \citep{lin2024evaluating, hu2023tifa} attempt localized checks, but their reliance on rigid proxy questions is conceptually misaligned with abstract instructions and inherently penalizes valid visual diversity.} Alternatively, recent VLM judges \citep{ku2024viescore, yang2025texttt} capture broader semantics but typically output holistic scores that lack interpretable, localized explanations. Crucially, none of these approaches address how to unpack an abstract instruction to systematically verify its execution. To bridge this gap, we introduce \eval, a framework that measures alignment to high-level intents through atomic, entity-level verification.

\paragraph{Approaches to Complex Abstract Instructions.}
Recent methods address complex instructions by using LLMs to decompose abstract prompts into explicit sub-tasks \citep{ji2025instruction, yu2025anyedit, iakovleva2024specify, zeng2025editworld}. Building on this, some frameworks execute these simple steps sequentially \citep{yeh2025beyond, yang2025texttt}, or introduce agentic loops where a VLM provides iterative visual feedback \citep{li2025editthinker, wang2024genartist, fang2025got, yao2026photoagent}. While effective, these computationally expensive strategies restrict visual diversity. Shifting the semantic burden to the text encoder forces rigid interpretations and leaves the diffusion model's innate mapping abilities unoptimized. To solve this, our pipeline provides scalable data, including a 4k-sample training set, enabling future work to directly optimize diffusion models for abstract edits. Complementing this, \eval\ offers entity-level diagnostics to target specific failures without exhaustive trial-and-error.

\section{Abstract Image Editing: Taxonomy and Definition}
\label{sec:defintion}

This work focuses on the underexplored complexity of \textit{semantic abstractness in human intent}, as opposed to previous studies that characterize complexity through instruction count \citep{yang2025texttt, wang2025gpt} or knowledge-based implicit references \citep{huang2024smartedit, qian2025pico}.
By shifting the focus of complex image editing toward interpretation, abstract editing measures a model's ability to translate underspecified intent into concrete visual modifications. This process quantifies the ``generative guesswork'' required to bridge the conceptual gap between the text instruction ($p$) and context image ($\mathcal{I}_c$) and a satisfying result. 

To formalize abstract image editing, we define two orthogonal axes spanning the human intent space: \textit{identification} and \textit{specificity}. \textbf{Identification} pinpoints the exact semantic or visual entities targeted by the edit, whether modifying existing elements or inserting new ones (the ``what''). \textbf{Specificity} determines the precise nature and extent of the visual transformations applied to those elements (the ``how''). \mv{Conceptually, the generative guesswork required to resolve these axes can be viewed as an \textbf{Editing Degree of Freedom ($eDoF$)}. Operating as a continuous spectrum, $eDoF$ represents the autonomy a model must exercise to interpret an instruction.} Theoretically, we could define $eDoF$ based on the size of the interpretation set $\mathcal{K}$, which represents all visually distinct edits that satisfy the text prompt $p$ given the context image $\mathcal{I}_c$: $eDoF \propto |\mathcal{K}(p | \mathcal{I}_c)|$.

\begin{figure*}[!ht] 
    \centering
    \includegraphics[width=1.0
    \textwidth]{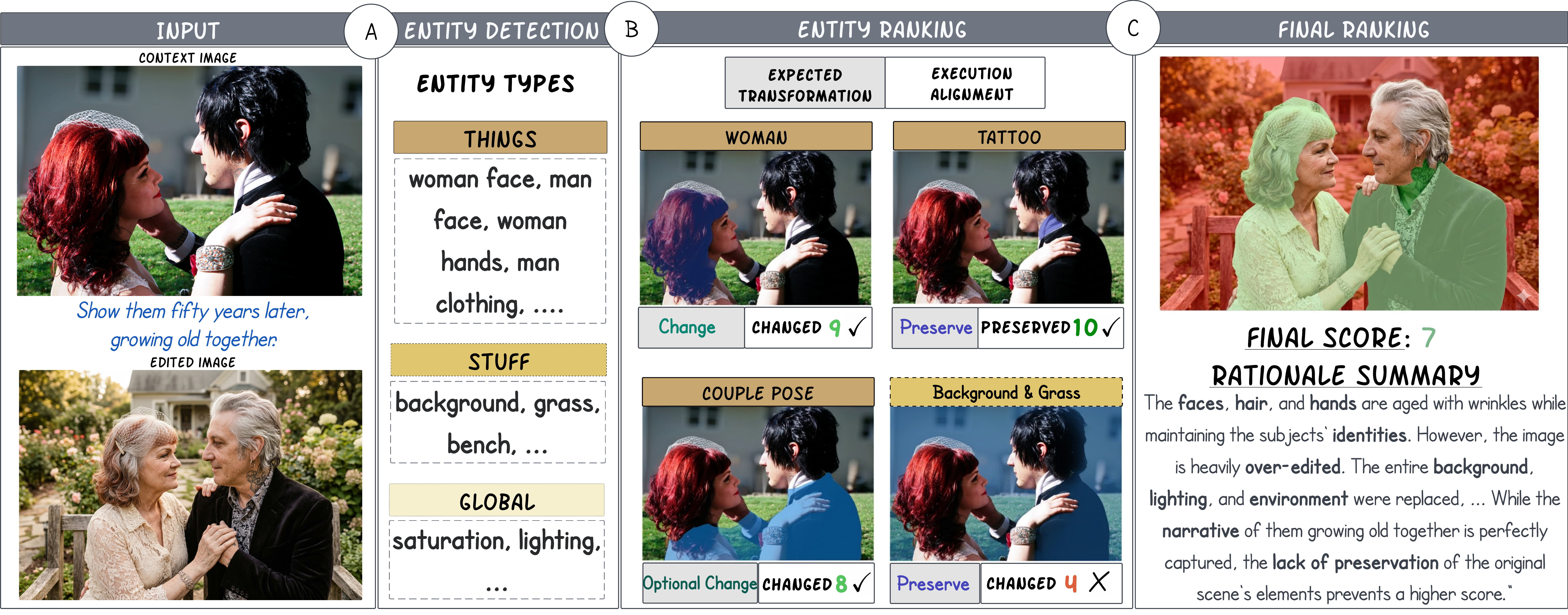}
    \caption{\textbf{Overview of the three-stage VLM-based \eval\ evaluation framework}. (A) Entity Detection identifies relevant entities. (B) Entity Ranking assigns an expected transformation to each entity (\textit{Change}, \textit{Optional}, \textit{Preserve}) and measures its execution alignment in the edited image. (C) Final Scoring aggregates these into a global rank and rationale. Results are visualized directly on-image via a red (incorrect) to green (correct) scale.}
    \label{fig:eval_scheme}
\end{figure*}

The magnitude of this interpretation set dictates the overall mapping of the edit (Fig. \ref{fig:intent}). When an instruction provides exact clarity along both axes, it yields a strict \textit{one-to-one} mapping where $|\mathcal{K}| \approx 1$ and the lowest $eDoF$. We divide these strictly mapped cases into two sub-types (Fig. \ref{fig:intent}). \textbf{Explicit} ($Exp$) edits are completely direct, mapping the literal instruction perfectly to the outcome (e.g., ``turn the toy green''\footnote{We acknowledge minor variations in hue but treat these as negligible for the purpose of interpretation.}). \textbf{Implicit} ($Imp$) edits also maintain a one-to-one mapping but require additional computational capabilities, such as specific domain knowledge (understanding ``man’s best friend'' is a dog) or robust visual detection (e.g. detecting the colors when instructed to ``swap the sofa's color with the toy's color''). Conversely, when ambiguity exists along the identification or specificity axes, the interpretation set expands into a \textit{one-to-many} mapping ($|\mathcal{K}| \gg 1$). These underspecified instructions demand a high $eDoF$ and are categorized as \textbf{abstract} ($Abs$) edits, as they remain open to a vast array of valid visual interpretations.

In practice, applying these axes is challenging because both offer endless possibilities. For identification, a target can be anything from a simple ``dog'' to a complex interaction like ``the dog's gaze on the toy'' \citep{kress2020reading}. For specificity, any chosen target can then be transformed into a multitude of valid ways. To manage this complexity in our evaluation (\S\ref{sec:eval}), we adopt a precision-over-recall approach. Rather than attempting to map the entire infinite space of what and how a model could have edited, we focus on verifying that the specific edits the model actually performed are logically grounded in the instruction. Further formalization is detailed in Appendix \S\ref{app:theory_definition}.

\section{\eval: Automatic Evaluation for Abstract Image Editing}
\label{sec:eval}

Evaluating abstract intent in image editing is uniquely challenging because a single instruction often triggers changes across multiple entities in various ways. As shown in Fig. \ref{fig:eval_scheme}, ``making a couple look older'' involves more than just graying a woman's hair, it entails adding facial wrinkles, updating outfits to reflect age, and potentially adjusting gaze to suggest a lifelong partnership. Effective evaluation must capture these nuances, distinguishing between essential transformations and optional creative choices for a weighted performance assessment. Furthermore, because these instructions are complex, models often succeed at some edits while failing at others. Current methods, whether based on image-text join-encoding as CLIP \citep{radford2021learning} or zero-shot VLM judgments \citep{ku2024viescore, yang2025texttt}, lack the precision to account for this multi-entity coordination and fail to provide the granular feedback necessary for such a high-dimensional task. To address this, we introduce \textbf{\eval}, a novel automatic evaluation framework centered on the granular inspection of entities, which inherently supports fast-to-interpret visual reporting.

Our framework is inspired by factuality research in the NLP domain, \mv{which frames semantic alignment as Natural Language Inference (NLI) \citep{bowman2015large, maynez2020faithfulness, kryscinski-etal-2020-evaluating}}, and verifies information by decomposing complex claims into ``atomic facts'' for individual validation \citep{gekhman2023trueteacher, min2023factscore, honovich2022true, wei2024long, chern2023factool}. As illustrated in Fig. \ref{fig:eval_scheme}, we adapt this concept to the visual domain by treating image entities as "atomic units". We decompose the evaluation into discrete logical steps, leveraging the VLM’s reasoning to perform a per-entity alignment and necessity assessment, given the context image, the candidate edited image and the instruction. This granular focus ensures that the final evaluation score is grounded in specific visual evidence, verifying both the correctness of modifications and the necessity of preservation, rather than relying on a global, surface-level explanation.

As illustrated in Fig. \ref{fig:eval_scheme}, the \eval\ Evaluation Framework consists of three main stages: entity detection, entity ranking, and final scoring. The first stage, \textbf{(A) Entity Detection}, focuses on identifying all entities\mv{, including global attributes,} within both the context and the edited images. To ensure comprehensive coverage, the VLM is prompted to partition these entities into three semantic groups: \textit{things} (distinct objects such as ``woman's face''), \textit{stuff} (amorphous background \citep{caesar2018coco}; environmental elements like ``grass''), and \textit{global} (global attributes as lighting). By categorizing entities this way, the framework captures intent-driven changes across every layer of the image.

Following this, the \textbf{(B) Entity Ranking} stage serves as the core of the framework. It addresses the tension between the expected edit and the actual candidate edited image produced by an image editing model. This stage includes two phases. First, the VLM determines the Expected Transformation for each entity. It iterates over all detected entities from the context image to assess their expected state based solely on the abstract instruction. This phase represents the identification axis, acting as a theoretical grounding step to determine what requires modification. Specifically, it assigns an expected state of $Change$, $Optional$ $Change$, or $Preserve$ before any visual comparison occurs, using only the context image and the text prompt as input.

Then, the evaluation moves to examining the edited image to analyze the modifications present in the candidate result. In this phase, Execution Alignment, the VLM first assesses whether a change occurred for each entity. If a change is detected, it is followed by a qualitative assessment of how the entity was transformed or preserved, producing a concise description of the modification (not shown in the figure). 
These two phases culminate in a per-entity rank, where the VLM calculates a score based on how well the observed changes align with the expected transformations and whether they increase overall alignment with the instruction. To ensure holistic quality, the VLM is also prompted to provide a global failure profile that audits the image for missing changes or over-editing artifacts not captured at the individual entity level (not shown in the figure). 

Finally, \textbf{(C) Final Scoring} and Rationale aggregates all previous findings into a comprehensive 1–10 assessment. This stage prioritizes three main criteria: how well the model followed the instruction, whether it met the expected level of change, and the overall coherence of the final image. By combining the data from all entity categories with the global examination, the framework generates a score and a written rationale. A score of 10 reflects perfect alignment where all required changes are executed flawlessly while preserving the essential context of the context image. Lower scores are assigned based on the severity of issues such as missed requirements, disjointed narratives, or over-editing artifacts. This synthesis ensures the final evaluation is a grounded reflection of performance across every image layer. The full prompt architecture is detailed in Appendix \S\ref{app:eval}.


To validate our automatic evaluation, we conducted an Amazon Mechanical Turk (AMT) study collecting 1,080 responses across 60 stratified \dataset\ samples (see Appendix \S\ref{app:human_evaluation}). Three annotators per task evaluated edits from the six top-performing models, assessing instruction following, content preservation, image quality, and granular per-entity alignment. Aggregating responses via mean vote, the study achieved moderate Inter-Annotator Agreement (quadratic weighted Fleiss' $\kappa$ of \mv{$0.47$}). Crucially, \eval\ achieves a Spearman's $\rho$ of \mv{$0.66$} with human judgments, confirming it as a high-fidelity proxy for instruction following while outperforming alternative metrics like VIE (\mv{$0.54$}) and image-text CLIP (\mv{$0.41$}).

\begin{figure*}[!ht] 
    \centering
    \includegraphics[width=0.9
    \textwidth]{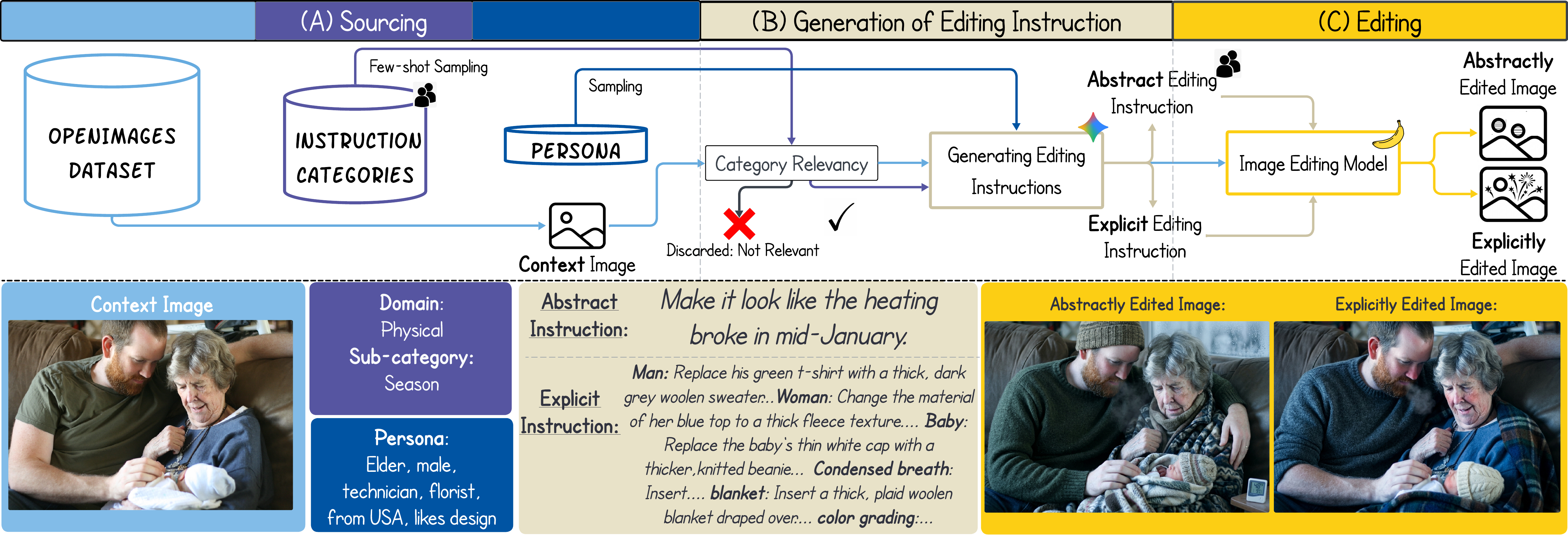}
    \caption{\textbf{The \dataset\ Automatic Curation Pipeline}. (A) Sourcing: Context images (Azure) from OpenImages are compiled alongside manual category examples (Purple) and diverse personas (Blue). (B) Instruction Generation: After filtering the images for category relevancy, an LLM, prompted by few-shot examples and a random persona, generates paired abstract and explicit instructions (Beige). (C) Editing: Both instructions are applied to the context image to produce the final edited pairs (Yellow). Bottom: A generated \dataset\ example.}
    \label{fig:dataset_scheme}
\end{figure*}

\section{The \dataset\ Dataset: Abstract Image Editing}
\label{sec:dataset}

Evaluating abstract image editing requires benchmarks that capture the inherent complexities of one-to-many mappings. Our analysis of the broader landscape of complex editing (Table \ref{tab:benchmarks_x_small}) reveals that existing commonly adopted datasets are often restricted by template-based language and narrow domain diversity. Most critically, these frameworks suffer from a diagnostic gap: they treat the editing process as a black box, failing to decouple high-level semantic inference from low-level generative execution. Without this distinction, it cannot be determined if a failure is due to misinterpreting the concept or an inability to perform the edit.

To address these gaps, we present \dataset, a dedicated benchmark for evaluating abstract image editing across diverse, real-world scenes. It is structured to support direct comparisons between paired abstract and explicit instructions. By providing both prompt types for each context image, the explicit edits serve as a direct baseline to isolate and quantify a model's abstract interpreting capabilities. The \dataset\ test set contains \mv{470} human-verified samples (via manual author review) to ensure they strictly adhere to our definition of abstract editing, as well as logically grounded. \mv{Additionally, we provide a training set of \mv{4k} samples to support future model development}.


\dataset\ spans 12 categories organized into four domains: Physical, Logical, Emotional, and Social. Each domain demands a varying level of generative autonomy, testing the model's capacity to resolve the inherent ambiguity of abstract intent. For instance, the Emotional domain demands high-level $eDoF$; Given a prompt to ``infuse empathy and a deep bond with animals'' (Fig. \ref{fig:main_fig_example}), the model should infer that the matador's aggressive posture should be replaced with a nurturing stance. Unlike existing benchmarks that focus on object detection or implicit references, \dataset\ targets abstract instructions that necessitate coordinated local and global edits across multiple entities. For additional details and statistics, see Appendix \S\ref{app:dataset_curation}.

 As illustrated in Fig. \ref{fig:dataset_scheme}, \dataset\ is generated via an automatic pipeline designed to capture a broad spectrum of authentic human intents through diverse and natural instructions. Our pipeline operates in three phases: (A) Sourcing, (B) Instruction Generation, and (C) Image Editing. The \textbf{Sourcing (A}) phase establishes the creative constraints of the dataset. We begin by selecting \mv{1,300} natural images from the Open Images v7 opensource dataset \citep{kuznetsova2020open}, prioritizing complex scenes with multiple interacting entities to ensure the model has a rich ``canvas'' to edit. To guide the abstraction, we manually define a taxonomy of the categories (e.g., \textit{Physical: Seasonal}), including instruction examples for each. We then utilize synthetic personas, defined by traits like age and expertise, to avoid predictable templates. Randomizing these traits spans a broad linguistic spectrum; a practical persona might dictate to ``make the room look freezing to test the insulation'', whereas a poetic persona might say ``infuse the room with the biting chill of a mid-January morning''.

\begin{figure}[htbp]
    \centering
        \captionof{table}{\textbf{Abstract Instruction Following Performance}. \textbf{Left Table}: Open-Source, OS with thinking modes, and closed models ranked by Abstract \eval\ scores and human annotations. \textbf{Failure Profile:} \textcolor{violet!90!black}{(Under-Editing)} $\leftarrow | \rightarrow$ \textcolor{red!80!white}{Over-editing}. Domains: \textbf{\textcolor{magenta}{Emotional}}, \textbf{\textcolor{blue}{Logical}}, \textbf{\textcolor{green!70!black}{Physical}}, and \textbf{\textcolor{orange!90!black}{Social}}. \textbf{Right Fig. \ref{fig:explicit_abstract}}: Prompt Type Comparison: Explicit (blue) and Abs. (striped bordeaux).}
        \vspace{-15pt}
        \label{tab:eval_complete_results}
    \begin{minipage}{0.75\textwidth}
        
        \scriptsize
        \setlength{\tabcolsep}{1.2pt} 
        \renewcommand{\arraystretch}{1.3} 
        \begin{tabular}{l l | c c c | c c c c}
            \toprule
            \textbf{Cat.} & \textbf{Model} & \textbf{Hum} $\uparrow$ & \textbf{Abs} $\uparrow$ & \textbf{F. Prof.} & 
            \textbf{\textcolor{magenta}{\makecell{Emo-\\tional}}} & 
            \textbf{\textcolor{blue}{\makecell{Logi-\\cal}}} & 
            \textbf{\textcolor{green!70!black}{\makecell{Physi-\\cal}}} & 
            \textbf{\textcolor{orange!90!black}{\makecell{So-\\cial}}} \\
            \midrule
            \multirow{4}{*}{\rotatebox{90}{\makecell{Open\\Source}}} 
                & Qwen-Image-Edit            & -    & \textbf{7.48} \sd{2.86} & {\failureprofile{0.15}{0.22}} & 8.00 \sd{2.89} & 6.89 \sd{3.02} & \textbf{7.43} \sd{2.92} & \textbf{8.14} \sd{2.41} \\
                & FLUX.2 [dev]               & 8.50 & 7.26 \sd{2.83} & {\failureprofile{0.13}{0.32}} & \textbf{8.29} \sd{2.04} & \textbf{7.34} \sd{2.89} & 6.85 \sd{3.24} & 7.16 \sd{2.57} \\
                & HiDream-E1                 & -    & 5.38 \sd{3.15} & {\failureprofile{0.33}{0.49}} & 7.91 \sd{2.12} & 4.18 \sd{3.05} & 4.97 \sd{3.23} & 6.56 \sd{2.63} \\
                & FLUX.1-Kontext [dev]       & 7.69 & 5.10 \sd{3.46} & {\failureprofile{0.48}{0.20}} & 6.17 \sd{3.48} & 4.34 \sd{3.44} & 5.21 \sd{3.62} & 5.72 \sd{3.16} \\
            \midrule
            \multirow{4}{*}{\rotatebox{90}{\makecell{OS w/\\Thinking}}} 
                & Step1X-Think-Reflect       & 7.97 & \textbf{6.90} \sd{3.20} & {\failureprofile{0.35}{0.05}} & \textbf{7.63} \sd{3.06} & \textbf{7.39} \sd{3.22} & 5.39 \sd{3.64} & \textbf{7.14} \sd{2.50} \\
                & Step1X                     & -    & 6.55 \sd{3.41} & {\failureprofile{0.39}{0.06}} & 7.61 \sd{3.06} & 6.90 \sd{3.38} & 5.36 \sd{3.83} & 6.64 \sd{2.97} \\
                & Bagel-Think                & -    & 5.80 \sd{3.50} & {\failureprofile{0.39}{0.21}} & 5.85 \sd{3.43} & 4.84 \sd{3.67} & \textbf{5.97} \sd{3.49} & 6.92 \sd{2.91} \\
                & Bagel                      & 6.61 & 4.45 \sd{3.42} & {\failureprofile{0.53}{0.21}} & 4.89 \sd{3.45} & 3.31 \sd{3.21} & 4.90 \sd{3.37} & 5.53 \sd{3.29} \\
            \midrule
            \multirow{5}{*}{\rotatebox{90}{\makecell{Closed\\Source}}} 
                & Gemini 3.1 Flash   & 9.66 & \textbf{9.52} \sd{1.37} & {\failureprofile{0.02}{0.06}} & \textbf{9.62} \sd{1.33} & 9.47 \sd{1.34} & \textbf{9.43} \sd{1.64} & \textbf{9.62} \sd{1.19} \\
                & GPT-Image                  & 9.67 & 9.34 \sd{1.37} & {\failureprofile{0.01}{0.11}} & 9.49 \sd{1.43} & 9.19 \sd{1.42} & 9.36 \sd{1.46} & 9.49 \sd{1.20} \\
                & Gemini 3 Pro     & -    & 9.27 \sd{1.78} & {\failureprofile{0.04}{0.07}} & 9.50 \sd{1.40} & \textbf{9.59} \sd{1.13} & 8.46 \sd{2.80} & 9.35 \sd{1.42} \\
                & Seed-Dream                 & -    & 9.21 \sd{1.76} & {\failureprofile{0.04}{0.08}} & 9.43 \sd{1.44} & 9.21 \sd{1.71} & 8.88 \sd{2.33} & 9.39 \sd{1.38} \\
                & Gemini 2.5 Flash Image     & -    & 8.67 \sd{2.40} & {\failureprofile{0.12}{0.07}} & 9.23 \sd{1.70} & 9.09 \sd{2.04} & 7.57 \sd{3.26} & 8.73 \sd{2.05} \\
            \bottomrule
        \end{tabular}
    \end{minipage}
    \hfill
    \begin{minipage}{0.2\textwidth}
        \centering
        \raisebox{-70mm}{\includegraphics[width=\textwidth]{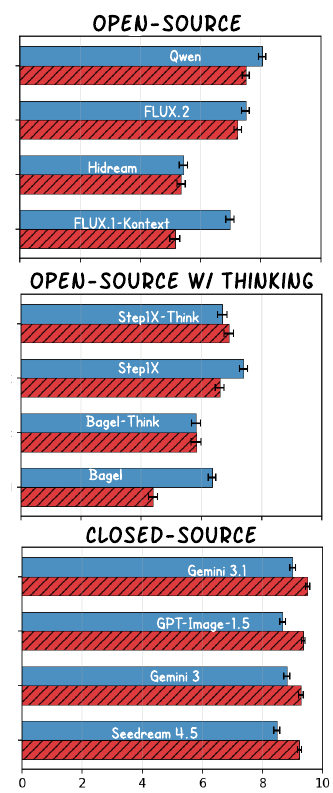}}
        \refstepcounter{figure}
        \label{fig:explicit_abstract}    
    \end{minipage}

    \includegraphics[width=1.0\textwidth]{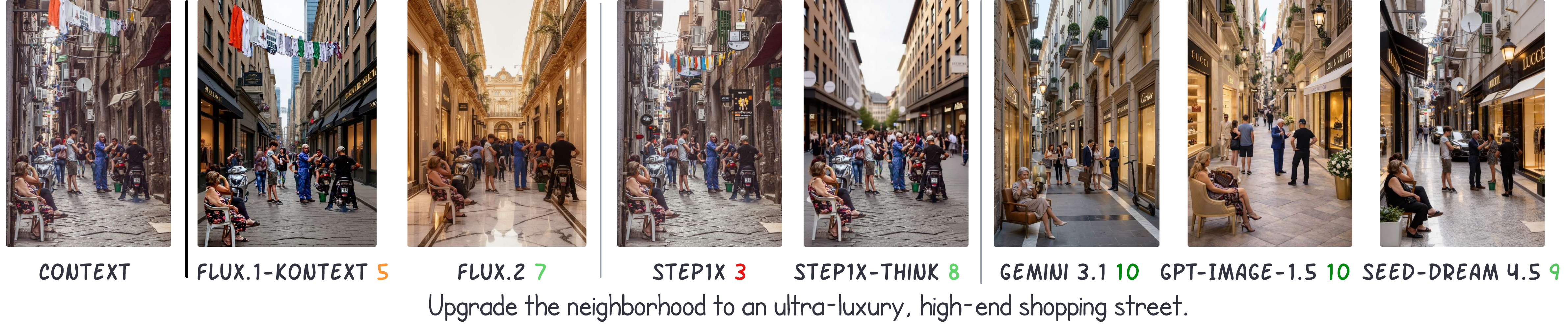}
    \captionof{figure}{Qualitative example edited by different models, representative from each model-category.}
    \label{fig:qualitative_eval_results}
\end{figure}

\textbf{Instruction Generation (B)} then translates these images, instruction examples and personas into actionable tasks. For every context image, we first perform a semantic relevance check to ensure the chosen category is grounded in the context image, for instance, we would not apply a ``Physical: Pose'' edit to an image of an empty desert. For valid pairings, we use a few-shot prompting approach where an LLM (Gemini 2.5 Pro \citep{comanici2025gemini}), adopting a specific random persona, generates two types of instructions: an abstract intruction representing underspecified human intent (e.g., ``The heating just broke''), and explicit instructions detailing the concrete visual modifications that ``solve'' the abstraction (e.g., ``Add frost to window edges and place a wool blanket on the sofa''). We then manually verify the abstractness of each sample \mv{of the test set}. In the \textbf{Editing phase (C)}, we execute both prompts to produce the final image pairs.

\section{Results \& Experimental Setup}
\label{sec:results}

We evaluate \mv{11} state-of-the-art image editing models selected for their diverse parameter scales and architectures. This comprises 8 open-source (OS) models, two of which include thinking variants, and 5 closed-source models (see Appendix \S\ref{app:exp_setup}). Table \ref{tab:eval_complete_results} presents the primary \eval\ scores on the \dataset\ test set and includes human evaluations for the top six performers, with Fig.~\ref{fig:qualitative_eval_results} providing a qualitative example of these abstract edits. Fig. \ref{fig:explicit_abstract} further illustrates the performance gap between abstract and explicit prompts. Supplementary preservation and instruction-following metrics, and complementary results are deferred to the Appendix (\S\ref{app:exp_setup} and \S\ref{app:results}) as well as qualitative results (\S\ref{app:qual_res}). Our analysis reveals several key trends:

\paragraph{Prompting Trade-offs in Open vs. Closed Models.}
Performance gap exists between closed-source and open-source models, with closed-source models achieving a high mean score of $9.2$ compared to top open-source performers like Qwen-Image-Edit ($7.48$). As presented in Fig. \ref{fig:explicit_abstract}, this divide is driven by a fundamental difference in instruction processing. Closed-source models thrive on abstract instructions, improving by up to $9\%$ with tighter score distributions when presented with high-level intent. In contrast, open-source models generally need explicit roadmaps to maintain alignment. \mv{However, this reliance on explicit instructions creates a severe trade-off for open-source architectures. Lengthy text often overwhelms these models, leading to excessive over-editing as text tokens dominate the original image context. Although switching to abstract prompts universally reduces this over-editing across all models (avg. $-13.3\%$), it merely shifts the failure mode. Unable to decipher high-level intents, open-source models fail to execute the edit entirely, trading over-editing for severe under-editing.} Thus, while open-source models require explicitness to function, abstraction serves as a necessary regularizer for preserving the original image. Domain-wise (Table \ref{tab:eval_complete_results}), we find that open-source models handle better with Emotional and Social domains (mean \mv{$6.88$}) rather than the Physical and Logical domains (\mv{$5.7$}). We hypothesize that this gap is a function of feature alignment. Emotional and Social edits often map to strongly-encoded semantic patterns (e.g., color shifts or facial expressions) that are robustly represented in the generated image latent space. In contrast, Physical and Logical domains require ``multi-hop'' compositional reasoning to resolve novel spatial dependencies and semantic rules that cannot be solved by simple pattern retrieval.

\paragraph{Failure Analysis.}
Open-source and closed-source models exhibit opposite failure modes. Open-source models tend toward under-editing, failing to capture latent requirements in abstract prompts (Table \ref{tab:eval_complete_results}; purple bar of Failure Profile); \mv{this results in inflated preservation metrics (Appendix Table \ref{tab:preservation_results})}. Conversely, closed-source models are prone to over-editing (red bar), discarding the original context to satisfy underspecified instructions.

\begin{wrapfigure}{r}{0.5\textwidth} 
    \centering
    \includegraphics[width=0.48\textwidth]{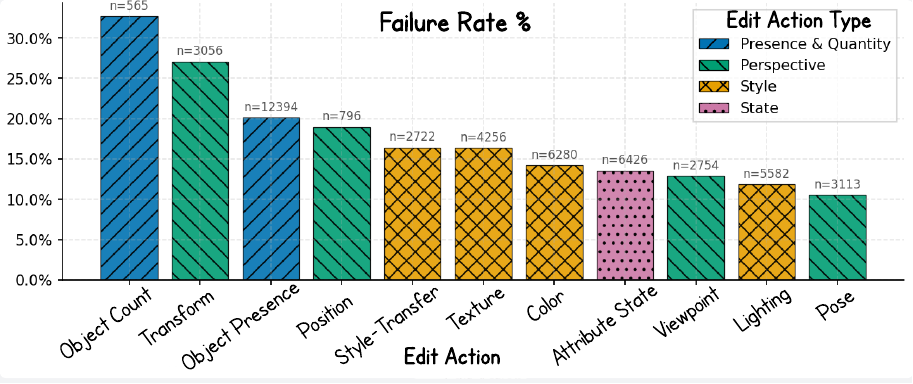}
    \caption{\textbf{Failure Rate per Entity Edit Action}. Normalized by the number of occurrences of each edit type across all models.}
    \label{fig:failure_modes_action_edit}
    \vspace{-10pt}
\end{wrapfigure}
Action-specific analysis (Fig. \ref{fig:failure_modes_action_edit}), \mv{based on the entity-ranking stage of our evaluation (\S\ref{sec:eval})}, reveals granular bottlenecks. The highest failure rates ($>30\%$) involve Object Counts, suggesting that numerical grounding remains the primary hurdle in image editing \citep{dahary2025decisive, binyamin2025make, tewel2024add}. This is followed by Perspective (scale, rotation, position) and Object Presence, which often yield missing or malformed entities. In contrast, models reliably handle ``surface'' edits like Style, Texture, and Attribute State. This confirms that the primary bottleneck in abstract instruction following is structural reasoning rather than aesthetic adjustment.

\paragraph{Thinking and Text Encoders as Drivers of Performance in Abstract Instruction Following.} Two key drivers improve instruction following for abstract prompts: advanced LLM backbones and iterative thinking processes. Among open-source models, the choice of text encoder is a primary performance factor. For instance, FLUX.2 [dev] utilizes a Mistral and T5 configuration ($7.26$) to outperform the CLIP-based Flux-Kontext ($5.1$), while Qwen-Image-Edit leverages a Qwen-VL-2.5 backbone. These are the highest-scoring open-source models. Beyond architecture, iterative thinking mechanisms significantly stabilize performance on abstract tasks. Step1X with its thinking mode improves its \eval\ score by $5.3\%$ over the base model, achieving a high human instruction-following score of $7.97$. Similarly, Bagel-Think demonstrates a $30.3\%$ increase in its score compared to the standard Bagel model. The efficacy of these methods suggests that the observed performance gap between closed-source and open-source models is largely driven by the adoption of similar inference-time thinking processes \citep{google_gemini_image_2026}. Furthermore, these thinking models offer a mixed benefit: they boost alignment on abstract instructions ($+1.4$ points for Bagel) but suffer a 'precision tax' on explicit prompts (Fig. \ref{fig:explicit_abstract}), where overthinking hurts direct execution.

\begin{wrapfigure}{r}{0.45\textwidth}
    \vspace{-20pt}
    \centering
    \includegraphics[width=0.44\textwidth]{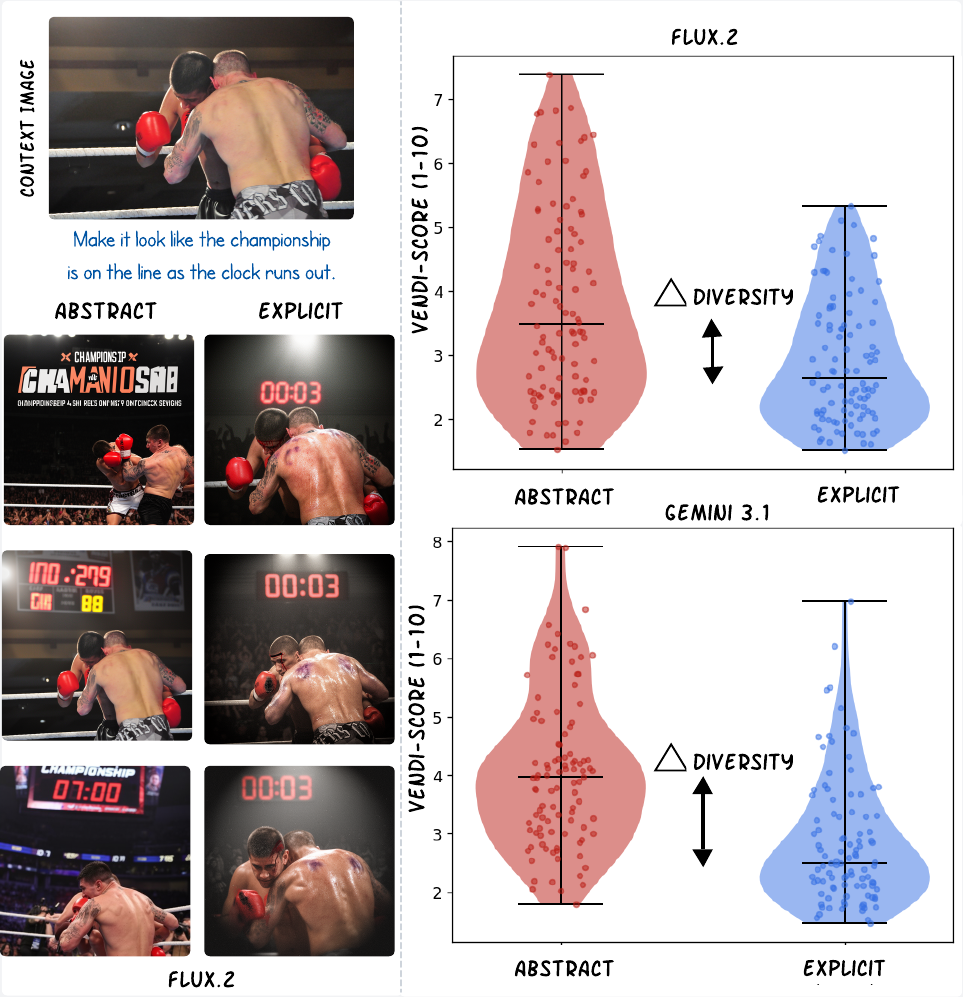}
    \caption{\textbf{Diversity Analysis}. Vendi score distributions for abstract vs. explicit instructions (right) and qualitative examples (left).}
    \label{fig:vendi}
    \vspace{-10pt}
\end{wrapfigure}

\paragraph{Abstractness Unlocks Diversity.} To examine the link between abstraction and creativity, we analyze how prompt types affect the diversity of the edited images. Using leading open-source (FLUX.2) and closed-source (Gemini 3.1) models, we generate 2,000 images by processing 100 random samples across 10 seeds for each prompt type, abstract and explicit (Fig. \ref{fig:vendi}). We measure the semantic and visual diversity of these outputs using the Vendi Score \citep{friedman2022vendi} on DINOv3 \citep{simeoni2025dinov3} features. We found that abstract prompts consistently produce a higher mean and density of diverse visual concepts than explicit ones. While this shift toward variety is conceptually expected, our results provide a clear quantification of the effect. The gap is especially large in the closed-source model, indicating that while explicit instructions force models toward a narrow output, abstract prompts allow them to explore a much wider manifold of creative interpretations.

\section{Conclusions and Discussion}
\label{sec:discussion}
In this work, we formalized abstract image editing, shifting the evaluation paradigm from holistic ``black box'' scoring to granular diagnostic inspection via our \eval\ framework and the \dataset\ benchmark. We demonstrate that treating image entities as atomic units of assessment provides interpretable, entity-level feedback that correlates with human judgment. Our analysis reveals open-source models are trapped in an under/over-editing trade-off that iterative thinking only superficially mitigates. Despite these insights, our evaluation methodology has inherent limitations. Current VLMs still occasionally struggle with accurate complex scene understanding, which can introduce evaluation noise. Furthermore, our taxonomy focuses on evaluating entities, potentially overlooking cross-entity relations.

Looking forward, \dataset\ can be expanded across diverse domains to test abstractness in new contexts, while \eval\ can be augmented with agentic capabilities like external image retrieval. Crucially, the utility of \eval\ extends beyond assessment: its granular feedback can actively steer generation, either as a reward signal for preference alignment or by driving test-time scaling through critique-and-revise loops. Ultimately, we hope our evaluation suite catalyzes the development of models that move beyond executing technical blueprints to fluidly navigating the ``one-to-many'' interpretation space of abstract human communication.

\section*{Acknowledgments}
{\small
We thank Eyal Ben-David and Avi Caciularu for helpful discussions regarding evaluation decomposition, and Alexandra Danzig for her valuable perspective on visual communication. We also thank Alan Arazi and Michael Toker for their meaningful feedback on the manuscript. Roi Reichart is supported by an Israel Ministry of Science and Technology (MOST) grant on multi-modal AI.}

\bibliographystyle{plain} 
\bibliography{custom}

\begin{thebibliography}{10}

\bibitem{avrahami2022blended}
Omri Avrahami, Dani Lischinski, and Ohad Fried.
\newblock Blended diffusion for text-driven editing of natural images.
\newblock In {\em Proceedings of the IEEE/CVF conference on computer vision and pattern recognition}, pages 18208--18218, 2022.

\bibitem{betker2023improving}
James Betker, Gabriel Goh, Li~Jing, Tim Brooks, Jianfeng Wang, Linjie Li, Long Ouyang, Juntang Zhuang, Joyce Lee, Yufei Guo, et~al.
\newblock Improving image generation with better captions.
\newblock {\em Computer Science. https://cdn. openai. com/papers/dall-e-3. pdf}, 2(3):8, 2023.

\bibitem{binyamin2025make}
Lital Binyamin, Yoad Tewel, Hilit Segev, Eran Hirsch, Royi Rassin, and Gal Chechik.
\newblock Make it count: Text-to-image generation with an accurate number of objects.
\newblock In {\em Proceedings of the Computer Vision and Pattern Recognition Conference}, pages 13242--13251, 2025.

\bibitem{bfl2025representation}
{Black Forest Labs}.
\newblock {FLUX.2}: Analyzing and enhancing the latent space of {FLUX} -- representation comparison, 2025.

\bibitem{bowman2015large}
Samuel Bowman, Gabor Angeli, Christopher Potts, and Christopher~D Manning.
\newblock A large annotated corpus for learning natural language inference.
\newblock In {\em Proceedings of the 2015 conference on empirical methods in natural language processing}, pages 632--642, 2015.

\bibitem{brooks2023instructpix2pix}
Tim Brooks, Aleksander Holynski, and Alexei~A Efros.
\newblock Instructpix2pix: Learning to follow image editing instructions.
\newblock In {\em Proceedings of the IEEE/CVF conference on computer vision and pattern recognition}, pages 18392--18402, 2023.

\bibitem{caesar2018coco}
Holger Caesar, Jasper Uijlings, and Vittorio Ferrari.
\newblock Coco-stuff: Thing and stuff classes in context.
\newblock In {\em Proceedings of the IEEE conference on computer vision and pattern recognition}, pages 1209--1218, 2018.

\bibitem{cai2025hidream}
Qi~Cai, Jingwen Chen, Yang Chen, Yehao Li, Fuchen Long, Yingwei Pan, Zhaofan Qiu, Yiheng Zhang, Fengbin Gao, Peihan Xu, et~al.
\newblock Hidream-i1: A high-efficient image generative foundation model with sparse diffusion transformer.
\newblock {\em arXiv preprint arXiv:2505.22705}, 2025.

\bibitem{chern2023factool}
I~Chern, Steffi Chern, Shiqi Chen, Weizhe Yuan, Kehua Feng, Chunting Zhou, Junxian He, Graham Neubig, Pengfei Liu, et~al.
\newblock Factool: Factuality detection in generative ai--a tool augmented framework for multi-task and multi-domain scenarios.
\newblock {\em arXiv preprint arXiv:2307.13528}, 2023.

\bibitem{clark1991grounding}
Herbert~H Clark and Susan~E Brennan.
\newblock Grounding in communication.
\newblock 1991.

\bibitem{comanici2025gemini}
Gheorghe Comanici, Eric Bieber, Mike Schaekermann, Ice Pasupat, Noveen Sachdeva, Inderjit Dhillon, Marcel Blistein, Ori Ram, Dan Zhang, Evan Rosen, et~al.
\newblock Gemini 2.5: Pushing the frontier with advanced reasoning, multimodality, long context, and next generation agentic capabilities.
\newblock {\em arXiv preprint arXiv:2507.06261}, 2025.

\bibitem{dahary2025decisive}
Omer Dahary, Yehonathan Cohen, Or~Patashnik, Kfir Aberman, and Daniel Cohen-Or.
\newblock Be decisive: Noise-induced layouts for multi-subject generation.
\newblock In {\em Proceedings of the Special Interest Group on Computer Graphics and Interactive Techniques Conference Conference Papers}, pages 1--12, 2025.

\bibitem{deng2025emerging}
Chaorui Deng, Deyao Zhu, Kunchang Li, Chenhui Gou, Feng Li, Zeyu Wang, Shu Zhong, Weihao Yu, Xiaonan Nie, Ziang Song, et~al.
\newblock Emerging properties in unified multimodal pretraining.
\newblock {\em arXiv preprint arXiv:2505.14683}, 2025.

\bibitem{fang2025got}
Rongyao Fang, Chengqi Duan, Kun Wang, Linjiang Huang, Hao Li, Shilin Yan, Hao Tian, Xingyu Zeng, Rui Zhao, Jifeng Dai, et~al.
\newblock Got: Unleashing reasoning capability of multimodal large language model for visual generation and editing.
\newblock {\em arXiv preprint arXiv:2503.10639}, 2025.

\bibitem{friedman2022vendi}
Dan Friedman and Adji~Bousso Dieng.
\newblock The vendi score: A diversity evaluation metric for machine learning.
\newblock {\em arXiv preprint arXiv:2210.02410}, 2022.

\bibitem{fu2023guiding}
Tsu-Jui Fu, Wenze Hu, Xianzhi Du, William~Yang Wang, Yinfei Yang, and Zhe Gan.
\newblock Guiding instruction-based image editing via multimodal large language models.
\newblock {\em arXiv preprint arXiv:2309.17102}, 2023.

\bibitem{gardenfors2004conceptual}
Peter Gardenfors.
\newblock {\em Conceptual spaces: The geometry of thought}.
\newblock MIT press, 2004.

\bibitem{gekhman2023trueteacher}
Zorik Gekhman, Jonathan Herzig, Roee Aharoni, Chen Elkind, and Idan Szpektor.
\newblock Trueteacher: Learning factual consistency evaluation with large language models.
\newblock {\em arXiv preprint arXiv:2305.11171}, 2023.

\bibitem{google_gemini_image_2026}
{Google DeepMind}.
\newblock Image generation with {Gemini} | {Gemini API} | {Google AI} for developers.
\newblock Online Documentation, 2026.
\newblock Accessed: April 6, 2026.

\bibitem{gemini_image_api}
{Google Developers}.
\newblock Gemini {API} documentation: Image generation and editing.
\newblock \url{https://ai.google.dev/gemini-api/docs/image-generation#gemini-image-editing}, 2026.
\newblock Accessed: April 18, 2026.

\bibitem{grice1975logic}
Herbert~P Grice.
\newblock Logic and conversation.
\newblock In {\em Speech acts}, pages 41--58. Brill, 1975.

\bibitem{he2025reasoning}
Qingdong He, Xueqin Chen, Chaoyi Wang, Yanjie Pan, Xiaobin Hu, Zhenye Gan, Yabiao Wang, Chengjie Wang, Xiangtai Li, and Jiangning Zhang.
\newblock Reasoning to edit: Hypothetical instruction-based image editing with visual reasoning.
\newblock {\em arXiv preprint arXiv:2507.01908}, 2025.

\bibitem{honovich2022true}
Or~Honovich, Roee Aharoni, Jonathan Herzig, Hagai Taitelbaum, Doron Kukliansy, Vered Cohen, Thomas Scialom, Idan Szpektor, Avinatan Hassidim, and Yossi Matias.
\newblock True: Re-evaluating factual consistency evaluation.
\newblock {\em arXiv preprint arXiv:2204.04991}, 2022.

\bibitem{hu2023tifa}
Yushi Hu, Benlin Liu, Jungo Kasai, Yizhong Wang, Mari Ostendorf, Ranjay Krishna, and Noah~A Smith.
\newblock Tifa: Accurate and interpretable text-to-image faithfulness evaluation with question answering.
\newblock In {\em Proceedings of the IEEE/CVF International Conference on Computer Vision}, pages 20406--20417, 2023.

\bibitem{huang2024smartedit}
Yuzhou Huang, Liangbin Xie, Xintao Wang, Ziyang Yuan, Xiaodong Cun, Yixiao Ge, Jiantao Zhou, Chao Dong, Rui Huang, Ruimao Zhang, et~al.
\newblock Smartedit: Exploring complex instruction-based image editing with multimodal large language models.
\newblock In {\em Proceedings of the IEEE/CVF Conference on Computer Vision and Pattern Recognition}, pages 8362--8371, 2024.

\bibitem{iakovleva2024specify}
Ekaterina Iakovleva, Fabio Pizzati, Philip Torr, and St{\'e}phane Lathuili{\`e}re.
\newblock Specify and edit: Overcoming ambiguity in text-based image editing.
\newblock {\em arXiv preprint arXiv:2407.20232}, 2024.

\bibitem{jeong2024visual}
Jaeseok Jeong, Junho Kim, Yunjey Choi, Gayoung Lee, and Youngjung Uh.
\newblock Visual style prompting with swapping self-attention.
\newblock {\em arXiv preprint arXiv:2402.12974}, 2024.

\bibitem{ji2025instruction}
Liya Ji, Chenyang Qi, and Qifeng Chen.
\newblock Instruction-based image editing with planning, reasoning, and generation.
\newblock In {\em Proceedings of the IEEE/CVF International Conference on Computer Vision}, pages 17506--17515, 2025.

\bibitem{jia2025compbench}
Bohan Jia, Wenxuan Huang, Yuntian Tang, Junbo Qiao, Jincheng Liao, Shaosheng Cao, Fei Zhao, Zhaopeng Feng, Zhouhong Gu, Zhenfei Yin, et~al.
\newblock Compbench: Benchmarking complex instruction-guided image editing.
\newblock {\em arXiv preprint arXiv:2505.12200}, 2025.

\bibitem{joseph2024iterative}
KJ~Joseph, Prateksha Udhayanan, Tripti Shukla, Aishwarya Agarwal, Srikrishna Karanam, Koustava Goswami, and Balaji~Vasan Srinivasan.
\newblock Iterative multi-granular image editing using diffusion models.
\newblock In {\em Proceedings of the IEEE/CVF winter conference on applications of computer vision}, pages 8107--8116, 2024.

\bibitem{kress2020reading}
Gunther Kress and Theo Van~Leeuwen.
\newblock {\em Reading images: The grammar of visual design}.
\newblock Routledge, 2020.

\bibitem{kryscinski-etal-2020-evaluating}
Wojciech Kryscinski, Bryan McCann, Caiming Xiong, and Richard Socher.
\newblock Evaluating the factual consistency of abstractive text summarization.
\newblock In Bonnie Webber, Trevor Cohn, Yulan He, and Yang Liu, editors, {\em Proceedings of the 2020 Conference on Empirical Methods in Natural Language Processing (EMNLP)}, pages 9332--9346, Online, November 2020. Association for Computational Linguistics.

\bibitem{ku2024viescore}
Max Ku, Dongfu Jiang, Cong Wei, Xiang Yue, and Wenhu Chen.
\newblock Viescore: Towards explainable metrics for conditional image synthesis evaluation.
\newblock In {\em Proceedings of the 62nd Annual Meeting of the Association for Computational Linguistics (Volume 1: Long Papers)}, pages 12268--12290, 2024.

\bibitem{kuznetsova2020open}
Alina Kuznetsova, Hassan Rom, Neil Alldrin, Jasper Uijlings, Ivan Krasin, Jordi Pont-Tuset, Shahab Kamali, Stefan Popov, Matteo Malloci, Alexander Kolesnikov, et~al.
\newblock The open images dataset v4: Unified image classification, object detection, and visual relationship detection at scale.
\newblock {\em International journal of computer vision}, 128(7):1956--1981, 2020.

\bibitem{labs2025flux}
Black~Forest Labs, Stephen Batifol, Andreas Blattmann, Frederic Boesel, Saksham Consul, Cyril Diagne, Tim Dockhorn, Jack English, Zion English, Patrick Esser, et~al.
\newblock Flux. 1 kontext: Flow matching for in-context image generation and editing in latent space.
\newblock {\em arXiv preprint arXiv:2506.15742}, 2025.

\bibitem{li2025editthinker}
Hongyu Li, Manyuan Zhang, Dian Zheng, Ziyu Guo, Yimeng Jia, Kaituo Feng, Hao Yu, Yexin Liu, Yan Feng, Peng Pei, et~al.
\newblock Editthinker: Unlocking iterative reasoning for any image editor.
\newblock {\em arXiv preprint arXiv:2512.05965}, 2025.

\bibitem{liang2022mind}
Victor~Weixin Liang, Yuhui Zhang, Yongchan Kwon, Serena Yeung, and James~Y Zou.
\newblock Mind the gap: Understanding the modality gap in multi-modal contrastive representation learning.
\newblock {\em Advances in Neural Information Processing Systems}, 35:17612--17625, 2022.

\bibitem{lin2024evaluating}
Zhiqiu Lin, Deepak Pathak, Baiqi Li, Jiayao Li, Xide Xia, Graham Neubig, Pengchuan Zhang, and Deva Ramanan.
\newblock Evaluating text-to-visual generation with image-to-text generation.
\newblock In {\em European Conference on Computer Vision}, pages 366--384. Springer, 2024.

\bibitem{liu2025step1x}
Shiyu Liu, Yucheng Han, Peng Xing, Fukun Yin, Rui Wang, Wei Cheng, Jiaqi Liao, Yingming Wang, Honghao Fu, Chunrui Han, et~al.
\newblock Step1x-edit: A practical framework for general image editing.
\newblock {\em arXiv preprint arXiv:2504.17761}, 2025.

\bibitem{liu2022design}
Vivian Liu and Lydia~B Chilton.
\newblock Design guidelines for prompt engineering text-to-image generative models.
\newblock In {\em Proceedings of the 2022 CHI conference on human factors in computing systems}, pages 1--23, 2022.

\bibitem{ma2024i2ebench}
Yiwei Ma, Jiayi Ji, Ke~Ye, Weihuang Lin, Zhibin Wang, Yonghan Zheng, Qiang Zhou, Xiaoshuai Sun, and Rongrong Ji.
\newblock I2ebench: A comprehensive benchmark for instruction-based image editing.
\newblock {\em Advances in Neural Information Processing Systems}, 37:41494--41516, 2024.

\bibitem{maynez2020faithfulness}
Joshua Maynez, Shashi Narayan, Bernd Bohnet, and Ryan McDonald.
\newblock On faithfulness and factuality in abstractive summarization.
\newblock In {\em Proceedings of the 58th annual meeting of the association for computational linguistics}, pages 1906--1919, 2020.

\bibitem{min2023factscore}
Sewon Min, Kalpesh Krishna, Xinxi Lyu, Mike Lewis, Wen-tau Yih, Pang Koh, Mohit Iyyer, Luke Zettlemoyer, and Hannaneh Hajishirzi.
\newblock Factscore: Fine-grained atomic evaluation of factual precision in long form text generation.
\newblock In {\em Proceedings of the 2023 Conference on Empirical Methods in Natural Language Processing}, pages 12076--12100, 2023.

\bibitem{norman2014some}
Donald~A Norman.
\newblock Some observations on mental models.
\newblock In {\em Mental models}, pages 7--14. Psychology Press, 2014.

\bibitem{openai2025gptimage}
{OpenAI}.
\newblock {\em Image generation}, December 2025.
\newblock GPT-Image-1.5 Model Documentation.

\bibitem{oppenlaender2024taxonomy}
Jonas Oppenlaender.
\newblock A taxonomy of prompt modifiers for text-to-image generation.
\newblock {\em Behaviour \& Information Technology}, 43(15):3763--3776, 2024.

\bibitem{pu2025picabench}
Yuandong Pu, Le~Zhuo, Songhao Han, Jinbo Xing, Kaiwen Zhu, Shuo Cao, Bin Fu, Si~Liu, Hongsheng Li, Yu~Qiao, et~al.
\newblock Picabench: How far are we from physically realistic image editing?
\newblock {\em arXiv preprint arXiv:2510.17681}, 2025.

\bibitem{qian2025pico}
Yusu Qian, Eli Bocek-Rivele, Liangchen Song, Jialing Tong, Yinfei Yang, Jiasen Lu, Wenze Hu, and Zhe Gan.
\newblock Pico-banana-400k: A large-scale dataset for text-guided image editing.
\newblock {\em arXiv preprint arXiv:2510.19808}, 2025.

\bibitem{qian2025gie}
Yusu Qian, Jiasen Lu, Tsu-Jui Fu, Xinze Wang, Chen Chen, Yinfei Yang, Wenze Hu, and Zhe Gan.
\newblock Gie-bench: Towards grounded evaluation for text-guided image editing.
\newblock {\em arXiv preprint arXiv:2505.11493}, 2025.

\bibitem{radford2021learning}
Alec Radford, Jong~Wook Kim, Chris Hallacy, Aditya Ramesh, Gabriel Goh, Sandhini Agarwal, Girish Sastry, Amanda Askell, Pamela Mishkin, Jack Clark, et~al.
\newblock Learning transferable visual models from natural language supervision.
\newblock In {\em International conference on machine learning}, pages 8748--8763. PmLR, 2021.

\bibitem{seedream2509seedream}
Team Seedream, Yunpeng Chen, Yu~Gao, Lixue Gong, Meng Guo, Qiushan Guo, Zhiyao Guo, Xiaoxia Hou, Weilin Huang, Yixuan Huang, et~al.
\newblock Seedream 4.0: Toward next-generation multimodal image generation, 2025.
\newblock {\em URL https://arxiv. org/abs/2509.20427}, 6.

\bibitem{sheynin2024emu}
Shelly Sheynin, Adam Polyak, Uriel Singer, Yuval Kirstain, Amit Zohar, Oron Ashual, Devi Parikh, and Yaniv Taigman.
\newblock Emu edit: Precise image editing via recognition and generation tasks.
\newblock In {\em Proceedings of the IEEE/CVF Conference on Computer Vision and Pattern Recognition}, pages 8871--8879, 2024.

\bibitem{simeoni2025dinov3}
Oriane Sim{\'e}oni, Huy~V Vo, Maximilian Seitzer, Federico Baldassarre, Maxime Oquab, Cijo Jose, Vasil Khalidov, Marc Szafraniec, Seungeun Yi, Micha{\"e}l Ramamonjisoa, et~al.
\newblock Dinov3.
\newblock {\em arXiv preprint arXiv:2508.10104}, 2025.

\bibitem{tewel2024add}
Yoad Tewel, Rinon Gal, Dvir Samuel, Yuval Atzmon, Lior Wolf, and Gal Chechik.
\newblock Add-it: Training-free object insertion in images with pretrained diffusion models.
\newblock {\em arXiv preprint arXiv:2411.07232}, 2024.

\bibitem{thrush2022winoground}
Tristan Thrush, Ryan Jiang, Max Bartolo, Amanpreet Singh, Adina Williams, Douwe Kiela, and Candace Ross.
\newblock Winoground: Probing vision and language models for visio-linguistic compositionality.
\newblock In {\em Proceedings of the IEEE/CVF Conference on Computer Vision and Pattern Recognition}, pages 5238--5248, 2022.

\bibitem{wang2025omnistyle}
Ye~Wang, Ruiqi Liu, Jiang Lin, Fei Liu, Zili Yi, Yilin Wang, and Rui Ma.
\newblock Omnistyle: Filtering high quality style transfer data at scale.
\newblock In {\em Proceedings of the Computer Vision and Pattern Recognition Conference}, pages 7847--7856, 2025.

\bibitem{wang2025gpt}
Yuhan Wang, Siwei Yang, Bingchen Zhao, Letian Zhang, Qing Liu, Yuyin Zhou, and Cihang Xie.
\newblock Gpt-image-edit-1.5 m: A million-scale, gpt-generated image dataset.
\newblock {\em arXiv preprint arXiv:2507.21033}, 2025.

\bibitem{wang2024genartist}
Zhenyu Wang, Aoxue Li, Zhenguo Li, and Xihui Liu.
\newblock Genartist: Multimodal llm as an agent for unified image generation and editing.
\newblock {\em Advances in Neural Information Processing Systems}, 37:128374--128395, 2024.

\bibitem{wei2024long}
Jerry Wei, Chengrun Yang, Xinying Song, Yifeng Lu, Nathan Hu, Jie Huang, Dustin Tran, Daiyi Peng, Ruibo Liu, Da~Huang, et~al.
\newblock Long-form factuality in large language models.
\newblock {\em Advances in Neural Information Processing Systems}, 37:80756--80827, 2024.

\bibitem{wu2025qwenimagetechnicalreport}
Chenfei Wu, Jiahao Li, Jingren Zhou, Junyang Lin, Kaiyuan Gao, Kun Yan, Sheng ming Yin, Shuai Bai, Xiao Xu, Yilei Chen, Yuxiang Chen, Zecheng Tang, Zekai Zhang, Zhengyi Wang, An~Yang, Bowen Yu, Chen Cheng, Dayiheng Liu, Deqing Li, Hang Zhang, Hao Meng, Hu~Wei, Jingyuan Ni, Kai Chen, Kuan Cao, Liang Peng, Lin Qu, Minggang Wu, Peng Wang, Shuting Yu, Tingkun Wen, Wensen Feng, Xiaoxiao Xu, Yi~Wang, Yichang Zhang, Yongqiang Zhu, Yujia Wu, Yuxuan Cai, and Zenan Liu.
\newblock Qwen-image technical report, 2025.

\bibitem{wu2025qwen}
Chenfei Wu, Jiahao Li, Jingren Zhou, Junyang Lin, Kaiyuan Gao, Kun Yan, Sheng-ming Yin, Shuai Bai, Xiao Xu, Yilei Chen, et~al.
\newblock Qwen-image technical report.
\newblock {\em arXiv preprint arXiv:2508.02324}, 2025.

\bibitem{yang2025texttt}
Siwei Yang, Mude Hui, Bingchen Zhao, Yuyin Zhou, Nataniel Ruiz, and Cihang Xie.
\newblock Complex-edit: Cot-like instruction generation for complexity-controllable image editing benchmark.
\newblock {\em arXiv preprint arXiv:2504.13143}, 2025.

\bibitem{yao2026photoagent}
Mingde Yao, Zhiyuan You, King-Man Tam, Menglu Wang, and Tianfan Xue.
\newblock Photoagent: Agentic photo editing with exploratory visual aesthetic planning.
\newblock {\em arXiv preprint arXiv:2602.22809}, 2026.

\bibitem{yeh2025beyond}
Chun-Hsiao Yeh, Yilin Wang, Nanxuan Zhao, Richard Zhang, Yuheng Li, Yi~Ma, and Krishna~Kumar Singh.
\newblock Beyond simple edits: X-planner for complex instruction-based image editing.
\newblock {\em arXiv preprint arXiv:2507.05259}, 2025.

\bibitem{yu2025anyedit}
Qifan Yu, Wei Chow, Zhongqi Yue, Kaihang Pan, Yang Wu, Xiaoyang Wan, Juncheng Li, Siliang Tang, Hanwang Zhang, and Yueting Zhuang.
\newblock Anyedit: Mastering unified high-quality image editing for any idea.
\newblock In {\em Proceedings of the Computer Vision and Pattern Recognition Conference}, pages 26125--26135, 2025.

\bibitem{yuan2024flexedit}
Tianshuo Yuan, Yuxiang Lin, Jue Wang, Zhi-Qi Cheng, Xiaolong Wang, Jiao GH, Wei Chen, and Xiaojiang Peng.
\newblock Flexedit: Marrying free-shape masks to vllm for flexible image editing.
\newblock {\em arXiv preprint arXiv:2408.12429}, 2024.

\bibitem{zamfirescu2023johnny}
J~Diego Zamfirescu-Pereira, Richmond~Y Wong, Bjoern Hartmann, and Qian Yang.
\newblock Why johnny can’t prompt: how non-ai experts try (and fail) to design llm prompts.
\newblock In {\em Proceedings of the 2023 CHI conference on human factors in computing systems}, pages 1--21, 2023.

\bibitem{zeng2025editworld}
Bohan Zeng, Ling Yang, Jiaming Liu, Minghao Xu, Yuanxing Zhang, Pengfei Wan, Wentao Zhang, and Shuicheng Yan.
\newblock Editworld: Simulating world dynamics for instruction-following image editing.
\newblock In {\em Proceedings of the 33rd ACM International Conference on Multimedia}, pages 12674--12681, 2025.

\bibitem{zhang2025re}
Huixuan Zhang and Xiaojun Wan.
\newblock Re-thinking the automatic evaluation of image-text alignment in text-to-image models.
\newblock {\em arXiv preprint arXiv:2506.08480}, 2025.

\bibitem{zhang2023magicbrush}
Kai Zhang, Lingbo Mo, Wenhu Chen, Huan Sun, and Yu~Su.
\newblock Magicbrush: A manually annotated dataset for instruction-guided image editing.
\newblock {\em Advances in Neural Information Processing Systems}, 36:31428--31449, 2023.

\bibitem{zhang2018unreasonable}
Richard Zhang, Phillip Isola, Alexei~A Efros, Eli Shechtman, and Oliver Wang.
\newblock The unreasonable effectiveness of deep features as a perceptual metric.
\newblock In {\em Proceedings of the IEEE conference on computer vision and pattern recognition}, pages 586--595, 2018.

\end{thebibliography}

\newpage
\appendix
\section*{Appendix}

\section*{Limitations}
\label{app:limitation}
While our dataset and evaluation framework provide a robust foundation for evaluating abstract instruction following, we acknowledge several limitations. A primary limitation involves the scope of abstract intents and cultural bias. Although our dataset spans physical, logical, emotional, and social domains, abstract concepts are inherently subjective and culturally dependent. Our prompts are constructed in English and may inadvertently reflect Western interpretations of concepts like professionalism or mood, suggesting future iterations should expand to multilingual and cross-cultural variations. Additionally, our benchmark suffers from inherited dataset biases. Because the context images in our dataset are sourced from the Open Images dataset, any underlying demographic or geographic representation biases present in that original data naturally carry over into our benchmark. 

\section*{Broader Impact}
\label{app:impact}
The ability of image generation models to understand abstract instructions has significant societal implications with both positive applications and potential risks. Our analysis provides valuable insights into model behavior, helping users navigate the gap between rigid explicit commands and open ended human intent. By revealing how different models interpret abstract instructions, our findings can guide users in crafting more effective prompts, making AI assisted design more accessible to those without extensive prompt engineering expertise. Furthermore, our evaluation provides developers with a concrete metric to identify and fix failure modes like context destruction, leading to safer and more reliable models. However, the primary risk of abstract editing lies in latent model biases. When an instruction is ambiguous, the model must fill in the unspecified details. Left unchecked, models may rely on harmful stereotypes to resolve this ambiguity and exacerbate representation biases. By formalizing the axes of identification and specificity, our taxonomy exposes exactly where these inferential leaps occur. We strongly encourage researchers to use our dataset not just for capability benchmarking but as an auditing tool to uncover how different architectures interpret socially sensitive abstractions.

\section*{Computational Requirements}
\label{app:resources}
Generating the dataset, establishing explicit baselines, and evaluating the models required specific computational resources and API access. Generating the prompt pairs required the Gemini API to create both abstract and explicit instructions for all relevant categories at once per context image, resulting in 1300 API calls across the 1300 context images sourced from the Open Images dataset. For model inference, we utilized a Google Cloud Platform VM equipped with a single NVIDIA A100 80GB GPU. Running inference for the eight open source models to generate the 940 test set samples per model took approximately 150 GPU hours in total. The evaluation phase required a VLM, specifically Gemini 3 Flash, to process two prompt types across 13 models and 470 samples. This evaluation pipeline required two separate API calls per sample to independently assess the initial context image and the final edited image, totaling 24,440 base API calls. This figure represents only the final evaluation pass and does not include the additional compute and API usage required for experimental prompt optimization during the development phase.

\clearpage
\section{Taxonomy and Definition}
\label{app:theory_definition}

\subsection{Clarifications}
\label{app:implicit_def}

\begin{figure*}[ht] %
    \centering
    \includegraphics[width=0.9\textwidth]{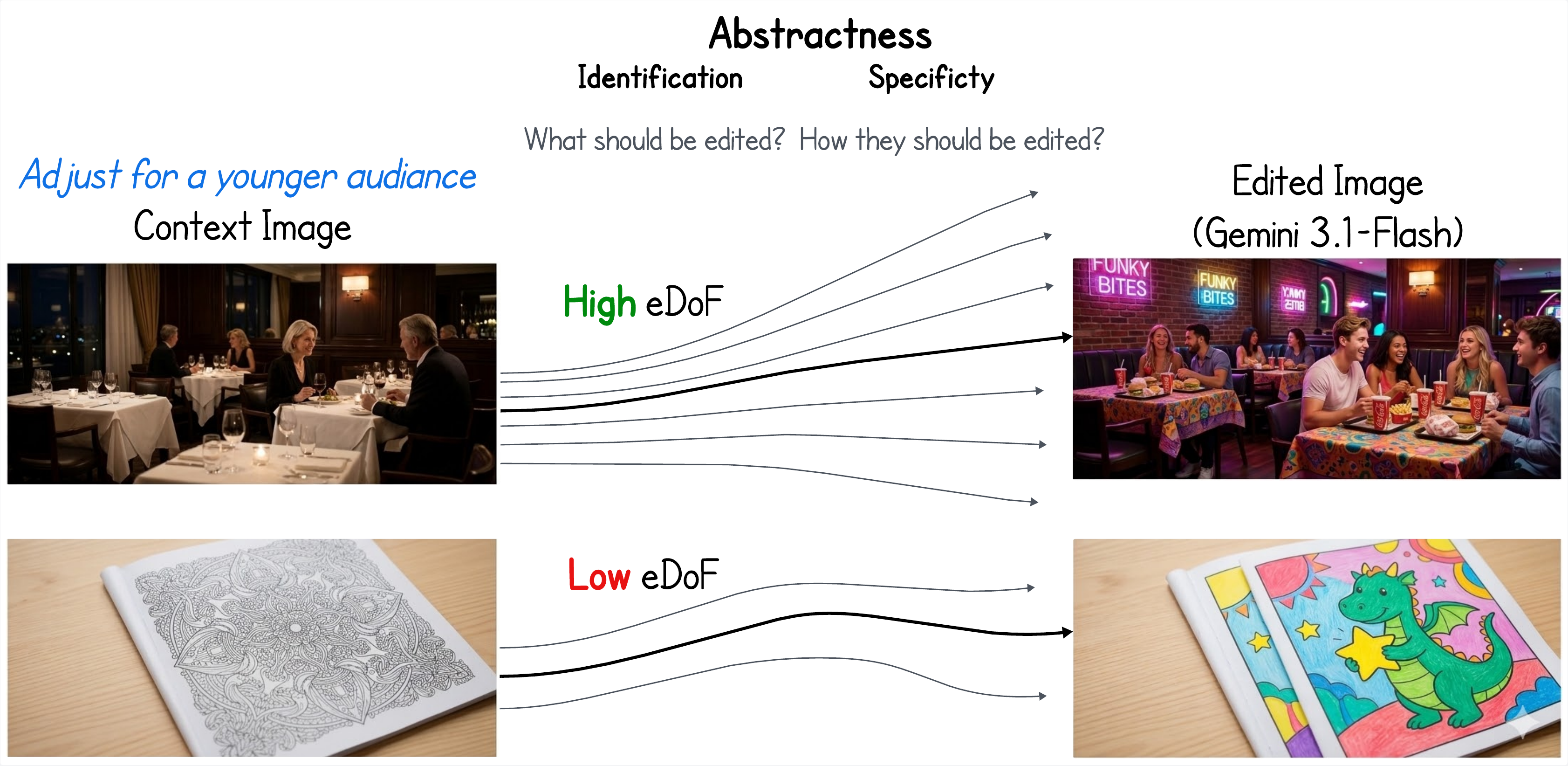}
    \caption{Context-Dependent $eDoF$. The same instruction, "Adjust for a younger audience'', requires varying levels of generative guesswork depending on the context image $\mathcal{I}_c$. In the High $eDoF$ case (top), the model must navigate a broad probability space to resolve both Identification (what to change) and Specificity (how to change it). In the Low $eDoF$ case (bottom), the visual environment provides explicit constraints, narrowing the model's interpretive path.}
    \label{fig:context_implicit}
\end{figure*}

\paragraph{Context Image as part of the definition.} The importance of the context image $\mathcal{I}_c$ lies in its role as the anchor for the instruction $p$, as the degree of generative guesswork is a function of the image-text pair rather than the text alone. An instruction can be interpreted differently depending on the visual environment; for instance (see Figure \ref{fig:context_implicit}), the prompt ``make it more formal'' might require adding a suit to a person in a Physical context, whereas in an Artistic context, it could involve shifting the image composition toward symmetry. Furthermore, the visual state of $\mathcal{I}_c$ determines the $eDoF$ level by defining the available constraints. Consider the instruction ``adjust for a younger audience''. In a photo of a serious, high-end restaurant, the model faces high $eDoF$ as it must creatively synthesize sweeping changes to the lighting, decor, and plating to match the concept. Conversely, in an image of a dark and grayscale coloring book page, the same instruction becomes relatively explicit with low $eDoF$ because the visual context clearly points to applying a vibrant, bright color palette. Thus, the context image provides the necessary grounding to resolve underspecified intent and quantify the model's reasoning.

\begin{figure}[!ht]
    \centering
    \includegraphics[width=1\linewidth]{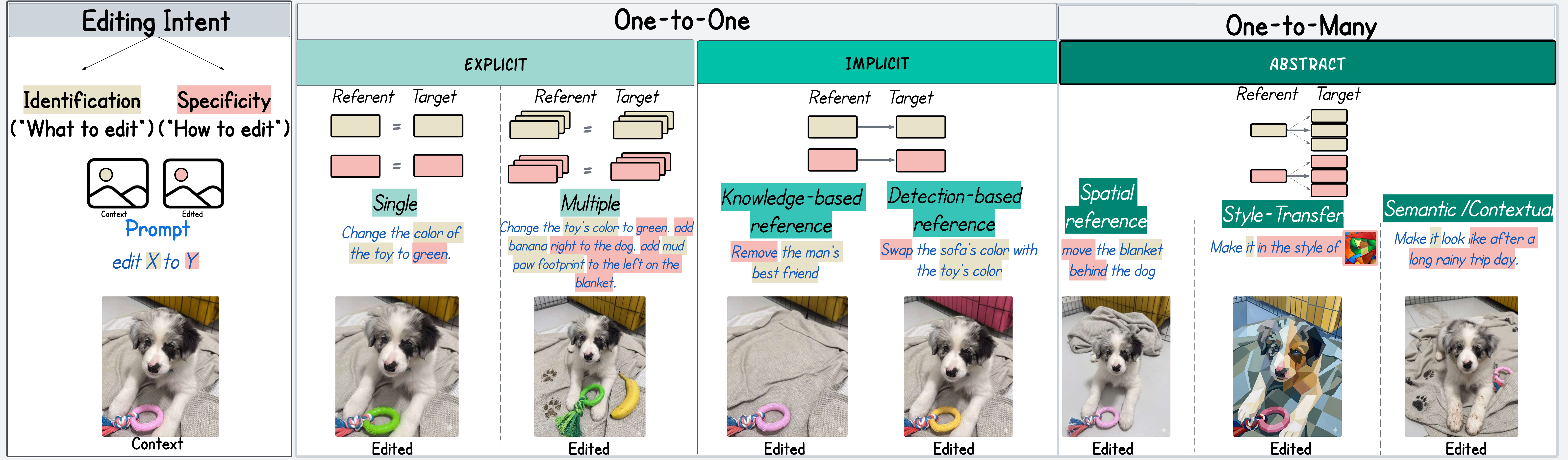}
    \caption{Expanded Cases of Editing Intent.}
    \label{fig:expanded_intent}
\end{figure}

\paragraph{Conceptualizing Editing Degrees of Freedom.} Editing Degrees of Freedom (eDoF) represents the continuous spectrum of valid interpretations an editing instruction can evoke. Quantifying this spectrum is inherently challenging, as assigning a strict metric to human imagination and semantic ambiguity quickly transitions from a computational problem into a philosophical one. Given this difficulty, it is more practical to conceptualize eDoF in ranges, specifically by mapping transitions from deterministic one-to-one instructions to open-ended one-to-many (or one-to-few) instructions based on the number of plausible outcomes. Explicit instructions represent the lowest eDoF, dictating direct and unambiguous changes. Still within the one-to-one regime are implicit instructions, which, rather than requiring abstract reasoning, rely on external computer vision capabilities like object detection or external knowledge retrieval to resolve the referent. Conversely, abstract instructions represent the highest eDoF, encompassing one-to-many scenarios such as spatial referencing, artistic transformations, and style transfer, as demonstrated in Figure \ref{fig:expanded_intent}. While this work primarily focuses on the semantic and contextual abstractness of human intent in image editing, expanding this framework to fully capture these broader categories of interpretation remains an exciting avenue for future research.

\paragraph{eDoF and Domains.} Crucially, a sample's position on the eDoF scale is inherently tied to its semantic domain \citep{gardenfors2004conceptual}. Physically bounded edits, such as changing the season to winter, usually carry lower interpretive variance and naturally align with lower $eDoF$ scores. Edits involving complex social or emotional shifts, such as making a scene appear formal, introduce vastly higher interpretive freedom, placing them at the upper limit of the scale. Furthermore, this level of abstractness is strictly independent of spatial resolution. Instructions can manifest as highly abstract local modifications (e.g., ``make the shirt more formal'') or completely explicit global transformations (e.g., ``reduce brightness by 1 in each color channel'').

\paragraph{From the Theoretical Consensus to the Tractable Practice.}
\mv{The practical implementation of this taxonomy is challenged by the continuous visual space abstract editing spans, making a consensus-based definition difficult to quantify. Beyond the mathematical intractability of mapping an infinite interpretation set, consensus is fractured by multi-level subjectivity, ranging from human cultural variance to the architectural biases of VLM judges and the spatial constraints of different generative seeds. It also suffers from the ambiguity of granularity, which directly impacts our core axes: Identification becomes a recursive problem, where the target can shift from a general entity (a ``dog'') to hyper-local components (the ``pose of its ear'') without a natural stopping point. Similarly, Specificity expands into second-order complexity ($O(n^{2})$) when an edit targets the relational interaction between entities (e.g., ``make them look in love'') rather than an isolated attribute, requiring the model to identify and modify the mutual dependency between multiple entities. To navigate this, our framework adopts a ``rough-cut'' approach based on three distinct methodological choices. First, we prioritize precision over recall in our evaluation; rather than attempting to account for the exhaustive recall of all valid creative interpretations, we focus on the precision of a model’s editing, ensuring the chosen output is logically grounded and verifiable. Second, we utilize a high-reasoning VLM as a proxy for a mutual judge across all experiments, employing fixed generative seeds within the image editing models to minimize stochastic variance. Third, we constrain the problem of granularity by grounding both identification and specificity to the immediate requirements of the visual context, effectively limiting the scope of modification to what is necessary for a successful edit.}

\subsection{Potential Abstractness Expansions}

While the specificity axis is central to framing abstract image edits, it alone provides an incomplete definition.

\paragraph{Causality Considerations.}
Specificity can be framed as a directional space where a single edit triggers a sequential cascade of subsequent changes. This ``edit cascade'' is defined by the causal relationships expected between explicit atomic edits. For example, the instruction ``make an earthquake'' in a crowded street scene implies a direct causal trajectory: cracked ground $\rightarrow$ structural damage $\rightarrow$ panicked pedestrians. Expanding the Specificity axis to include these trajectories allows for a more accurate formalization of abstract editing. However, this approach introduces significant challenges inherent to causality itself, such as determining the ground truth order of effects in multi-turn edits and distinguishing between correlation and true causation. While we believe integrating causality is essential for a complete understanding of the field, it remains beyond the scope of this work; we hope to inspire future research into this intersection.


\paragraph{Spatial Considerations.}
The Specificity axis is further influenced by a critical challenge that was previously addressed \cite{tewel2024add}: determining where to insert a new object, or where to move an existing one, and at what scale relative to the surrounding context. This extends the problem beyond semantics into the realm of spatial reasoning. Even when an entity is explicitly named, its position and scale may remain entirely abstract, requiring the model to infer placement based on scene understanding. Specifically, we identify \textit{abstract location}, where the model must determine a plausible placement for an object when the instruction is general, such as deciding where to "Add Snoopy", and abstract size, where the model must determine the appropriate scale of an object for actions like adding, replacing, or transforming. These two dimensions introduce significant spatial ambiguity into the editing process. While our test set includes physical domain categories such as size and point of view, the broader uncertainty regarding object placement remains beyond the scope of this work.

\paragraph{Relational Interaction Considerations.} A more holistic evaluation would ideally account for the interactions between entities within a scene. However, this introduces substantial scalability and complexity issues; for $n$ entities, the relational space expands at $O(n^2)$. Assessing the nuanced changes in these dependencies, such as spatial shifts or physical contact, is computationally demanding and difficult to quantify. Consequently, while these interactions are vital for scene coherence, evaluating them requires a careful balance between modeling the full relational web and maintaining a tractable evaluation framework.

\paragraph{Generalizing to Stylistic Intents.} While artistic style transfer \citep{jeong2024visual, wang2025omnistyle, yao2026photoagent} involves global transformations, these works primarily evaluate familiarity with pre-existing visual distributions, such as specific artistic movements. In contrast, \dataset\ focuses on situational intents that do not map to a single stylistic category. For example, a style prompt dictates a uniform aesthetic change, whereas an abstract prompt like ``make the dog look like it just returned from a long rainy trip'' requires the model to autonomously deduce and execute specific semantic edits, such as adding mud or wetting fur. Consequently, style transfer represents a constrained form of abstraction relying on specific external knowledge, while our work addresses the open-ended interpretation of complex human intent.
Integrating elements of artistic reference with open-ended editing instructions remains a compelling direction for future work, potentially bridging the gap between concept-driven stylization and intent-driven content modification.

\section{Dataset Statistics \& Curation Details}
\label{app:dataset_curation}

\begin{figure*}[ht]
\centering
\includegraphics[width=1.0\linewidth]{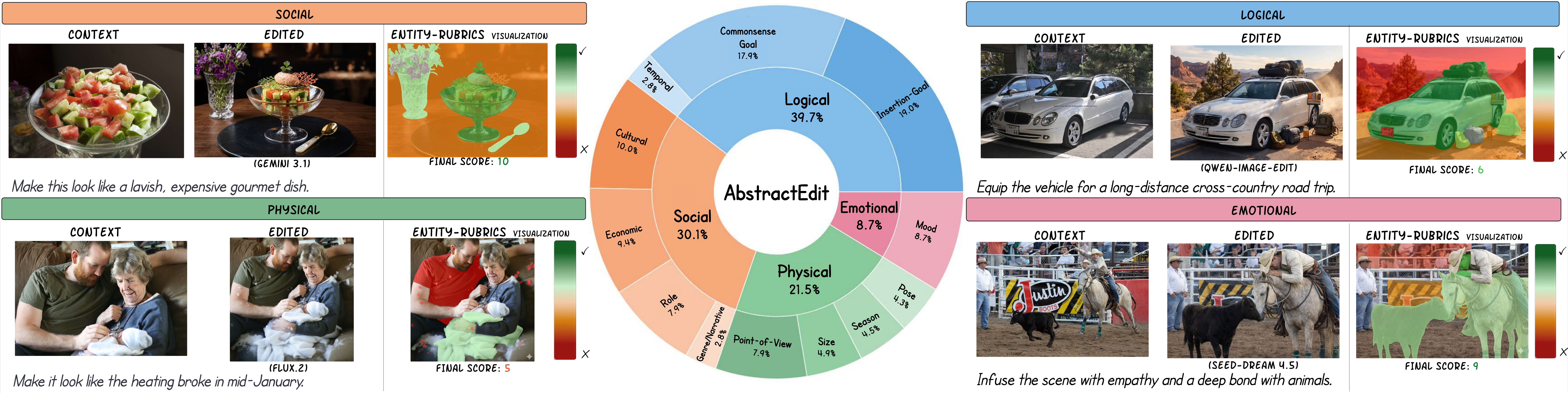}
\caption{Composition and Evaluation of the \dataset\ Benchmark. The composition of the \dataset\ benchmark is illustrated across four primary domains and 12 subcategories in the middle panel. Surrounding this distribution are representative samples from each domain, featuring context images paired with candidate edits produced by various models. Each output is overlaid with our \eval\ automatic granular evaluation, providing a visual layout of performance that ranges from red (incorrect) to green (correct) at the entity level.}
\label{fig:sub_viz_bench760}
\end{figure*}

\begin{figure}
    \centering
    \includegraphics[width=0.7\linewidth]{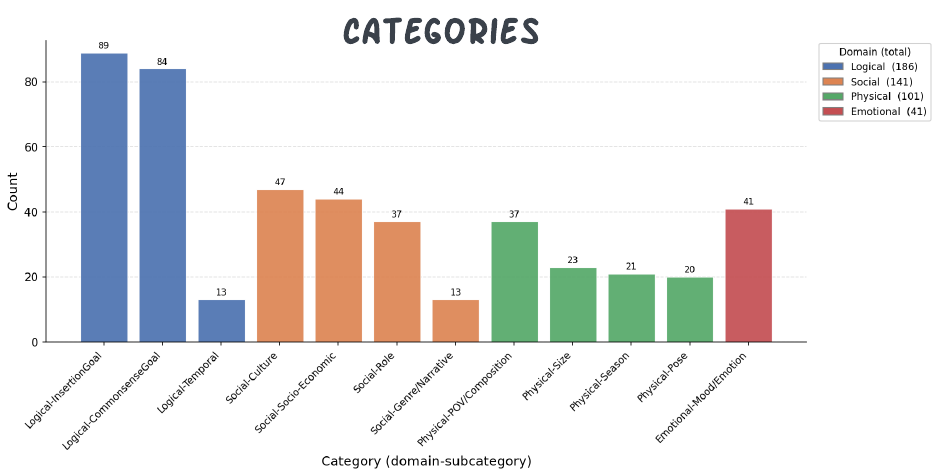}
    \caption{\dataset\ distribution of categories.}
    \label{fig:counts_categories}
\end{figure}

\paragraph{Domain and Subcategory Distribution of \dataset.} Fig. \ref{fig:sub_viz_bench760} and 
Fig. \ref{fig:counts_categories} detail the composition of the \dataset\ benchmark across its four primary domains and their fine-grained categories. The quantitative distribution: Logical (186), Social (141), Physical (101), and Emotional (41), is intentionally designed to reflect the varied cognitive demands required for abstract image editing. The \textbf{Logical} domain is the most heavily represented, driven primarily by the \textit{Insertion Goal} (89) and \textit{Commonsense Goal} (84) subcategories. This prioritization is motivated by the fact that functional, deductive reasoning, inferring \textit{what} specific objects or state changes are required to fulfill an abstract need (e.g., "prepare for rain" resolving to "add an umbrella"), forms the foundational backbone of abstract human instruction. The \textbf{Social} domain follows closely, including complex human-centric categories such as \textit{Culture} (47), \textit{Socio-Economic} (44), and \textit{Role} (37). The strong emphasis on this domain challenges the model's capacity to translate abstract sociological concepts into concrete visual markers (e.g., clothing, environment), a task that requires deep semantic grounding often absent in standard diffusion models. The \textbf{Physical} domain focuses on abstract intents that resolve into structural modifications, such as shifting \textit{POV/Composition} (37) or \textit{Size} (23) to alter the perceived focus or power dynamic of a scene. Finally, while the \textbf{Emotional} domain has the lowest sample count, its singular focus on \textit{Mood/Emotion} (41) provides a critical test of atmospheric interpretation, requiring the model to manipulate global image semantics, such as lighting, weather, and color grading, to evoke subjective feelings. Together, this hierarchical distribution ensures the benchmark rigorously tests a wide spectrum of abstraction, ranging from objective problem-solving to subjective atmospheric translation. See Table \ref{tab:train_category_dist} for the domain distribution of the train set (4000 samples).

\begin{table}[!ht]
    \centering
    \caption{Category Distribution in \dataset\ Training Set.}
    \label{tab:train_category_dist}
    \begin{tabular}{lrr}
        \toprule
        \textbf{Domain} & \textbf{Count} & \textbf{Percentage (\%)} \\
        \midrule
        Social & 1,470 & 35.7\% \\
        Physical & 1,470 & 35.7\% \\
        Logical & 882 & 21.4\% \\
        Emotional & 294 & 7.1\% \\
        \midrule
        \textbf{Total} & \textbf{4,116} & \textbf{100.0\%} \\
        \bottomrule
    \end{tabular}
\end{table}

\begin{figure}
    \centering
    \includegraphics[width=0.85\linewidth]{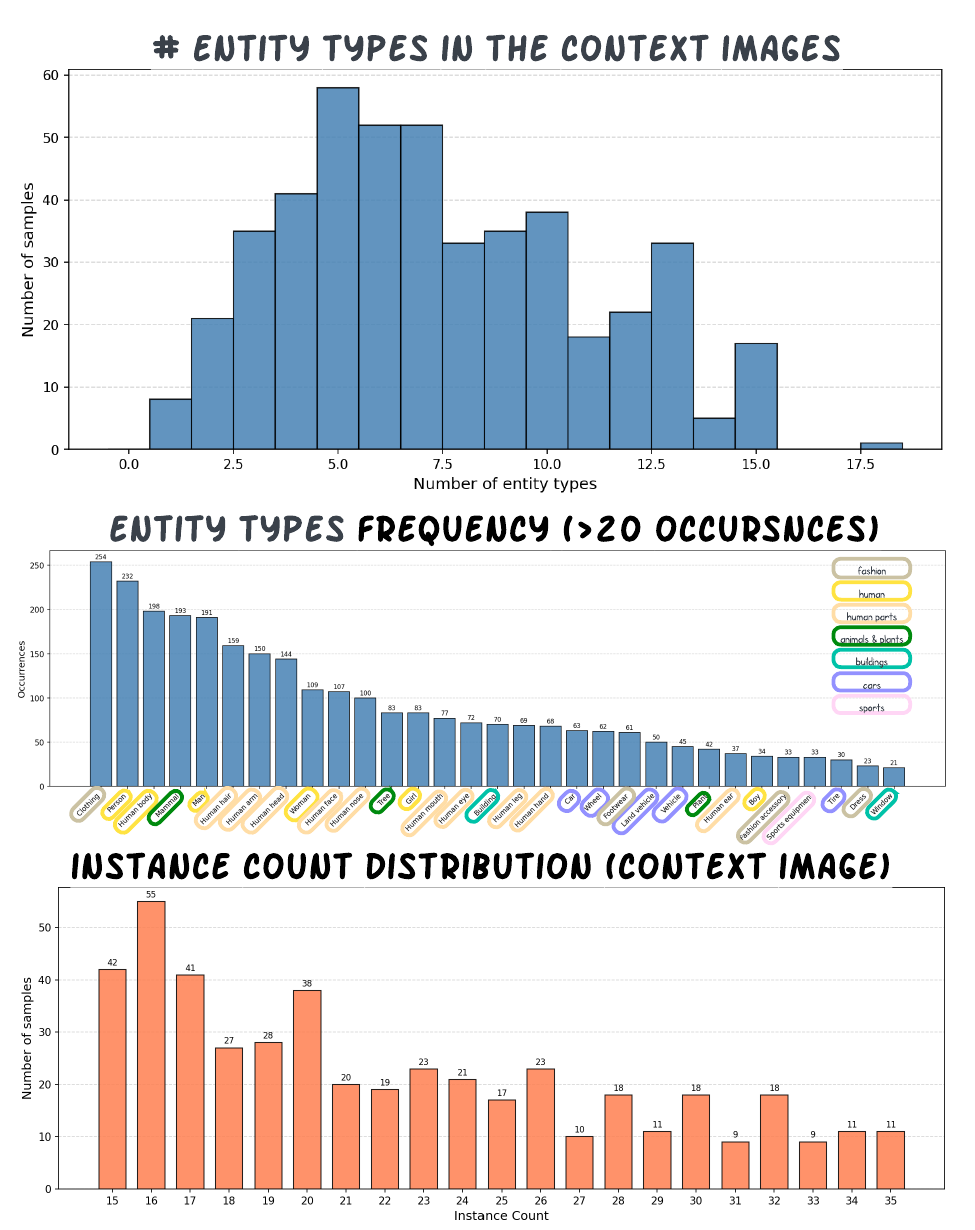}
    \caption{Distribution of Common Entities in \dataset\ Context Images (OpenImages).}
    \label{fig:context_image_entity_types_stats}
\end{figure}

\paragraph{Context Image Composition and Density.}
Figure \ref{fig:context_image_entity_types_stats} illustrates the semantic diversity of the base context images selected from the OpenImages v7 dataset \citep{kuznetsova2020open}. The top distribution demonstrates the variety of distinct entity types present within individual samples; the majority of images contain between 2 and 12 unique categories, peaking prominently at 5 distinct entity types [24]. This guarantees that the benchmark evaluates conceptual editing within compositionally rich environments rather than simplified, single-object backgrounds. The frequency analysis of these entities (middle) reveals a strong prevalence of human-centric and everyday subjects, heavily dominated by categories such as ``Clothing'' (254 occurrences) [19, 32], ``Person'' (232 occurrences) [34, 40], and ``Human body'' (196 occurrences)[37, 41]. This grounds the dataset in relatable, real-world scenarios, which is a prerequisite for supporting the complex interpersonal and structural interactions required by the Social and Physical evaluation domains. Finally, the instance count distribution (bottom) highlights the high visual density of the chosen images, with most samples containing between 15 and 35 distinct bounding box instances [106, 123], and a notable peak at 16 instances [63, 107]. This deliberate level of visual clutter introduces realistic scene distractors, ensuring that models must not only interpret the abstract intent but also accurately localize the edit without inadvertently corrupting a dense surrounding context.

\begin{figure}[!ht]
    \centering
    \includegraphics[width=0.6\linewidth]{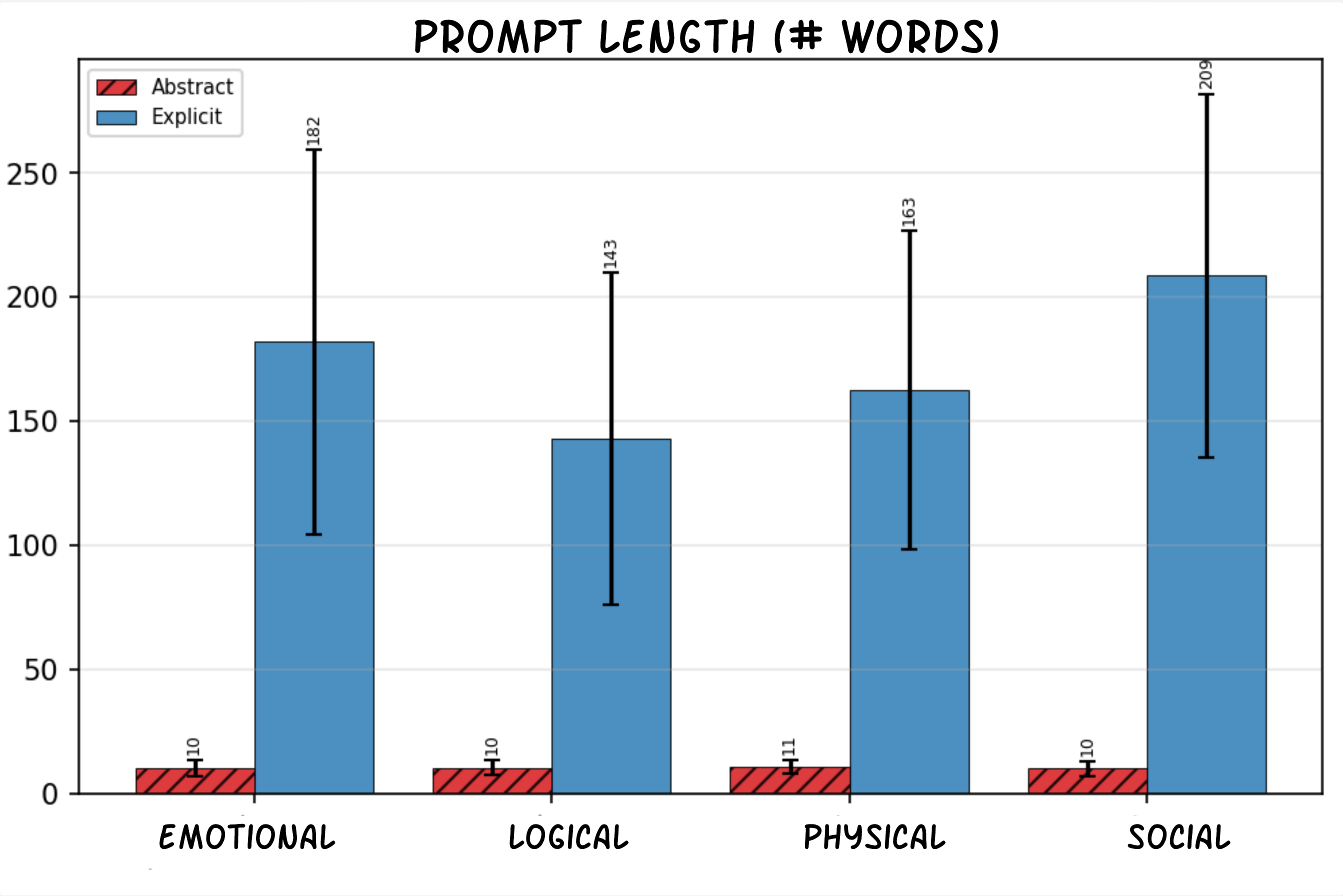}
    \caption{Explicit vs Abstract Prompt Length.}
    \label{fig:prompt_length_abstract_explicit}
\end{figure}

\paragraph{Abstract and Explicit Prompt Lengths.}
Figure \ref{fig:prompt_length_abstract_explicit} presents the distribution of prompt lengths, measured in words, for both abstract and explicit instructions across the four evaluation domains (Emotional, Logical, Physical, and Social) in \dataset. The most prominent observation is the stark contrast in length and variance between the two prompt types, which quantitatively illustrates the "semantic gap" inherent in abstract editing. Abstract instructions are consistently brief, averaging roughly 10 to 11 words as was limited in the curation process ($<15$ words), directly reflecting their high-level, underspecified nature. Conversely, the paired explicit instructions are significantly longer and exhibit substantial variance, acting as comprehensive, step-by-step blueprints for execution. Among the domains, the \textbf{Social} category requires the most extensive explicit instructions (averaging 209 words), likely due to the complexity of describing interpersonal dynamics and contextual scene modifications. In contrast, the \textbf{Logical} domain requires the fewest words on average (143 words) to explicitly ground the intended edits. This massive length discrepancy highlights the heavy lifting required, by either the human annotator or the model's text encoder, to translate a concise abstract concept into a complete set of executable visual steps.

\paragraph{Personas.}
To ensure instruction diversity, we employ a high-dimensional persona sampling strategy that simulates a wide array of human backgrounds and behaviors (see the prompt box). By sampling features such as age, country, profession, and personality as independent variables, the model generates instructions spanning different cultural contexts and linguistic styles. This approach captures the intersection of a user's technical ability, ranging from AI enthusiasts to tech-phobic individuals, with their specific creative motivations like artistic expression or practical utility. By grounding each request in a unique combination of visual language and hobby, the system prevents mode collapse and produces a dataset that reflects the authentic variety of real-world image editing needs. This feature space yields $10^{10}$ unique persona combinations.

\clearpage
\begin{tcolorbox}[
    enhanced,
    breakable,
    colback=pastelgreen, 
    colframe=gray!40, 
    sharp corners, 
    boxrule=0.5pt,
    title= Automatic Curation pipeline of \dataset\ Curation,
    coltitle=black,
    fonttitle=\bfseries,
    fontupper=\small\ttfamily 
]
\small
\label{box:instruction_gen}
\begin{verbatim}
random_paersona = persona_prompt_generator()
system_prompt_2 = f"""

{random_paersona} Your goal is to generate creative and detailed editing 
instructions for an image based on a set of approved categories. Your goal 
is also to translate creative, high-level ideas (**Abstract Prompts**) into 
specific, technical, and deterministic steps (**Explicit Instructions**).
Each response must be sampled at random from the distribution.

## CONTEXT
- **Image Entities**: You will be given a list of entities detected in the 
image: {', '.join(entities)}. You can detect additional entities if you 
think they are relevant for the edit, but you cannot remove any of the 
provided entities.
- **Relevant Categories**: You will be given a list of categories, each with 
an example of a high-level creative prompt.

## YOUR TASK
Your overall task is to generate editing instructions for a list of categories. 
This list includes all of the `Relevant Categories` provided to you, **PLUS one 
new, creative category of your own invention (the category name will start 
with NEW_ prefix)** that you believe is a good fit for the image.
For **EACH** category in this combined list (both provided and self-generated), 
you must generate a complete set of editing instructions by following these steps:

1. **Create an Abstract Prompt (The "What")**: Write a new, creative, and 
high-level prompt inspired by the category's theme. This defines the overall 
'vibe' of the edit. The instruction must be open to interpretation in terms of 
How exactly to achieve it, I.e, different people will interpret it differently.
2. **Probablity Sampling**: Ensure that each generated response is sampled 
randomly from the distribution.
3. **Define Entity-Specific Edits (The "How")**: 
    - **Discovery**: Actively identify more relevant entities in the image or 
    context that were not provided in the {entities} list. Include them in 
    your explicit list.
        For each relevant `Image Entity`, determine its role in the edit. This 
        is the technical process. It is precise, measurable, and leaves no 
        room for interpretation. It focuses on concrete properties:
        Write a clear, precise, and actionable instruction for the entity with 
        specific guidance. Keep it blank if no change is needed to the entity 
        to advance the alignment to abstract prompt.             
        -- **Granularity**: Relate to the relevant entity components (e.g., for 
        a 'person', specify edits for their clothing, hair, and posture in 
        separate sentences).
        -- **Density**: Aim for a high density of edits. Each sentence must 
        contain exactly ONE technical change. Multiple sentences can be 
        generated for the same entity as long as they refer to different 
        components or aspects of edit of the entity.  
        -- **Technical Properties**: Use specific values: "Increase roughness 
        to 80%," "Shift hue to #228B22," "Reduce focal depth to 2.0," or 
        "Add 15% film grain."
4. **Define a General Instruction**: Write instructions that describes any 
global changes to the image's overall e.g., style, lighting, mood, or color 
palette. This applies to the entire scene.
5. **Insertion/Removal Entities**: You may suggest new entities to be 
inserted in the image to better align with the abstract prompt as an explicit 
instruction specifying the entity and its placement. 
    These should be realistic and contextually appropriate. If no 
    insertions/removals are needed, leave this blank.
    For each insertion, you MUST provide coordinates or relative placement 
    (e.g., "Place a [Object] 10cm to the left of [Entity] at a 30-degree angle").

## OUTPUT REQUIREMENTS
- You **must** process every category provided in the input.
- You MUST return your complete analysis of all categories by calling the 
`DiverseImageAnalysisResult_abstractExplicit` function.
- You must ensure diversity in your generated prompts by sampling randomly 
from the distribution.
- Each abstract prompt must not be longer than 15 words, and has to keep its 
abstractness while being specific to the image content and user context.
- Each abstract prompt has to look like user-generated by the persona mentioned 
in the beginning, suitable for them using a mobile phone on the go, trying to 
quickly convey a creative idea.
- Your generated **explicit instructions** and **general instruction** must 
**NEVER** use subjective or interpretive words. only specific technical clear 
words for specific change.
- High-density output is mandatory: provide as many specific, atomic 
instructions as possible per entity.
- Avoid adjectives like "whimsical", "vast", "vibrant", which are not 
user-common.

**Begin your analysis for the following categories. Examples are provided only 
for inspiration. Do not use them as templates. Follow each category guidelines:**
{relevant_cats_with_examples_str}
"""
\end{verbatim}
\end{tcolorbox}


\clearpage
\begin{tcolorbox}[
    colback=Lavender!30, 
    colframe=gray!40,    
    arc=4pt,             
    boxrule=0.5pt,       
    fontupper=\small\sffamily, 
    title=\centering\textbf{Persona-Driven Instruction Diversity},
    fonttitle=\small\sffamily,
    coltitle=black,      
    titlerule=0.5pt,     
    label=box:persona_prompt 
]
\textbf{Feature Space} $\mathcal{F}$:\\[1mm] 
\textbf{Age:} Gen Alpha, Gen Z, Millennial, Gen X, Boomer, Elderly, Teenager, Toddler-parent, Young Adult, Retiree \\
\textbf{Country:} Japan, Brazil, Norway, Kenya, India, USA, France, Australia, Egypt, Mexico \\
\textbf{Gender:} Non-binary, Female, Male, Agender, Genderfluid, Trans-masculine, Trans-feminine, Bigender, Queer, Undisclosed \\
\textbf{Hobby:} Urban Exploration, Gardening, Cyber-gaming, Birdwatching, Cosplay, Minimalist Design, Oil Painting, Skateboarding, Baking, Street Photography \\
\textbf{Profession:} Florist, Data Scientist, History Teacher, DJ, Architect, Mechanic, Social Media Influencer, Nurse, Student, Chef \\
\textbf{Tech Skill:} Smartphone novice, Power user, Professional Editor, Casual social poster, Tech-phobic, AI enthusiast, Tablet artist, Analog lover, Gamer, Software Dev \\
\textbf{Visual Lang:} Cinematic, Abstract, High-Contrast, Pastel, Gritty/Industrial, Surreal, Symmetrical, Over-saturated, Muted/Desaturated, Low-Poly \\
\textbf{Personality:} Sarcastic, Highly optimistic, Melancholic, Efficient/Direct, Whimsical, Detail-oriented, Impatient, Poetic, Skeptical, Cheerful \\
\textbf{Name:} Yuki, Mateo, Astrid, Akachi, Priya, Jax, Elodie, Liam, Zaynab, Santiago \\
\textbf{Motivation:} Artistic expression, Quick fix, Social media clout, Memory preservation, Creating a gift, Practical utility, Boredom, Satire, Visual storytelling, Professional portfolio 

\tcblower 

\textbf{System Prompt Template} $\mathcal{S}$:\\[1mm]
\textit{``You are an expert image editing assistant. You currently simulate how a user named \textbf{\{name\}} from \textbf{\{country\}}—a \textbf{\{age\}} \textbf{\{gender\}} who works as a \textbf{\{profession\}} and enjoys \textbf{\{hobby\}}—would generate creative abstract editing instructions. This user is a \textbf{\{tech\_skill\}} motivated by \textbf{\{motivation\}}. They describe things using a \textbf{\{visual\_lang\}} visual language and have a \textbf{\{personality\}} personality.''}
\end{tcolorbox}

\begin{tcolorbox}[
    breakable, 
    colback=softgold, 
    colframe=black, 
    width=\columnwidth, 
    sharp corners, 
    boxrule=0.5pt,
    fontupper=\small\sffamily, 
    fonttitle=\small\sffamily\bfseries, 
    title=\centering Domains and Categories: Guidelines,
    colbacktitle=softgold,
    coltitle=black,
    toptitle=1mm,
    bottomtitle=1mm,
    before skip=10pt,
    after skip=10pt,
    label=box:domains_guide,
]
\raggedright

\textbf{Emotional-Mood/Emotion:} Fundamentally alters the emotional atmosphere of the image through changes in color, light, and context, or of people/animals emotions. \textit{Requirement: Use different moods / emotions such as joyful, peaceful, sad, angry, fearful, surprised, curious, proud, lonely, hopeful.} 

\textbf{Physical-Material:} Manipulates the sensory surface qualities and physical materials of objects. \textit{Requirement: Must contain objects with distinct, changeable materials such as smooth, rough, glossy, soft, grainy, metal, wood, glass, fabric, stone.} 

\textbf{Physical-Disaster/Event:} Introduces a transformative, chaotic, or disastrous event into the scene, even if not logical for the initial setting. \textit{Requirement: Use both natural and human-made events such as earthquake, flood, wildfire, hurricane, volcanic eruption, war, protest, explosion, building collapse, oil spill.} 

\textbf{Physical-POV/Composition:} Changes the camera angle, framing, or composition to alter the viewer's narrative relationship to the scene. 

\textbf{Physical-Pose:} Alters a subject's body language to convey a different emotion, status, or intent. \textit{Requirement: Must contain a 'person', 'animal', or anthropomorphic figure.} 

\textbf{Physical-Season:} Changes the time of year, including atmospheric conditions and indirect clues for indoor scenes. 


\textbf{Physical-Size:} Alters the scale of an exisiting object or subject in the image, without mentioning a very specific size. 

\textbf{Social-Culture:} Transports the subjects or scene into a different cultural context, affecting fashion, architecture, or historical period. \textit{Requirement: Best with people, fashion, and architecture that can be re-interpreted. Use different nationalities and cultures [e.g. Thai, Indian, Nigerian, Moroccan, Peruvian, Greek, Scottish, Ethiopian, Mexican, Maori, Mongolian, Egyptian, Brazilian, Norwegian].} 

\textbf{Social-Environment/Setting:} Changes the primary location to a new environment with a distinct character or narrative purpose. 

\textbf{Social-Genre/Narrative:} Re-contextualizes the scene to take place during a specific social gathering, ceremony, or ritual. \textit{Requirement: Use different events [e.g. wedding, birthday, festival, protest, parade, funeral, carnival, market, concert, sports event, religious ceremony]. It can apply a distinct visual philosophy or artistic movement or specific artistic style to the entire image.} 

\textbf{Social-Role:} Recasts a person into a new professional or archetypal role by altering their appearance and context. \textit{Requirement: Must contain a 'person'. Use different professions [e.g. firefighter, chef, astronaut, doctor, teacher, artist, musician, athlete, scientist, pilot, engineer, farmer, dancer, writer, actor].} 

\textbf{Social-Socio-Economic:} Adjusts the visual narrative of wealth, poverty, or societal status through changes to the environment and personal effects. \textit{Requirement: Use states such as homeless, wealthy, poor, middle-class, luxurious, impoverished, unemployed, white-collar, upper-class, working-class.} 





\textbf{Logical-Temporal:} Manipulates the timeline by showing the scene at a different point in its story (past, future, before, after). \textit{Requirement: The image must imply an action or process that can be shifted in time.} 

\textbf{Logical-CommonsenseGoal:} Creatively disrupts the scene's logical purpose or common-sense state. \textit{Requirement: The scene must have a clear function that can be completed or subverted, without mentioning how to do it. Do not mention how to change, focus on the goal.} 

\textbf{Logical-InsertionGoal:} Adds a general object or subject (something, an object, an equipment, etc) that has a clear, logical purpose within the scene. \textit{Requirement: Ask to add **something**, not a specific object, DO NOT mention what to add, the focus is on the goal. The goal has to relate to the overall scene purpose.}
\end{tcolorbox}

\section{Entity-Rubrics Evaluation}
\label{app:eval}

\paragraph{Evaluation Prompt Templates.} Our \eval\ evaluation employs a two-stage prompting strategy. The first call establishes a theoretical baseline of expectations from the context image alone (see Figure \ref{box:expectation_prompt}), while the second call performs a comprehensive comparison between the context and edited images against those expectations (Figure \ref{box:expectation_reality_prompt}). The output schema is presented in Figure \ref{box:eval_output_scheme}.

\begin{tcolorbox}[colback=gray!5,colframe=gray!50,title=Call 1: Entity Expectations Prompt,label={box:expectation_prompt}]
\footnotesize
\textbf{Role:} Expert Visual Analyst \\
\textbf{Inputs:} Instruction ($p$), Context Image ($\mathcal{I}_c$).

\textbf{Analysis Steps:}
\begin{itemize}
    \item \textbf{Step 1: Define Categories.} Identify \textit{Things} (distinct people/animals/objects; multiple instances must be differentiated with unique IDs and parts), \textit{Stuff} (general environmental and structural elements), and \textit{Global} (abstract image properties like luminance or spatial complexity).
    \item \textbf{Step 2: Generate Element List.} Create a comprehensive master list of all entities and parts present in $\mathcal{I}_c$ relevant to the edit.
    \item \textbf{Step 3: Per-Element Evaluation.} Assign each entity its group and evaluate its editing expectation:
    \begin{itemize}
        \item \texttt{EXPECTED\_CHANGE}: Highly expected to change. Obvious and necessary changes that everyone would expect (e.g., adding snow for a "winter scene").
        \item \texttt{OPTIONAL\_CHANGE}: May change to improve alignment without hurting identity preservation. Not everyone would think of it, but it is a positive creative interpretation (e.g., adding an umbrella for "make it rainy").
        \item \texttt{EXPECTED\_PRESERVATION}: Should be preserved and NOT change. Irrelevant to the instruction.
    \end{itemize}
\end{itemize}
\textbf{Output Format:} JSON object containing \texttt{entity\_expectations}.
\end{tcolorbox}


\begin{tcolorbox}[colback=gray!5,colframe=gray!50,title=Call 2: Comprehensive Evaluation Prompt, label={box:expectation_reality_prompt}]
\footnotesize
\textbf{Role:} Micro-Detail Observer and Expert Visual Analyst \\
\textbf{Inputs:} Instruction ($p$), Context Image ($\mathcal{I}_c$), Edited Image ($\mathcal{I}_e$), Entity Expectations.

\textbf{Analysis Steps:}
\begin{itemize}
    \item \textbf{Steps 1 \& 2: Define \& Generate List.} Expand the master list by scanning both images to include new elements introduced in $\mathcal{I}_e$.
    \item \textbf{Step 3: Per-Element Evaluation.} For each item, determine:
    \begin{itemize}
        \item \textit{Group \& Change Description:} Describe the literal visual delta between the images.
        \item \textit{Edit Action:} Categorize the change type (e.g., \texttt{POSITION}, \texttt{STYLE\_TRANSFER}).
        \item \textit{Reality vs. Expectation:} Compare actual change to the baseline expectation:
        \begin{itemize}
            \item \texttt{GOOD\_EXPECTED\_CHANGE}: Changed, and needed to. Explicitly requested or mandatory.
            \item \texttt{BAD\_EXPECTED\_CHANGE}: Changed and needed to, but actively hurts alignment (failure of core intent).
            \item \texttt{GOOD\_OPTIONAL\_CHANGE}: Changed without being obvious, but actively improves alignment (positive creative interpretation).
            \item \texttt{BAD\_OPTIONAL\_CHANGE}: Changed without being obvious, but actively hurts alignment or preservation (negative creative interpretation).
            \item \texttt{GOOD\_EXPECTED\_PRESERVATION}: Did not change, and this is the correct outcome.
            \item \texttt{BAD\_EXPECTED\_PRESERVATION}: Did not change, but should have to successfully fulfill the instruction.
        \end{itemize}
    \end{itemize}
    \item \textbf{Step 4: Global Assessment.} Evaluate missing changes, over-editing, and overall narrative coherence.
    \item \textbf{Step 5: Overall Score Synthesis.} Synthesize an overall 1--10 alignment score:
    \begin{itemize}
        \item \textbf{10 (Perfect Alignment):} Perfectly embodies the instruction with all required changes executed flawlessly. Essence is preserved; narrative is cohesive.
        \item \textbf{8--9 (Strong Alignment):} Strongly reflects the instruction with most required changes executed well. Minor imperfections exist but do not detract overall.
        \item \textbf{6--7 (Moderate-High Alignment):} Good level of alignment. Most required changes executed, but some minor imperfections or missed requirements exist.
        \item \textbf{4--5 (Moderate Alignment):} Some required changes executed, but noticeable imperfections, missed requirements, or signs of over-editing. Disjointed narrative.
        \item \textbf{2--3 (Low Alignment):} Many required changes missing; significantly different from what was requested. Substantial over-editing or preservation issues.
        \item \textbf{1 (Very Low Alignment):} Fails to align. Required changes mostly missing; output unrelated to instruction. Incoherent narrative and severe preservation issues.
    \end{itemize}
\end{itemize}
\textbf{Output Format:} JSON containing \texttt{entity\_evaluations} and \texttt{global\_evaluations}.
\end{tcolorbox}

\begin{tcolorbox}[
    colback=blue!5!white,
    colframe=blue!75!black,
    title=\textbf{Entity-Rubrics Evaluation: Output Schema},
    label={box:eval_output_scheme},
    fonttitle=\sffamily\bfseries
]
\begin{lstlisting}[language=TeX, basicstyle=\footnotesize\ttfamily]
{
  "entity_evaluations": {
    "<entity_name>": {
      "group": "things" | "stuff" | "global",
      "change_description": "string",
      "change_occured": boolean,
      "edit_action": "NO_CHANGE" | "OBJECT_PRESENCE" | 
                     "OBJECT_COUNT" | "POSITION" | 
                     "TRANSFORM" | "POSE" | "VIEWPOINT" | 
                     "COLOR" | "TEXTURE" | "LIGHTING" | 
                     "STYLE_TRANSFER" | "ATTRIBUTE" | 
                     "STATE" | "OTHER",
       "edit_expectation": "EXPECTED_CHANGE" | 
                            "OPTIONAL_CHANGE" | 
                            "EXPECTED_PRESERVATION"
      "edit_execution": "GOOD_EXPECTED_CHANGE" | 
                        "BAD_EXPECTED_CHANGE" | ...,
      "entity_edit_rationale": "string",
      "entity_overall_score": integer [1-10],
    }
  },
  "global_evaluation": {
    "missing_changes": boolean,
    "over_editing": boolean,
    "overall_narrative_coherence": boolean,
    "final_rank": integer [1-10],
    "final_rationale": "string"
  }
}
\end{lstlisting}
\end{tcolorbox}

\paragraph{Entities Granularity.} When defining entities, whether they are things, stuff (i.e., background elements like sky, floor, or walls), or global features, there is a significant hierarchy of resolution to consider. For instance, a ``dog'' can be further decomposed into its mouth, ears, and paws, extending even further to specific attributes like the ``left canine'' or the ``tongue''. The depth of granularity required is highly context-dependent, as not every resolution is relevant to a specific edit. While an instruction for a man to move from "standing" to ``sitting'' may not require detailing individual body parts for instruction-following alignment, an edit where a man's mood changes from sad to happy necessitates explicit references to the mouth or eyes. Consequently, both our dataset curation for explicit instruction generation and our evaluation framework prioritize high density and granularity. As presented in the prompt \ref{box:instruction_gen}, we define Granularity as the requirement to relate to relevant entity components, such as specifying edits for a person's clothing, hair, and posture in separate sentences, and we define Density as the maintenance of a high concentration of edits by ensuring each sentence contains exactly one technical change, allowing for multiple sentences to be generated for the same entity as long as they refer to distinct components or different aspects of the transformation.

\section{Human Evaluation}
\label{app:human_evaluation}

\begin{figure}[!htbp]
    \centering
    \includegraphics[width=1\linewidth]{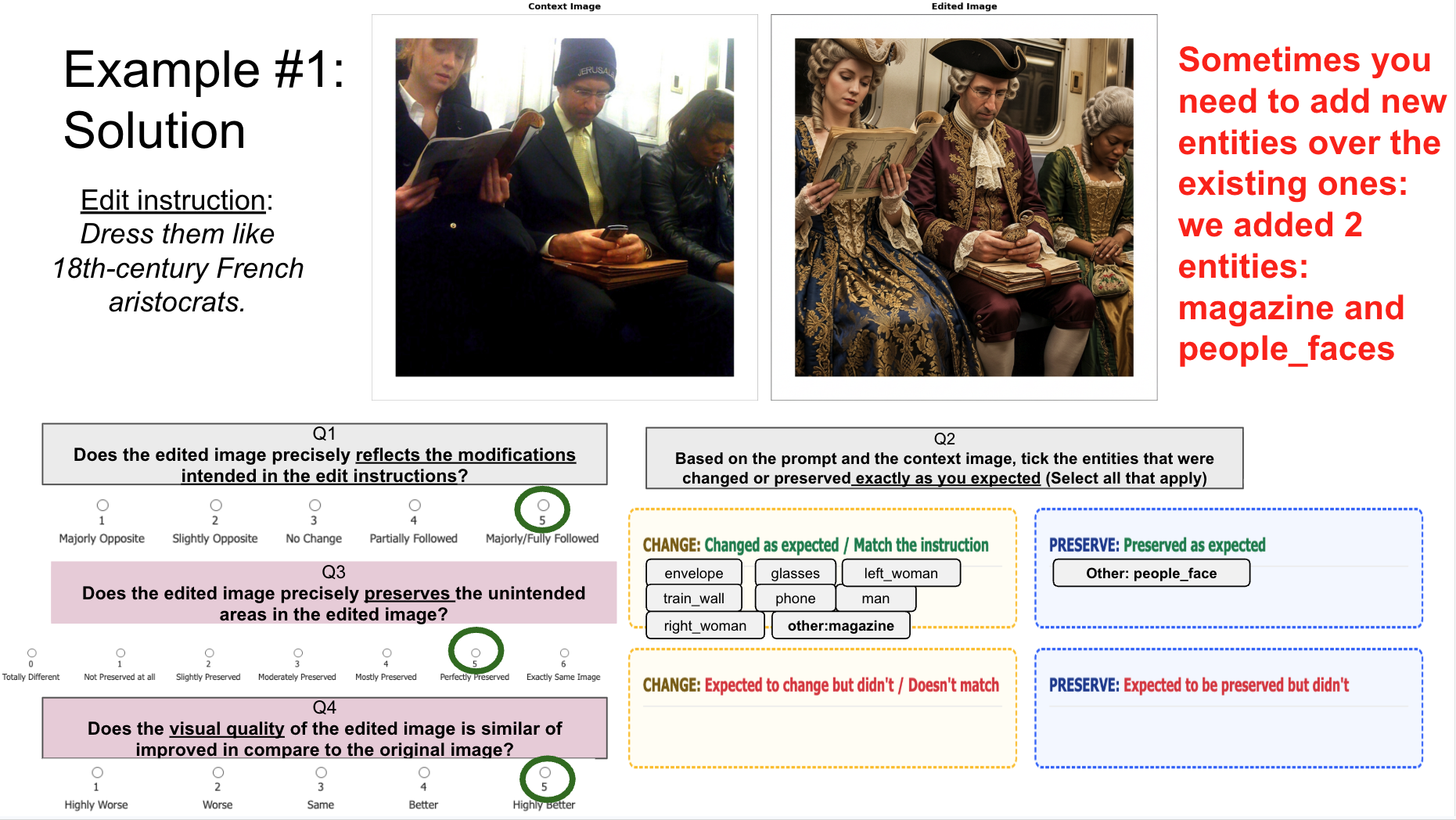}
    \caption{Annotation guideline example for Amazon Mechanical Turk workers.}
    \label{fig:human_guidelines}
\end{figure}

\begin{figure}[!htbp]
    \centering
    \includegraphics[width=0.7\linewidth]{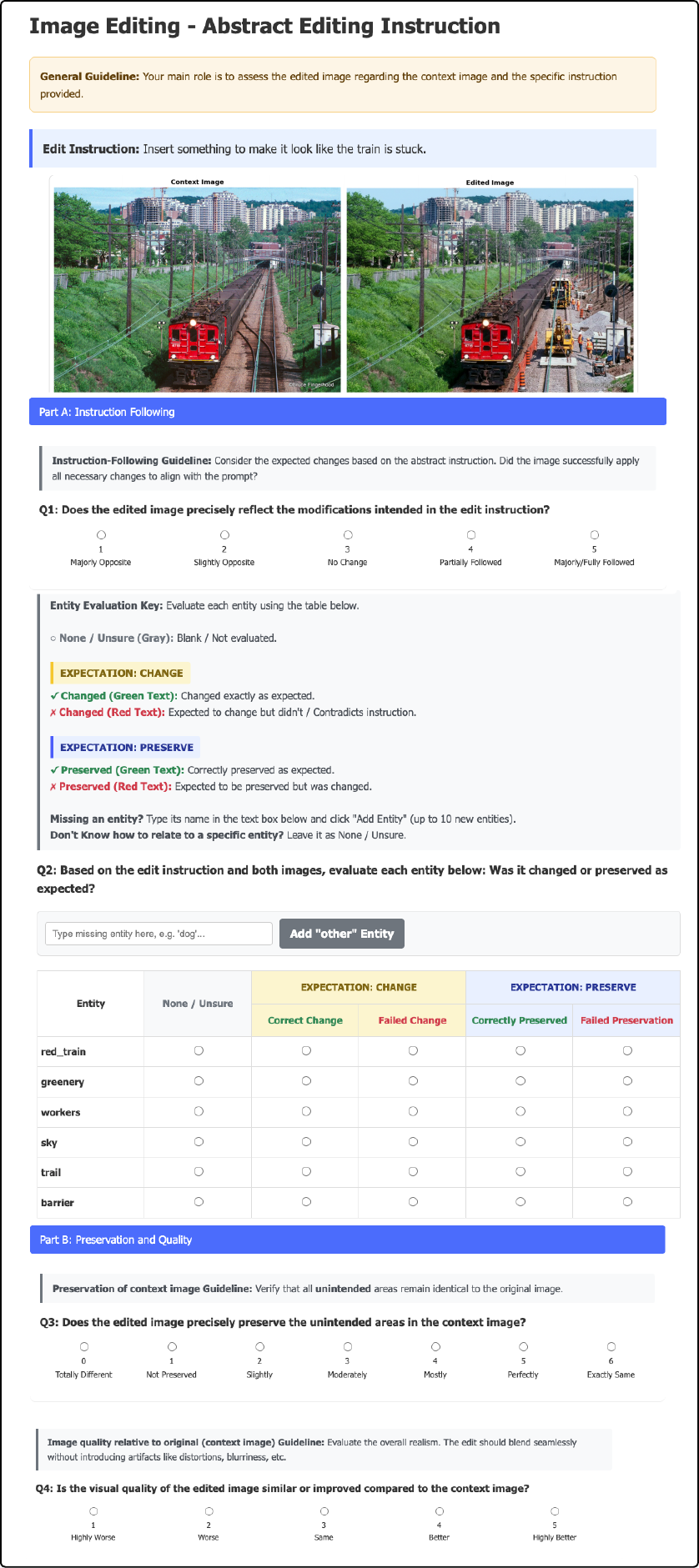}
    \caption{Human Evaluation Questionnaire in Amazon Mechanical Turk (AMT).}
    \label{fig:human_eval_mturk}
\end{figure}

\paragraph{Human Evaluation Questionnaire and Guidelines}
To rigorously validate our automated evaluation \eval, and assess image editing model performance, we conducted a comprehensive human evaluation study using Amazon Mechanical Turk (AMT). To ensure high-quality and reliable annotations, we restricted the pool to fluent English speakers who successfully passed a qualification task. Ultimately, a curated pool of eight highly qualified workers participated in the study, with each specific editing task being evaluated by three independent annotators to establish consensus. Workers were compensated fairly at a rate of 0.35\$ per completed task (which takes approximately 1-2 minutes). As illustrated in Figure \ref{fig:human_eval_mturk}, the evaluation interface presents the context and edited images side-by-side alongside a structured, two-part questionnaire. Part A focuses on Instruction Following, requiring annotators to rate the overall prompt alignment on a 1–5 scale (Q1) and conduct a fine-grained, entity-level evaluation (Q2) to verify if specific elements were correctly changed or preserved. Part B assesses Preservation and Quality, measuring how accurately unintended areas were maintained (Q3) and scoring the overall realism and visual quality compared to the original image (Q4). To standardize the assessment and minimize subjective bias, workers were provided with detailed visual guidelines and solved examples. The study achieved moderate Inter-Annotator Agreement (quadratic weighted Fleiss' $\kappa$ of \mv{$0.47$}\footnote{IAA breakdown: instruction following $0.54$, image preservation $0.61$, per-entity alignment $0.41$, and image quality $0.28$.}). As demonstrated in Figure \ref{fig:human_guidelines}, these examples explicitly trained annotators on how to correctly assign scores and how to manually add and evaluate new or missing entities (such as adding "magazine" or "people faces") that frequently emerge during complex, abstract image edits.

\paragraph{Human Evaluation and Ethics} This study conducted image annotation tasks involving human crowd-workers. The protocol was determined to be exempt from formal IRB review based on the following criteria: (1) the research involved minimal risk to participants, (2) the tasks did not expose workers to sensitive, offensive, or harmful content, and (3) no Personally Identifiable Information (PII) was collected or stored, ensuring participant anonymity.

\clearpage

\section{Experimental Setup}
\label{app:exp_setup}

\paragraph{Image Editing Models.} We primarily evaluate \mv{11} state-of-the-art image editing models, selected to span a diverse range of parameter scales and architectural designs (see Table \ref{tab:baselines_arch}). The open-source models include FLUX.1-Kontext [dev]\citep{labs2025flux}, FLUX.2 [dev] \citep{bfl2025representation}, Qwen-Image-Edit-2511 \citep{wu2025qwen, wu2025qwenimagetechnicalreport}, Step1X-Edit \citep{liu2025step1x}, Bagel \citep{deng2025emerging}, and HiDream-E1-1 \citep{cai2025hidream}. For Step1X and Bagel, we additionally evaluate their respective thinking variants ("think-and-reflect" and "think" modes correspondingly). The evaluated closed-source models consist of GPT-Image-1.5 \citep{openai2025gptimage}, Seedream-4.5 \citep{seedream2509seedream} and the Gemini Family; Gemini-3 Pro Image Preview (Nano Banana), Gemini 3.1 Flash Image Preview \citep{gemini_image_api} (Nano Banana 2) and Gemini-2.5 Flash Image \citep{comanici2025gemini}. We evaluated all models using their default inference parameters (see Table \ref{tab:model_configs_balanced}).

\paragraph{Instruction Following Evaluation Metrics.}
We examine three commonly adopted instruction-following metrics alongside our proposed evaluation \eval. First, $\Delta$CLIP computes the difference in cosine similarity between the abstract instruction and the edited image versus the context image; this distance measures the net change in image-text alignment. For the second and third methods, we utilize VLM-based approaches that prompt a VLM with the instruction, the context image, and the candidate edited image to generate a global instruction-following score and a brief explanation. Specifically, VIE-Score \citep{ku2024viescore} incorporates penalties for over-editing within its system prompt, while ComplexEdit \citep{yang2025texttt} utilizes discrete per-score descriptions to guide the model. For all VLM-based metrics, including \eval, we employ Gemini 3 Flash with a thinking budget as the backbone evaluator. This ensures a high-reasoning, consistent baseline for scoring, with each prompt instructing the VLM to conclude with a single, finalized score.

\paragraph{Preservation Metrics.} Alongside the instruction-following evaluation metrics, we measure semantic preservation of the context image, identity preservation as well as image quality based on VLM \citep{yang2025texttt} and perceptual similarity to the context image (LPIPS \citep{zhang2018unreasonable}).

\begin{table*}[ht]
    \centering
    \footnotesize 
    \caption{Architecture Summary of Baseline Image Editing Models. DiT notes Diffusion-Transformer. \ding{51} and \ding{55} denote the presence or absence of a dedicated thinking mode, respectively.}
    \label{tab:baselines_arch}
    
    \resizebox{\textwidth}{!}{
    \begin{NiceTabular}{l l l l l l c}
        \toprule
        \textbf{Category} & \textbf{Model} & \textbf{Company} & \textbf{Text-Encoder} & \textbf{Image Generator} & \textbf{Size/Params} & \textbf{Thinking} \\
        \midrule
        \Block{6-1}{Open-source} 
                                   & Flux-Kontext [dev] & Black Forest Labs & CLIP + T5 & DiT & $\sim$12B & \ding{55} \\
                                   & Flux2 [dev] & Black Forest Labs & Mistral + T5 & DiT & $\sim$32B & \ding{55} \\
                                   & Qwen-Image-Edit & Alibaba Cloud & Qwen-VL-2.5 & DiT & 20B & \ding{55} \\
                                   & Step1X-Edit & Stepfun & Qwen-VL & DiT & $\sim$30B & \ding{51} \\
                                   & HiDream-E1-1 & HiDream.ai & T5 + CLIPs + Llama & DiT & $\sim$25B & \ding{55} \\
                                   & Bagel & ByteDance & Bagel-MoT & Custom Unified & 14B (7B active) & \ding{51} \\
        \midrule
        \Block{5-1}{Closed-source} 
                                   & Gemini 2.5 Flash & Google & X & X & - & \ding{55} \\
                                   & Gemini 3 Pro & Google & X & X & - & \ding{55} \\
                                   & Gemini 3.1 Flash & Google & X & X & - & \ding{55} \\
                                   & GPT-Image-1.5 & OpenAI & X & X & - & \ding{55} \\
                                   & Seedream-4.5 & ByteDance & X & X & - & \ding{55} \\
        \bottomrule
    \end{NiceTabular}%
    } 
\end{table*}

\begin{table}[h]
\centering
\scriptsize
\setlength{\tabcolsep}{3pt} 
\caption{\textbf{Model Inference Configurations.}}
\label{tab:model_configs_balanced}
\begin{tabular}{@{} l c c p{1.8cm} c c c @{}} 
\toprule
\textbf{Model} & \textbf{Category} & \textbf{\makecell{Inference\\Steps}} & \textbf{\makecell{CFG\\Scale}} & \textbf{Resolution} & \textbf{Thinking} & \textbf{Seed} \\
\midrule
FLUX.2 & Open & 20 & 2.5 & 1024 & \xmark & 42 \\
FLUX.1-Kontext & Open & 20 & 2.5 & 1024 & \xmark & 42 \\
Qwen-image-edit & Open & 40 & 4.0 \tiny(Tr) & --- & \xmark & 0 \\
Step1x-edit & Open & 28 & 6.0 \tiny(Tr) & --- & \xmark & 42 \\
Step1x-edit-think & Open & 28 & 6.0 \tiny(Tr) & --- & \checkmark & 42 \\
Bagel & Open & 50 & \makecell[l]{4.0 \tiny(Tx)\\2.0 \tiny(Im)} & --- & \xmark & 42 \\
Bagel-think & Open & 50 & \makecell[l]{4.0 \tiny(Tx)\\2.0 \tiny(Im)} & --- & \checkmark & 42 \\
HiDream-e1 & Open & 28 & 1.5 \tiny(Im) & --- & \xmark & 42 \\
\midrule
Gemini (All) & Closed & --- & --- & --- & \xmark & 42 \\
GPT Image 1.5 & Closed & --- & --- & --- & \xmark & --- \\
Seedream-4.5 & Closed & --- & --- & 2048 & \xmark & --- \\
\bottomrule
\end{tabular}
\end{table}

\section{Complementary Results and Analysis}
\label{app:results}

\begin{figure}[!htbp]
    \centering
    \caption{Performance Comparison of Image Editing Models. \textbf{Instruction Following:} Ent-R (Entity-Rubrics), VIE (VIE Score) \citep{ku2024viescore}, CE (Complex-Edit) \citep{yang2025texttt}, $\Delta$C ($\Delta$CLIP) \citep{radford2021learning} Scores presented visually as well on the right.}
    \label{fig:metrics}
    \begin{minipage}{0.48\textwidth}
        \centering
        \scriptsize 
        \setlength{\tabcolsep}{2pt} 
        \begin{NiceTabular}{l | l | c c c c}[cell-space-limits=2pt]
            \toprule
            \textbf{Cat.} & \textbf{Model} & \textbf{Ent-R.} & \textbf{VIE} & \textbf{CE} & \textbf{$\Delta$C} \\
            \midrule
            \Block{8-1}{Open} 
                & Qwen-image-edit   & \textbf{7.48} & 6.00 & 8.43 & 0.05 \\
                & FLUX.2            & 7.26 & 5.13 & 8.36 & 0.05 \\
                & Hidream-e1        & 5.38 & 3.53 & 6.60 & 0.05 \\
                & FLUX.1-Kontext    & 5.10 & 4.09 & 5.87 & 0.03 \\

                & Step1x-edit-thk   & 6.9 & 6.24 & 7.34 & 0.03 \\
                & Step1x-edit       & 6.55 & 6.08 & 7.00 & 0.03 \\
                & Bagel-think       & 5.80 & 4.56 & 6.61 & 0.04 \\
                & Bagel             & 4.45 & 3.91 & 5.55 & 0.03 \\
            \midrule
            \Block{5-1}{Closed} 
                & Gem. 3.1 (Fl.)    & \textbf{9.52} & 8.17 & 9.78 & 0.05 \\
                & GPT Image 1.5     & 9.34 & 7.70 & 9.88 & 0.05 \\
                & Gem. 3 (Pro)      & 9.27 & 7.88 & 9.56 & 0.05 \\
                & Seedream-4.5      & 9.21 & 7.91 & 9.41 & 0.04 \\
                & Gem. 2.5 (Fl.)    & 8.67 & 6.82 & 8.46 & 0.04 \\
            \bottomrule
        \end{NiceTabular}
    \end{minipage}
    \hfill 
    \begin{minipage}{0.48\textwidth}
        \centering
        \includegraphics[width=\linewidth]{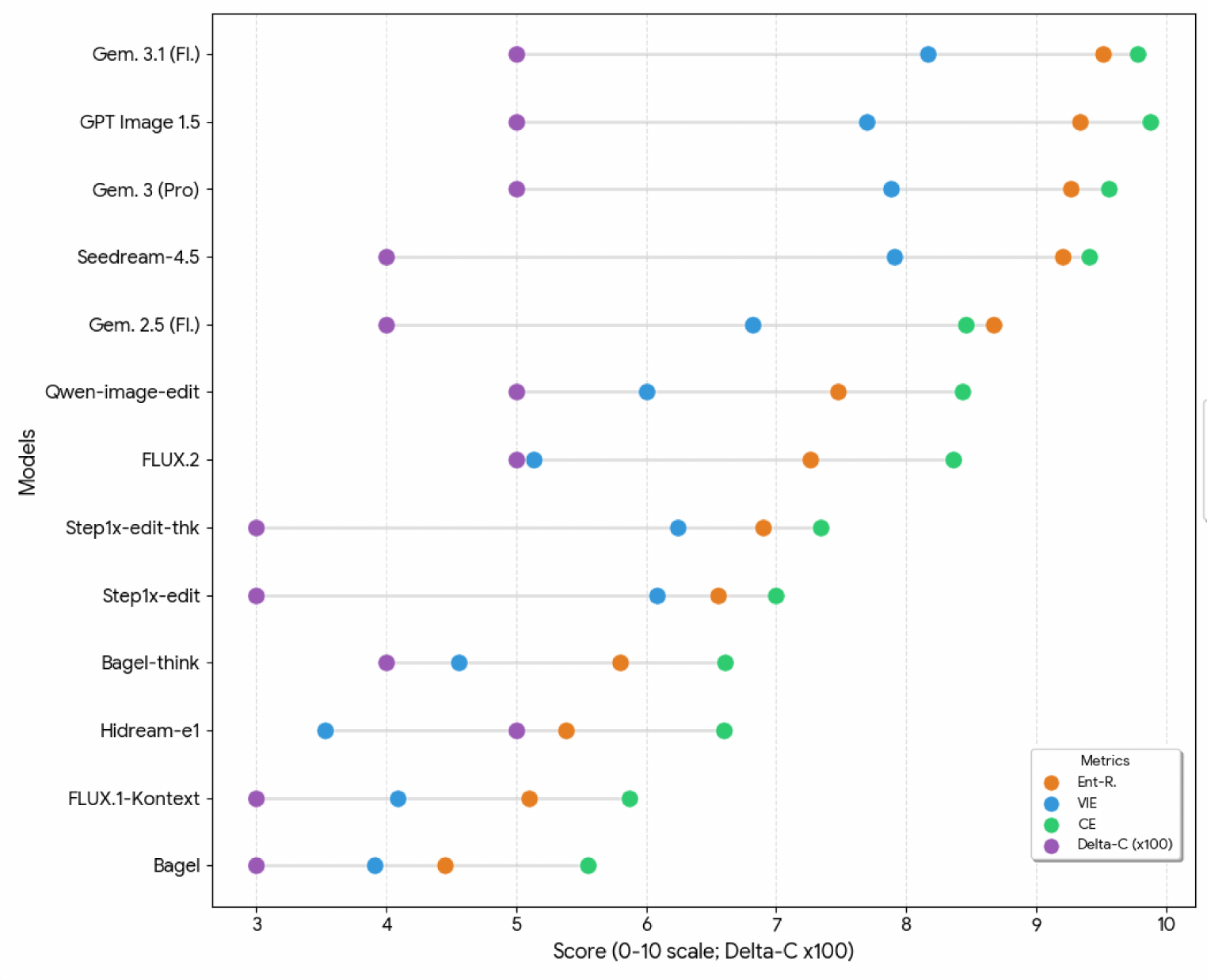}
    \end{minipage}
\end{figure}

\begin{figure}[!ht]
    \centering
    \includegraphics[width=0.5\linewidth]{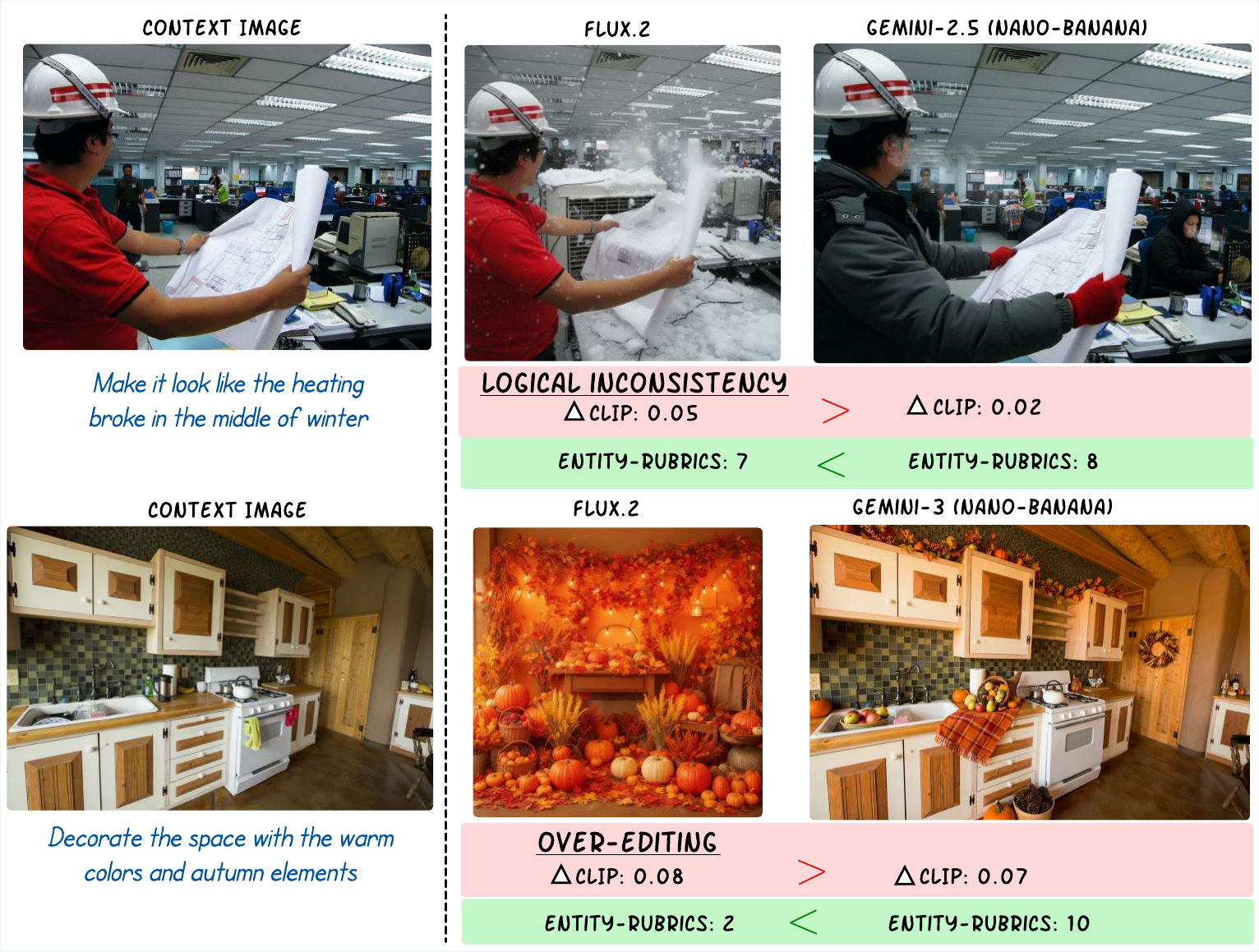}
    \caption{$\Delta$CLIP vs \eval\ Evaluation Qualitative Comparison (Samples from the \dataset\ test set. Logical Inconsistency (top): While the prompt asked for the heating to be "broken," FLUX generated a literal heating generator and snow that didn't exist in the original office, whereas the Gemini 2.5 model correctly inferred that a broken heater means people would wear coats and gloves. CLIP favors the literal alignment.
Over-Editing (bottom): When tasked with decorating a kitchen, FLUX.2 completely replaces the room with a dense pile of pumpkins and leaves. CLIP ignores the contextual anchor.}
    \label{fig:clip_is_bad}
\end{figure}

\paragraph{Discrepancies in evaluation metrics.} An analysis of the global results reveals significant limitations in existing automated metrics. 
We examined CLIP \citep{radford2021learning}, encoding-based metric, where we compute the difference between the CLIP score of the edited image with the instruction to the context image with the instruction. Moreover, we examine two commonly adopted VLM-based approaches, VIE-Score \citep{ku2024viescore} and ComplexEdit \citep{yang2025texttt} of assessing instruction-following. As presented in Table/Figure \ref{fig:metrics}, the $\Delta$CLIP metric exhibits minimal variance between models, proving uninformative and incapable of capturing subtle, fine-grained semantic changes. It highly favors logical inconsistencies and over-editing, as presented in Figure \ref{fig:clip_is_bad}. Conversely, while the VLM-based metrics are highly correlated, ComplexEdit demonstrates a strong leniency bias, consistently yielding inflated scores that fail to strictly separate high-quality edits from mediocre ones. While the VIE Score accounts for structural preservation, it frequently over-penalizes legitimate global changes and lacks interpretable, detailed explanations for its scoring. These discrepancies highlight the necessity of our \eval\ evaluation, which provides a more balanced, diagnostically transparent evaluation.

\paragraph{Context Preservation and Image Quality.}
Table \ref{tab:preservation_results} details the preservation and perceptual quality metrics across the evaluated models. To establish a human baseline, we report the human preservation score (\textbf{Hum-P}), which represents the mean rating from our user questionnaire multiplied by two to align with a standardized scale. Closed-source models, such as Gemini 3.1 Flash and Seed-Dream, demonstrate a balanced performance, maintaining strong identity preservation and high perceptual quality while executing complex abstract edits. Interestingly, the most aggressive preservation scores, such as the exceptionally low LPIPS distances observed in models like Step1X and Bagel, require careful interpretation. Qualitative analysis reveals that these seemingly superior preservation results are frequently an artifact of severe under-editing. Because these models often struggle to resolve the semantic ambiguity of abstract instructions, they fail to modify the image meaningfully. Consequently, they leave the original context largely untouched, yielding artificially inflated preservation metrics that mask their failure to follow the underlying instruction.

\begin{table}[!ht]
    \centering
    \caption{\eval\ Preservation and Image Quality. \textbf{Hum-P:} Human ground truth for preservation. \textbf{LP:} LPIPS $\downarrow$, \textbf{Id:} Identity preservation $\uparrow$, \textbf{Qual:} Perceptual Quality $\uparrow$.}
    \label{tab:preservation_results}
    \scriptsize
    \setlength{\tabcolsep}{3pt} 
    \renewcommand{\arraystretch}{1.5} 
    \begin{NiceTabular}{l l | c | c c c }
        \toprule
        \textbf{Cat.} & \textbf{Model} & \textbf{Hum-P} $\uparrow$ & \textbf{LP} $\downarrow$ & \textbf{Id} $\uparrow$ & \textbf{Qual} $\uparrow$ \\
        \midrule
        \Block{4-1}{\rotatebox{90}{\makecell{Open\\Source}}} 
            & Qwen-Image-Edit            & -    & 0.50 \sd{0.22} & 5.89 \sd{2.87} & 5.11 \sd{1.72} \\
            & FLUX.2 [dev]               & 7.17 & 0.67 \sd{0.15} & 5.51 \sd{3.26} & 5.95 \sd{1.77} \\
            & HiDream-E1                 & -    & 0.63 \sd{0.19} & 4.17 \sd{3.01} & 4.85 \sd{1.57} \\
            & FLUX.1-Kontext [dev]       & 7.64 & 0.63 \sd{0.17} & 6.89 \sd{3.02} & 5.74 \sd{1.73} \\
        \midrule
        \Block{4-1}{\rotatebox{90}{\makecell{OS w/\\Thinking}}} 
            & Step1X-Think-Reflect       & 8.65 & 0.35 \sd{0.19} & 8.38 \sd{2.24} & 5.77 \sd{1.76} \\
            & Step1X                     & -    & 0.33 \sd{0.19} & 8.52 \sd{2.07} & 5.77 \sd{1.70} \\
            & Bagel-Think                & -    & -             & -             & -             \\
            & Bagel                      & 7.09 & 0.30 \sd{0.30} & 7.43 \sd{3.24} & 5.75 \sd{2.42} \\
        \midrule
        \Block{5-1}{\rotatebox{90}{\makecell{Closed\\Source}}} 
            & Gemini 3.1 Flash (NB 2)    & 7.89 & 0.51 \sd{0.24} & 8.17 \sd{2.24} & 6.84 \sd{1.92} \\
            & GPT-Image                  & 7.07 & 0.64 \sd{0.14} & 7.39 \sd{2.39} & 7.00 \sd{1.65} \\
            & Gemini 3 Pro (NB)          & -    & 0.45 \sd{0.26} & 8.25 \sd{2.43} & 6.63 \sd{1.93} \\
            & Seed-Dream                 & -    & 0.46 \sd{0.25} & 8.21 \sd{2.35} & 7.29 \sd{1.79} \\
            & Gemini 2.5 Flash Image     & -    & 0.47 \sd{0.22} & 8.19 \sd{2.44} & 7.04 \sd{1.75} \\
        \bottomrule
    \end{NiceTabular}
\end{table}

\begin{figure}[!ht]
    \centering
    \includegraphics[width=0.75\linewidth]{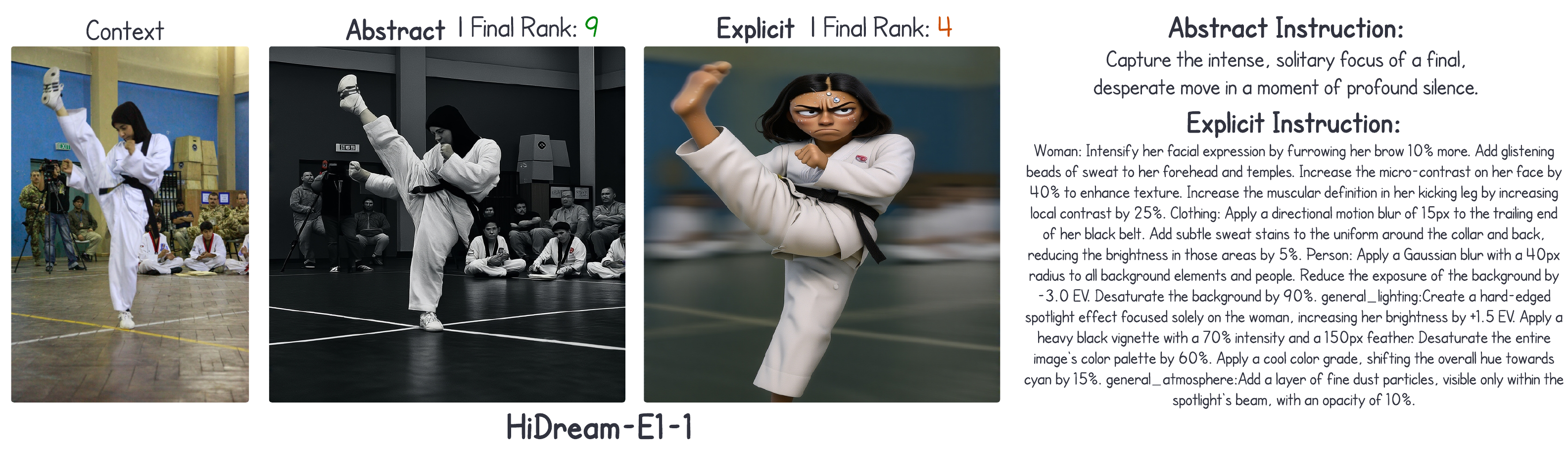}
    \caption{A qualitative example demonstrating over-editing by HiDream-E1 when given an explicit instruction.}
    \label{fig:hidream_overediting}
\end{figure}

\paragraph{Text Clues as Abstractness Bypass in Closed-Source Models.}
We find that closed-source models insert entities with text/sign/cue in their names at a 52\% higher rate than open-source models (2.27\% vs 1.49\% of inserted entities), suggesting they are more prone to increase the alignment to the abstract prompt via text-related visual elements, bypassing challenging visual edits (see examples in Appendix Figure \ref{fig:text_entities}).

\begin{figure}[!ht]
    \centering
    \includegraphics[width=0.95\linewidth]{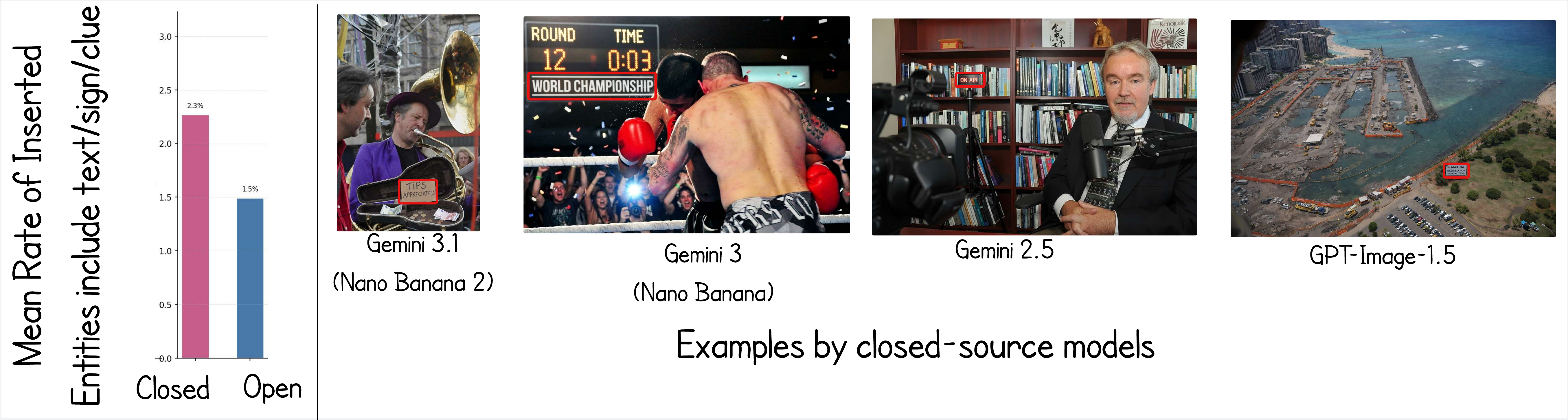}
    \caption{Qualitative Examples across Closed Source models of cases where they insert more textual clues to align better with the abstract instruction.}
    \label{fig:text_entities}
\end{figure}

\paragraph{Complementary Results.} 
Table \ref{tab:explicit} presents the \eval\ scores on the explicit prompt, as presented in bar graphs in the main paper. 

Figure \ref{fig:vendi} presents the violin distributions of Vendi scores for image representations (DINOv3 \citep{simeoni2025dinov3}) generated from both abstract and explicit prompt types across 10 seeds each. We compare two models, FLUX.2 and Gemini 3.1, and provide qualitative examples from FLUX.2 illustrating the limited diversity of explicit prompts relative to the richness accessed through abstract ones. Notably, this diversity gap is more pronounced in closed-source model. 

\begin{figure}[!ht]
    \centering
    \includegraphics[width=0.95\linewidth]{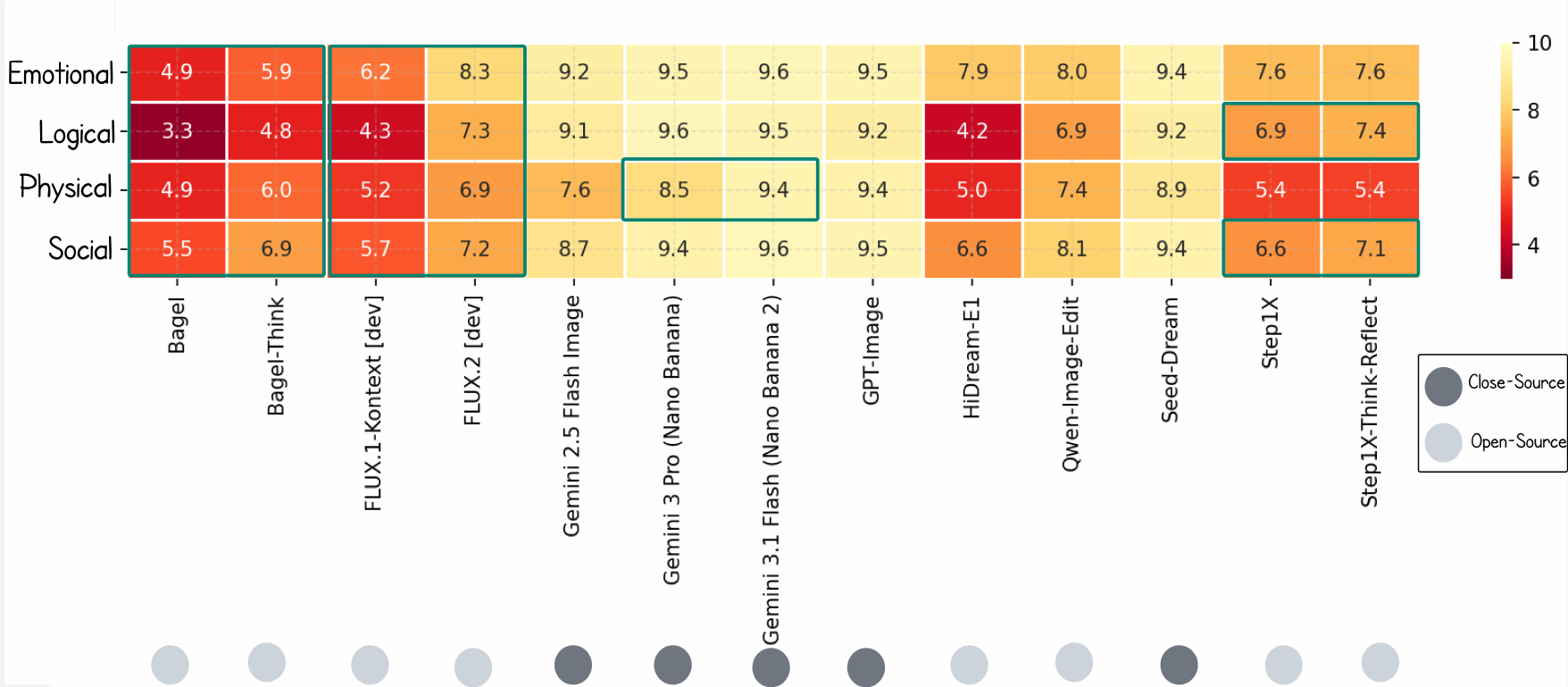}
    \caption{Heatmap Visualization of the \eval\ scores per model (x-axis) and per domain (y-axis).}
    \label{fig:domains_models}
\end{figure}

\begin{figure}[!ht]
    \centering
    \includegraphics[width=0.75\linewidth]{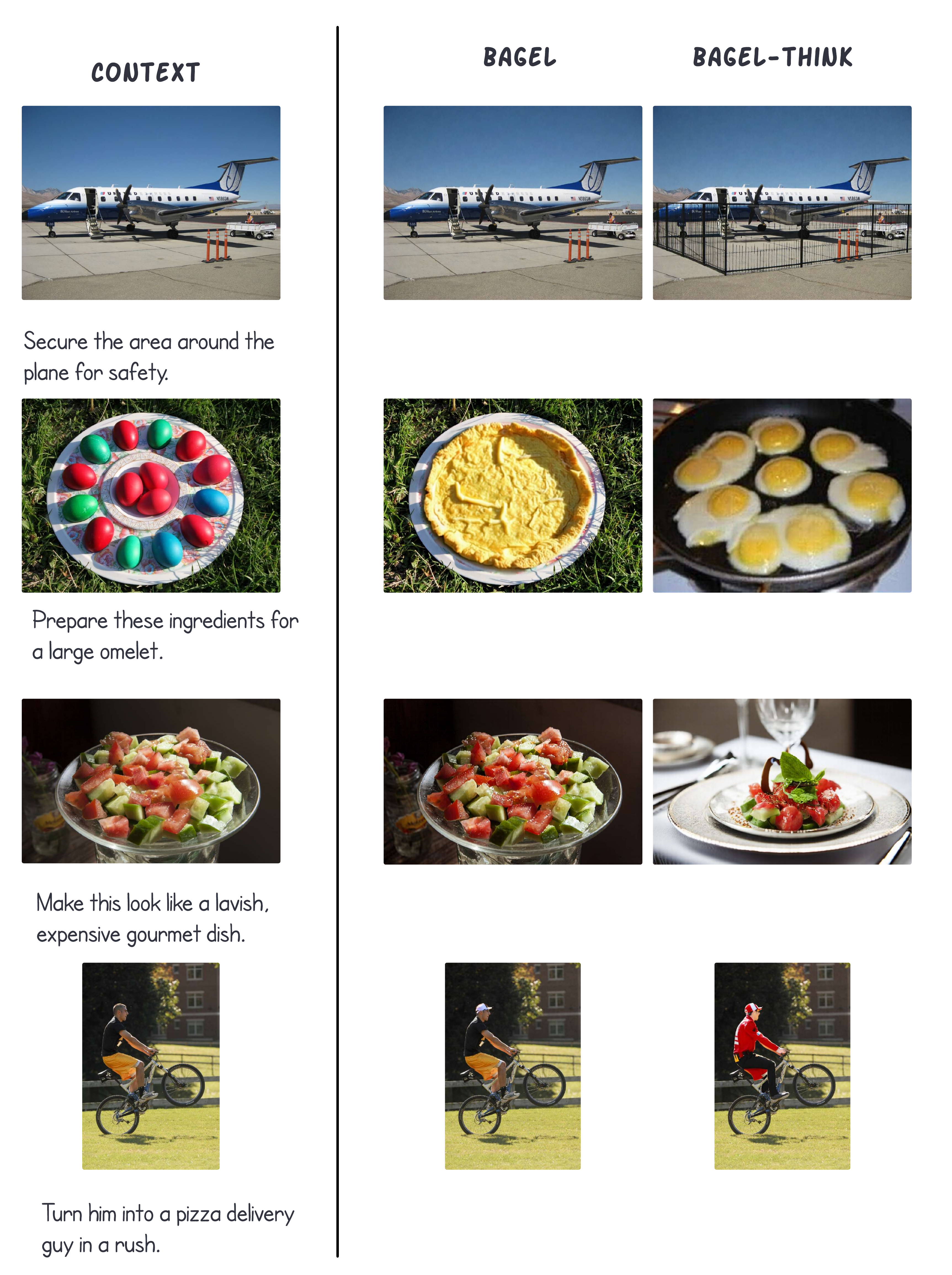}
    \caption{A qualitative comparison of Bagel vs. Bagel-Think.}
    \label{fig:qualitative_bagel_thinking}
\end{figure}

\begin{figure}[!ht]
    \centering
    \includegraphics[width=0.75\linewidth]{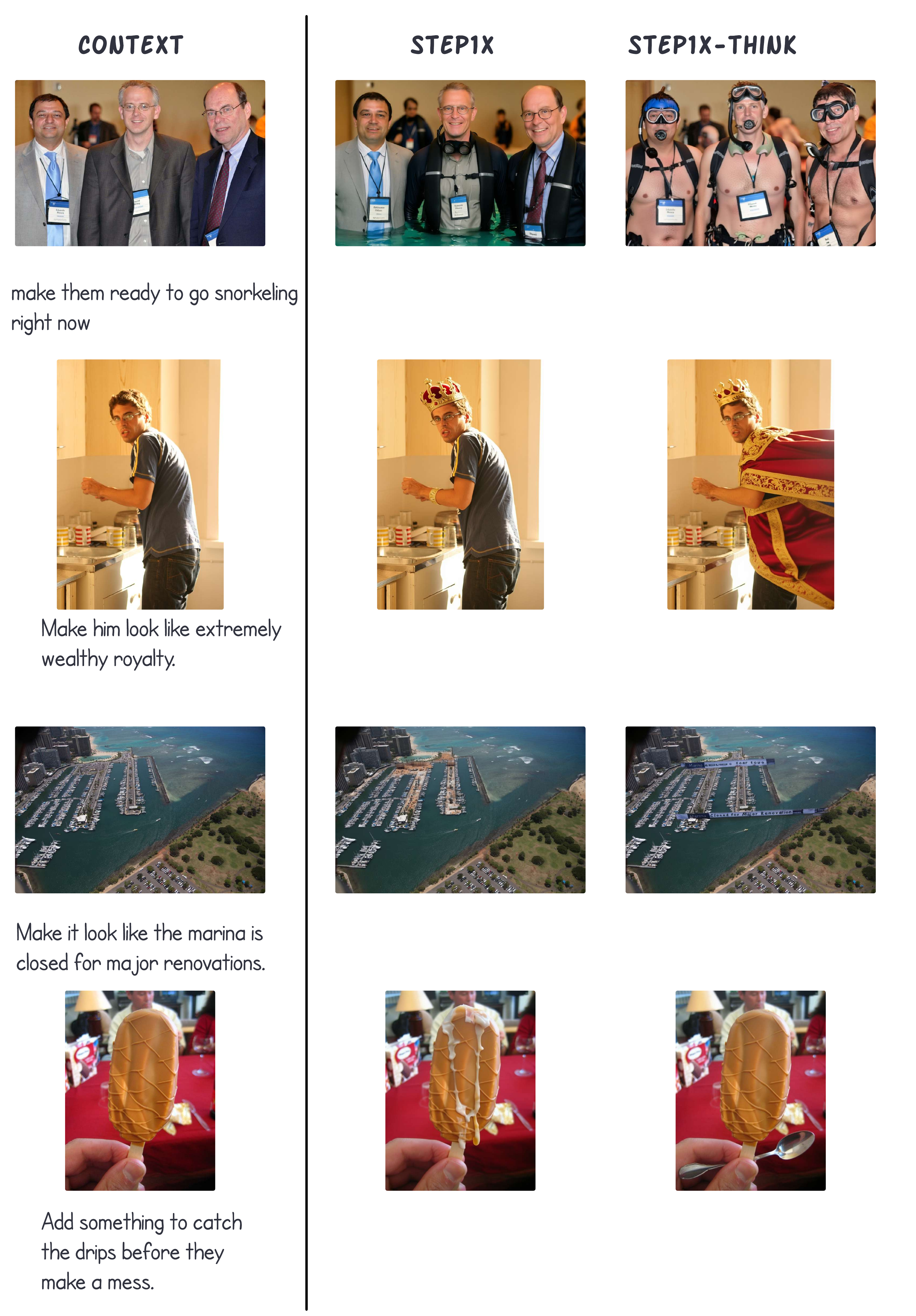}
    \caption{A qualitative comparison of Step1X vs. Step1X-Think-Reflect.}
    \label{fig:qualitative_step1_thinking}
\end{figure}

Figure \ref{fig:domains_models} presents a heatmap of \eval\ scores decomposed by domain. The results highlight that integrating a "thinking" mode yields notable performance improvements (e.g., from Bagel to Bagel-Think, and Step1X to Step1X-Think). For Step1X, the most substantial gains are observed in the logical and social domains, suggesting that reasoning capabilities help the model decompose complex abstract intents into aligned, explicit sub-instructions. The gains are presented qualitatively as well in Figures \ref{fig:qualitative_bagel_thinking}, \ref{fig:qualitative_step1_thinking}.
Furthermore, while the performance of closed-source models remains relatively stable overall, we observe a significant leap from Gemini 3 Pro to Gemini 3.1 Flash specifically within the physical domain.

\begin{figure}[!ht]
    \centering
    \includegraphics[width=0.9\linewidth]{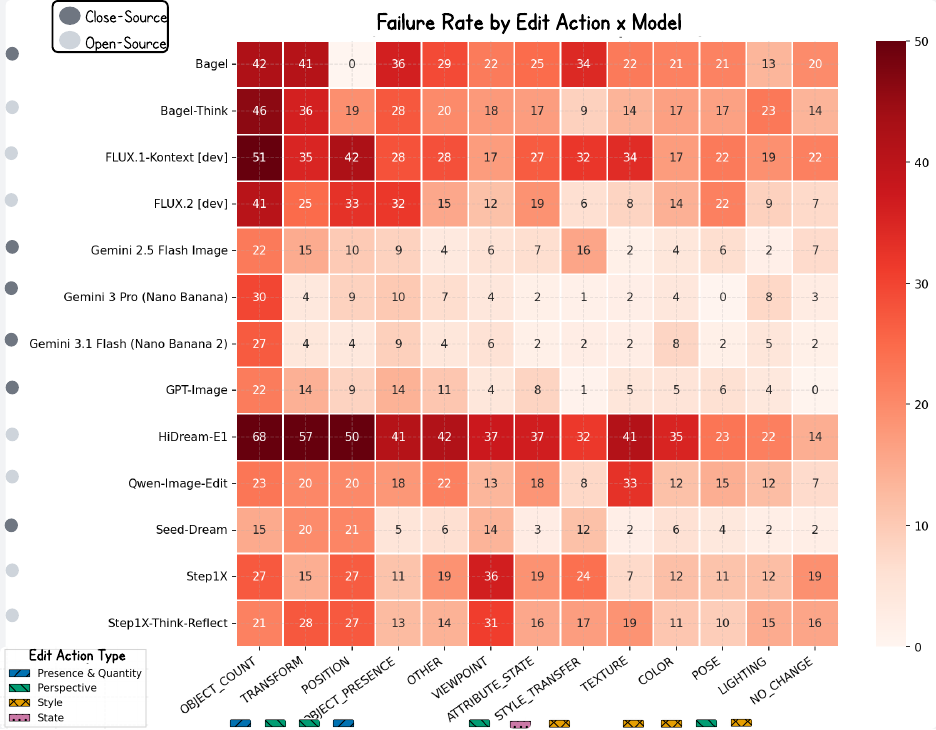}
    \caption{Heatmap illustrating the failure rate across different edit actions (x-axis) for each model (y-axis). More intense red shades correspond to a higher percentage of failure out of all entity-level edits.}
    \label{fig:edit_action_model}
\end{figure}

\paragraph{Failure Modes by Edit Action.}
Figure \ref{fig:edit_action_model} visualizes the failure rates of each model broken down by specific granular edit actions. Most notably, the heatmap exposes a universal bottleneck in spatial and quantitative reasoning. OBJECT COUNT yields the highest failure rates across virtually all models, serving as the primary point of failure even for the most advanced closed-source systems. Similarly, structural edits such as TRANSFORM and POSITION prove significantly more difficult to execute than stylistic changes. Furthermore, analyzing the thinking-enhanced variants reveals a surprising "trade-off" effect. While integrating a thinking phase successfully mitigates failures in semantic and stylistic mappings, such as Bagel-Think dropping its STYLE TRANSFER failure rate from 34 to 9, it can paradoxically destabilize fundamental operations. For instance, Step1X-Think-Reflect exhibits increased failure rates in TRANSFORM and TEXTURE compared to its base counterpart, suggesting that explicit multi-step reasoning might occasionally override robust, low-level visual priors. Finally, the data exposes a sharp dichotomy within the open-source tier: models like FLUX.2 rival closed-source systems on surface-level adjustments like TEXTURE (6) and LIGHTING (9), yet entirely collapse on structural tasks like OBJECT COUNT (41) and POSITION (33), indicating that current open-source alignment is heavily skewed toward appearance rather than spatial composition.

\begin{table}[!ht]
    \centering
    \caption{Summary of model performance (\eval) on Explicit prompts of the \dataset\ test set. \textbf{Bold} values indicate the winning score within each specific category group.}
    \label{tab:explicit}
    \begin{tabular}{l l | c}
        \toprule
        \textbf{Cat.} & \textbf{Model} & \textbf{Exp.} $\uparrow$ \\
        \midrule
        \multirow{4}{*}{\rotatebox{90}{\makecell{Open\\Source}}} 
            & Qwen-Image-Edit            & \textbf{8.05} \sd{2.51} \\
            & FLUX.2 [dev]               & 7.50 \sd{2.61}          \\
            & HiDream-E1                 & 5.43 \sd{2.80}          \\
            & FLUX.1-Kontext [dev]       & 6.97 \sd{2.92}          \\
        \midrule
        \multirow{4}{*}{\rotatebox{90}{\makecell{OS w/\\Thinking}}} 
            & Step1X-Think-Reflect       & 6.70 \sd{3.25}          \\
            & Step1X                     & \textbf{7.39} \sd{2.92} \\
            & Bagel-Think                & 5.82 \sd{3.19}          \\
            & Bagel                      & 6.35 \sd{2.79}          \\
        \midrule
        \multirow{4}{*}{\rotatebox{90}{\makecell{Closed\\Source}}} 
            & Gemini 3.1 Flash (NB 2)    & \textbf{8.99} \sd{1.97} \\
            & GPT-Image                  & 8.66 \sd{1.98}          \\
            & Gemini 3 Pro (NB)          & 8.80 \sd{2.08}          \\
            & Seed-Dream                 & 8.47 \sd{2.23}          \\
        \bottomrule
    \end{tabular}
\end{table}

\clearpage
\section{Qualitative Results}
\label{app:qual_res}

\begin{figure}[!ht]
    \centering
    \includegraphics[width=0.9\linewidth]{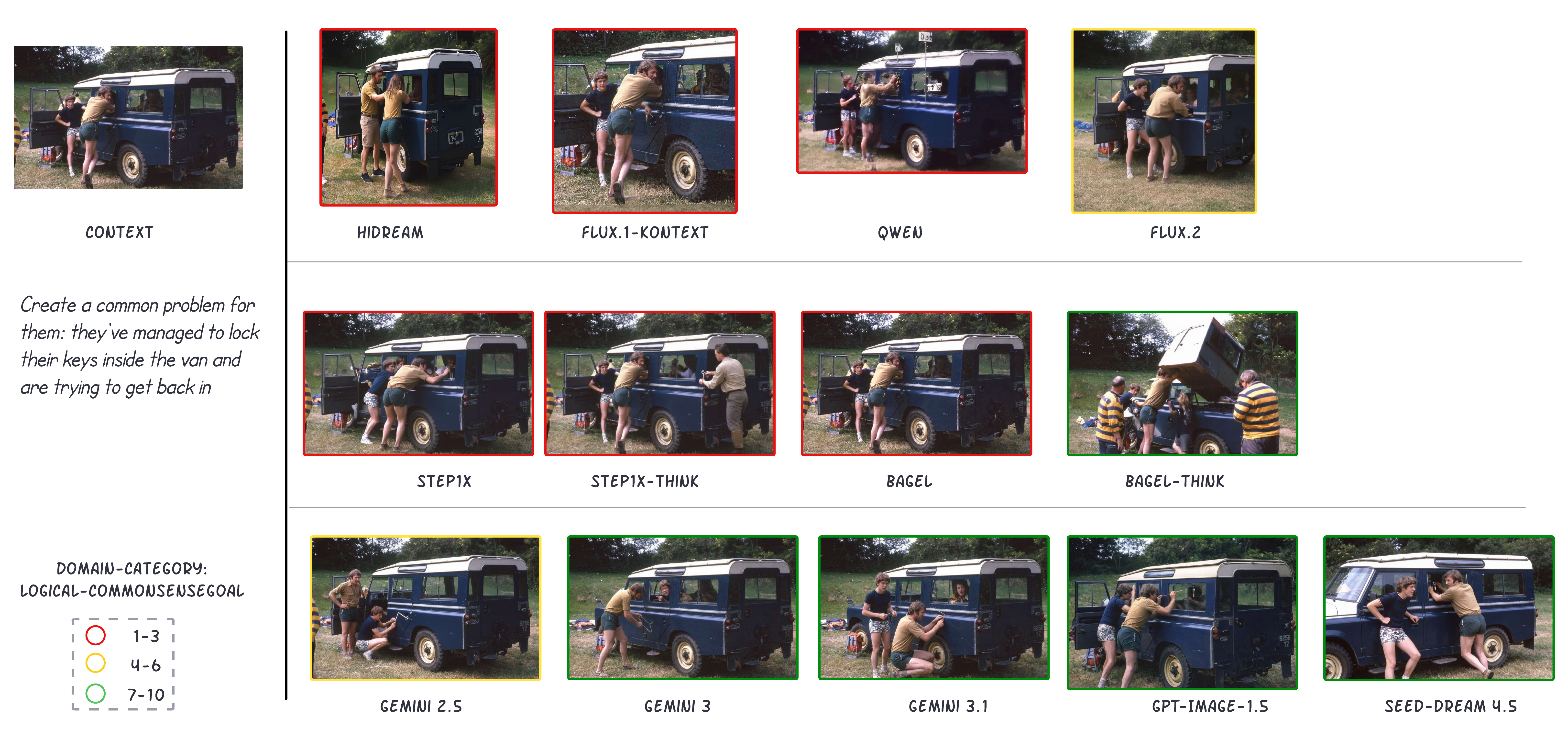}
    \caption{A qualitative comparison of all evaluated models on a sample from the logical domain. Frame colors indicate the corresponding \eval\ bin scores.}
    \label{fig:example_logical}
\end{figure}

\begin{figure}[!ht]
    \centering
    \includegraphics[width=0.9\linewidth]{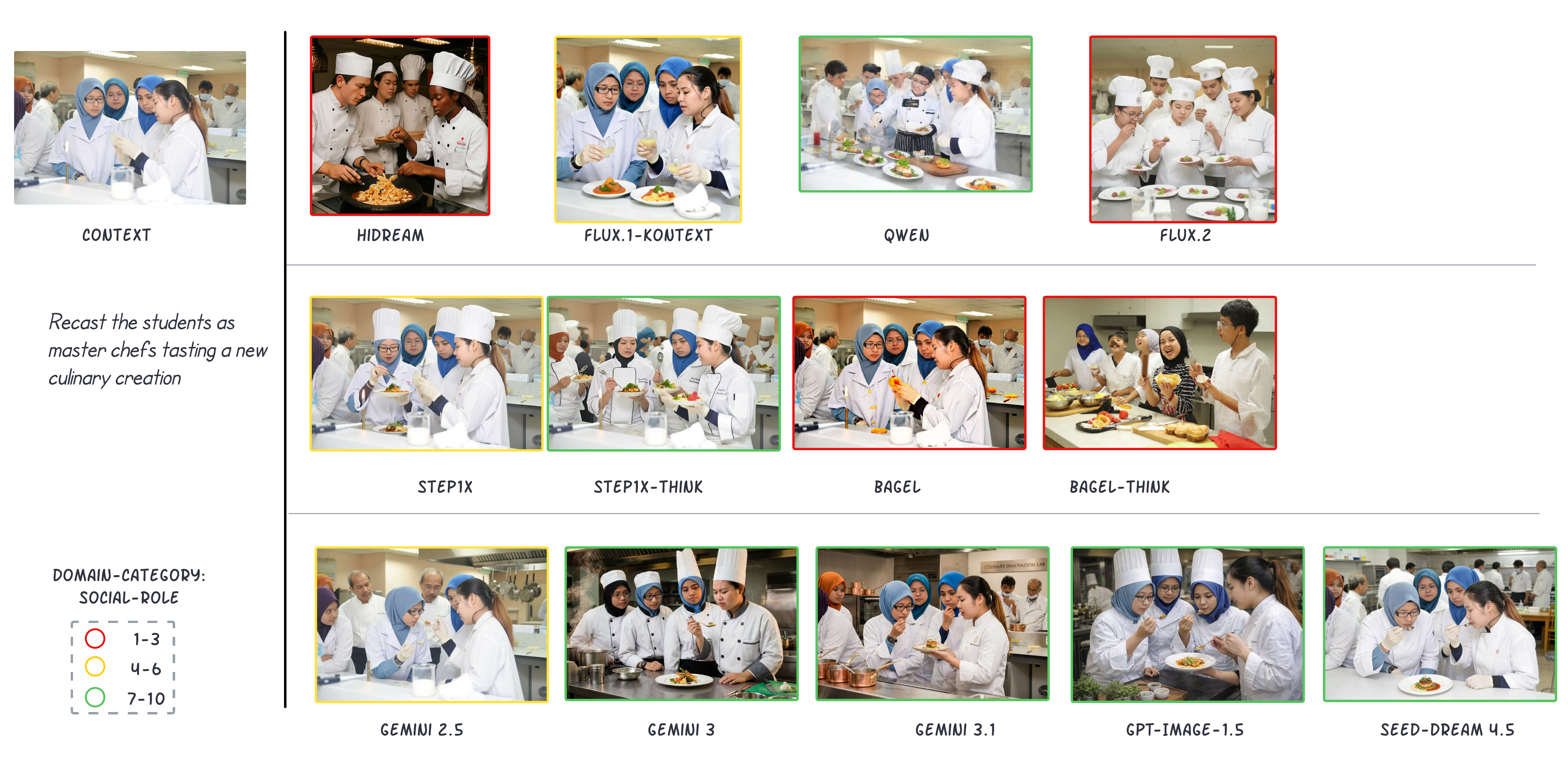}
    \caption{A qualitative comparison of all evaluated models on a sample from the social domain. Frame colors indicate the corresponding \eval\ bin scores.}
    \label{fig:example_social_role}
\end{figure}

\begin{figure}[!ht]
        \centering
        \includegraphics[width=0.9\linewidth]{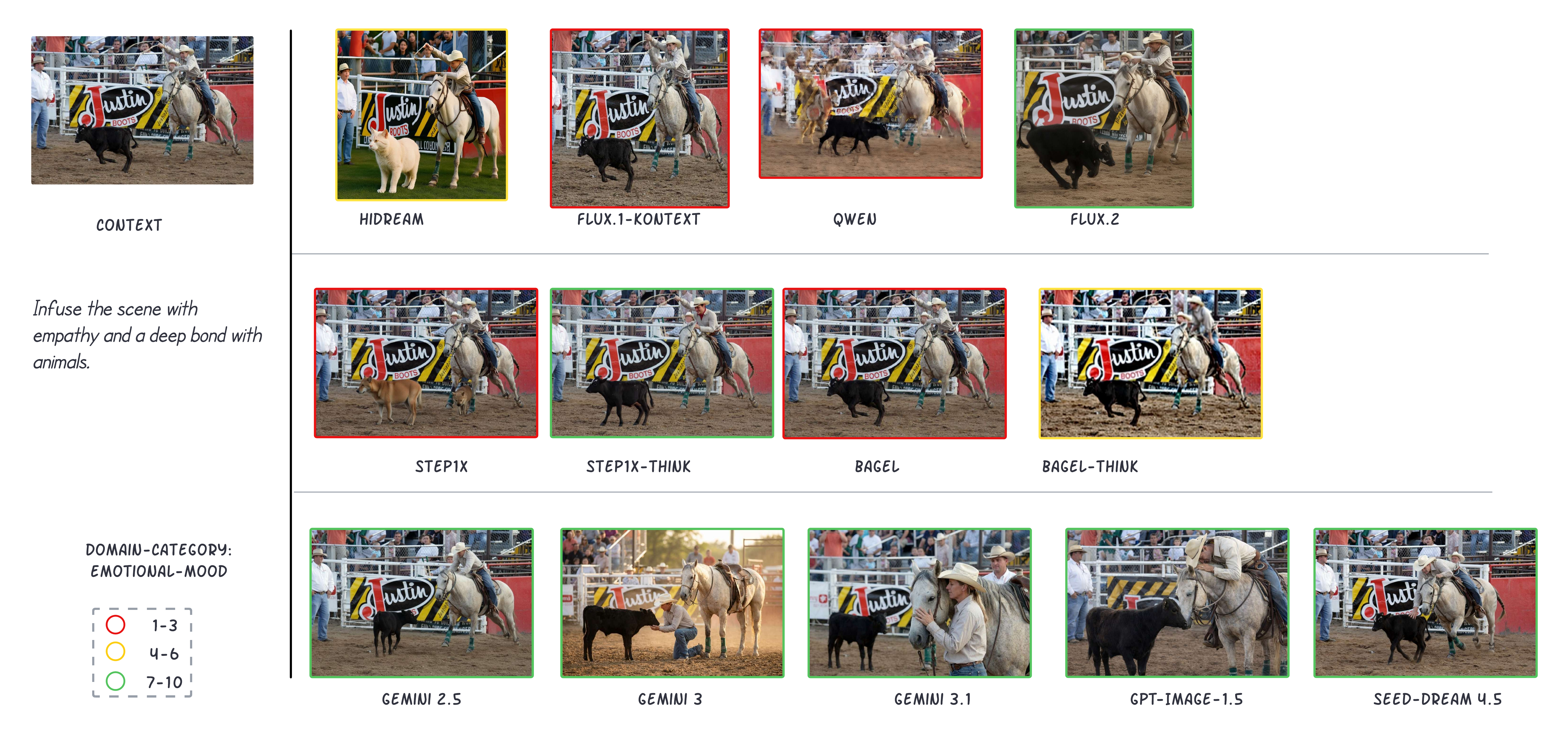}
        \caption{A qualitative comparison of all evaluated models on a sample from the emotional domain. Frame colors indicate the corresponding \eval\ bin scores.}
        \label{fig:example_emotional}
    \end{figure}

\begin{figure}[!ht]
    \centering
    \includegraphics[width=0.9\linewidth]{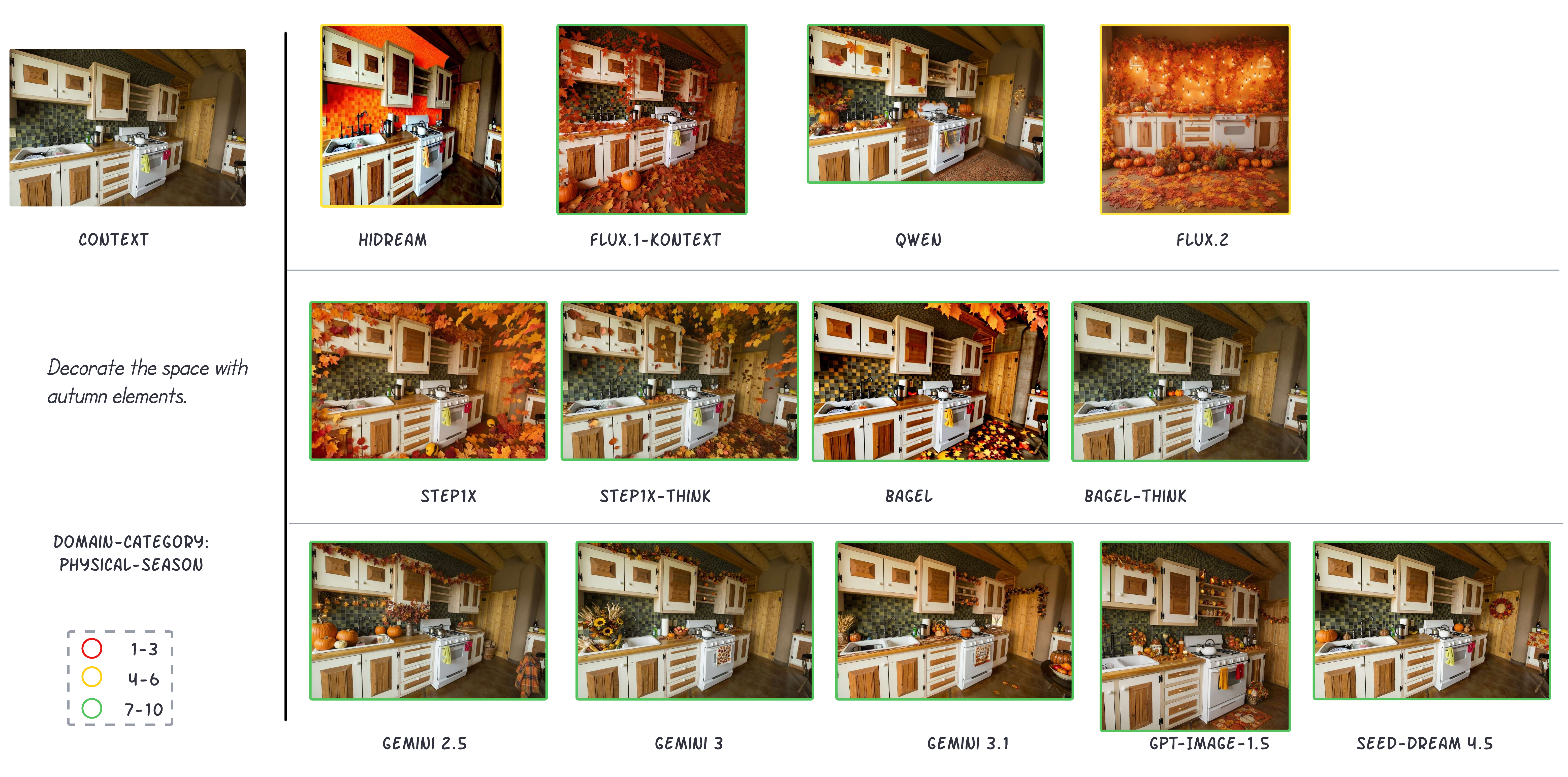}
    \caption{A qualitative comparison of all evaluated models on a sample from the physical domain. Frame colors indicate the corresponding \eval\ bin scores.}
    \label{fig:example_physic}
\end{figure}

\begin{figure}[!ht]
    \centering
    \includegraphics[width=0.9\linewidth]{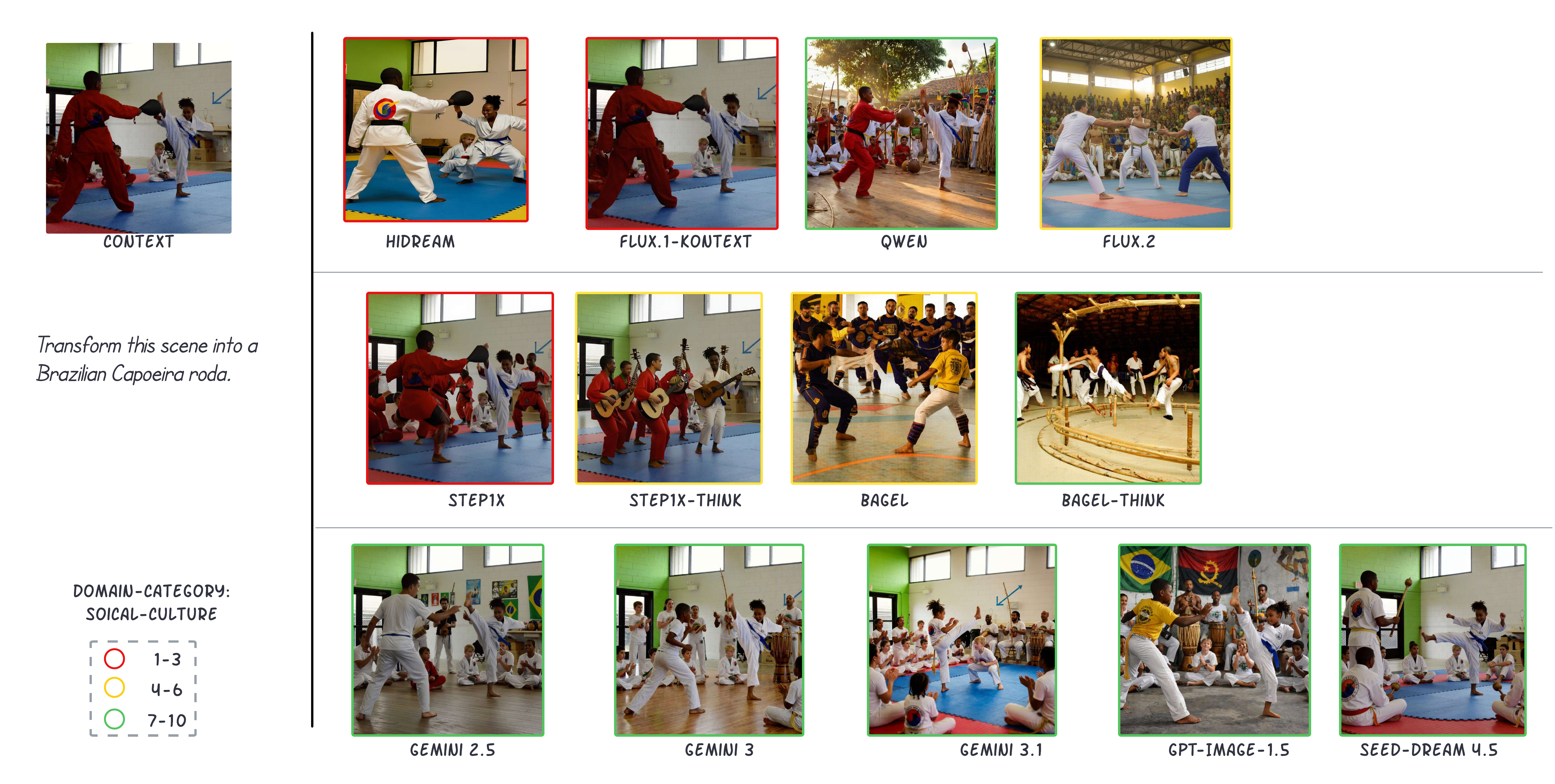}
    \caption{A qualitative comparison of all evaluated models on a sample from the social domain. Frame colors indicate the corresponding \eval\ bin scores.}
    \label{fig:example_social_culture}
\end{figure}






\end{document}